\documentclass[lettersize,journal]{IEEEtran}
\usepackage{amsmath,amsfonts}
\usepackage{algorithm}
\usepackage{array}
\usepackage{textcomp}
\usepackage{stfloats}
\usepackage{url}
\usepackage{verbatim}
\usepackage{graphicx}
\usepackage{cite}

\usepackage[noend]{algpseudocode}
\usepackage{enumitem}
\usepackage{subcaption}
\usepackage{xcolor}
\usepackage{amsthm, amssymb}
\usepackage{booktabs}
\usepackage{wrapfig}

\newtheorem{theorem}{Theorem}[section]
\newtheorem{assumption}[theorem]{Assumption}
\newtheorem{definition}[theorem]{Definition}
\newtheorem{lemma}[theorem]{Lemma}
\newtheorem{proposition}[theorem]{Proposition}

\DeclareMathOperator*{\argmin}{arg\,min}
\DeclareMathOperator*{\li}{li}
\DeclareMathOperator*{\Beta}{Beta}

\renewcommand{\textcolor}[2]{#2}
\renewcommand{\color}[2]{#2} 

\begin{document}

\title{Resource-Constrained Decentralized Federated Learning via Personalized Event-Triggering}

\author{Shahryar~Zehtabi,~\IEEEmembership{Student Member,~IEEE,}
        Seyyedali~Hosseinalipour,~\IEEEmembership{Senior Member,~IEEE,}
        and~Christopher~G.~Brinton,~\IEEEmembership{Senior Member,~IEEE}
\thanks{A preliminary version of this work \cite{zehtabi2022decentralized} appeared in the 2022 IEEE Conference on Decision and Control (CDC).}
\thanks{S. Zehtabi and C. G. Brinton are with the Elmore Family School of Electrical and Computer Engineering, Purdue University, West Lafayette, USA (e-mail: {szehtabi, cgb}@purdue.edu}
\thanks{S. Hosseinalipour is with the Department of Electrical Engineering, University at Buffalo–SUNY, NY, USA (email: alipour@buffalo.edu).}
}

\markboth{IEEE/ACM Transactions on Networking Special Issue on AI and Networking, January~2025}%
{Zehtabi \MakeLowercase{\textit{et al.}}: Event-Triggered Decentralized Federated Learning}


\maketitle

\begin{abstract}
Federated learning (FL) is a popular technique for distributing machine learning (ML) across a set of edge devices. In this paper, we study fully decentralized FL, where in addition to devices conducting training locally, they carry out model aggregations via cooperative consensus formation over device-to-device (D2D) networks. \textcolor{blue}{We introduce asynchronous, event-triggered communications among the devices to handle settings where access to a central server is not feasible.} To account for the inherent resource heterogeneity and statistical diversity challenges in FL, we define personalized communication triggering conditions at each device that weigh the change in local model parameters against the available local network resources. We theoretically recover the $\mathcal{O}{(\ln{k} / \sqrt{k})}$ convergence rate to the globally optimal model of decentralized gradient descent (DGD) methods in the setup of our methodology. We provide our convergence guarantees for the last iterates of models, under relaxed graph connectivity and data heterogeneity assumptions compared with the existing literature. To do so, we demonstrate a $B$-connected information flow guarantee in the presence of sporadic communications over the time-varying D2D graph. Our subsequent numerical evaluations demonstrate that our methodology obtains substantial improvements in convergence speed and/or communication savings compared to existing decentralized FL baselines.
\end{abstract}


\section{Introduction}\label{sec:intro}
\IEEEPARstart{F}{ederated} learning (FL) has emerged as a popular technique to distribute machine learning (ML) model training across a network of devices \cite{kairouz2021advances}. With initial deployments including next-word prediction across mobile devices \cite{konevcny2016federated}, FL is envisioned to serve many intelligence applications in edge/fog computing and the Internet of Things (IoT) \cite{chiang2016fog}.

In the conventional FL architecture, a set of devices is connected to a central server (e.g., at a base station) in a star topology configuration \cite{konevcny2016federated}. Devices conduct model updates locally based on their individual datasets, and the server periodically aggregates these local models into a global model, synchronizing devices with this global model to begin the next round of training. Several works in the past few years have built functionality into this architecture to manage different dimensions of heterogeneity that manifest in fog, edge, and IoT networks, including varying communication and computation abilities of devices \cite{wang2021network,wang2019adaptive} and varying statistical properties of local device datasets \cite{bonawitz2019towards,li2020federated}.

\textcolor{blue}{However, access to a central server is not practical in some cases; For example, when conducting online training over dynamic intelligent systems, such as smart vehicles that move across a city, they can rely on vehicle-to-vehicle (V2V) links rather than vehicle-to-edge/server communications \cite{beltran2023decentralized}.} In addition, the model aggregation step in FL can be resource intensive when it requires frequent uplink transmission of large models \cite{hosseinalipour2020federated}. In particular, it can lead to longer delays and larger bandwidth utilization for network links since there are multiple layers of network devices between the devices and the server \cite{hosseinalipour2022multi}. In wireless networks where device-to-server connectivity is energy intensive or unavailable, ad hoc structures formed through device-to-device (D2D) links serve as an efficient alternative \cite{chiang2016fog}. The proliferation of such settings motivates \textit{fully decentralized FL} \cite{lalitha2019peer, koloskova2020unified, liu2023yoga}, where the model aggregation step is distributed across devices (in addition to the data processing step).

In this paper, we propose a novel methodology to facilitate fully decentralized FL and analyze its convergence characteristics. Our methodology has two key components. The first component involves combining updates based on stochastic gradient descent of local ML models with \textit{cooperative consensus formation over the D2D graph topology} available across devices. Compared with more traditional decentralized optimization problems, the FL context introduces new challenges for these procedures due to (i) heterogeneity in device resources and local ML objectives, due to diversity in local datasets, and (ii) heterogeneity in wireless communication resources, which impacts the ability of devices to carry out consensus iterations. We consider these unique properties of decentralized FL in our algorithm design and analysis.

The central server in FL is also commonly employed for timing synchronization, i.e., to determine the time between global aggregations \cite{hosseinalipour2022multi}. To overcome the lack of a central timing mechanism in decentralized FL, and to alleviate resource utilization, the second component of our methodology is an \textit{asynchronous, event-triggered communication framework} for distributed ML consensus. Event-triggered communications offer several benefits in our setting. One, redundant communications can be reduced by defining event triggering conditions based on the variation of each device's model parameters. Also, eliminating the assumption that devices communicate in every iteration opens up the possibility of alleviating straggler problems, which is a prevalent concern in FL \cite{hosseinalipour2020federated}. Third, we can eliminate redundant computations at each device by limiting the aggregation to only when new parameters are received.

\subsection{Related Work}
\begin{table}[t]
    \centering
    \setlength{\tabcolsep}{0.3mm}
    \caption{\textcolor{blue}{Comparison of our work against representative works in the decentralized FL literature.}}
    \textcolor{blue}{\begin{tabular}{c|c|c|c|c|c|c}
        \toprule
        Paper & \begin{tabular}{c} Time- \\ Varying \\ Graph \end{tabular} & \begin{tabular}{c} Sporadic \\ Comm. \end{tabular} & \begin{tabular}{c} Smooth. \\ Assump. \end{tabular} & \begin{tabular}{c} Constant \& \\ Diminishing \\ Step Size \end{tabular} & \begin{tabular}{c} Convex \& \\ Non-Convex \\ Analysis \end{tabular} & \begin{tabular}{c} Last \\ Iterates \\ Conv. \\ (Convex) \end{tabular}
        \\
        \midrule
        \cite{koloskova2020unified} & & \checkmark & \checkmark & & \checkmark &
        \\
        \hline
        \cite{nedic2009distributed, nedic2013distributed} & \checkmark & & & & &
        \\
        \hline
        \cite{sundhar2010distributed} & \checkmark & & & \checkmark & &
        \\
        \hline
        \cite{sun2022decentralized, mishchenko2022proxskip} & & & \checkmark & & & \checkmark
        \\
        \hline
        \textbf{Ours} & \pmb{\checkmark} & \pmb{\checkmark} & \pmb{\checkmark} & \pmb{\checkmark} & \pmb{\checkmark} & \pmb{\checkmark}
        \\
        \bottomrule
    \end{tabular}}
    \label{tab:comparison}
\end{table}

We discuss related work in (i) distributed learning through consensus on graphs and (ii) FL over heterogeneous network systems. Our work lies at the intersection of these areas, and Table~\ref{tab:comparison} illustrates the contributions of our work compared to existing literature.

\subsubsection{Consensus-based distributed optimization}
There is a rich literature on distributed optimization on graphs using consensus algorithms, for example, \cite{nedic2009distributed, nedic2010constrained, pu2021distributed, nedic2014distributed, xin2018linear, koloskova2020unified}. For connected, undirected graph topologies, symmetric and doubly-stochastic transition matrices can be constructed for consensus iterations. In typical approaches \cite{nedic2009distributed, nedic2010constrained}, each device maintains a local gradient of the target system objective (e.g., minimizing the consensus error across nodes), with the consensus matrices designed to satisfy additional convergence criteria outlined in \cite{boyd2004fastest, xiao2004fast}.
More recently, gradient tracking optimization techniques have been developed in which the global gradient is learned simultaneously along with local parameters \cite{pu2021distributed}. Also, \cite{liu2024decentralized, nguyen2023performance} present a variation of gradient tracking algorithms where devices conduct multiple local gradient steps at each iteration. Other works have considered the distributed optimization problem over directed graphs, which is harder since constructing doubly-stochastic transition matrices is not a straightforward task for general directed graphs \cite{gharesifard2010does}. To resolve this issue, methods such as the push-sum protocol \cite{nedic2014distributed} have been proposed, where an extra optimizable parameter is introduced at each device in order to independently learn the right (or left) eigenvector corresponding to the eigenvalue of 1 of the transition matrices \cite{xin2018linear}. More recently, dual transition matrices have been studied, where two distinct transition matrices are designed to exchange model parameters and gradients separately, one column-stochastic and the other row-stochastic \cite{saadatniaki2020decentralized, pu2020push, xin2018linear}. Moreover, asynchronous communications have also been researched in the literature \cite{assran2020asynchronous, bof2018multiagent, liao2024asynchronous}.

On the other hand, event-triggered methods have received significant research attention in the conventional distributed optimization literature \cite{yang2019survey}. However, in federated learning (FL), there are only a handful of papers thus far which have studied event-triggered communication mechanisms, e.g., \cite{george2020distributed}. Event-triggering for inter-device communications poses unique challenges in the FL context due to different types of heterogeneity which become extremely pronounced in these setups, \textcolor{blue}{where a model is being trained} over real-world wireless devices \cite{kairouz2021advances}: (i) diversity in local dataset statistics \cite{zhao2018federated}, which can have significant impacts on convergence behavior, since gradient iterations on these local datasets will tend to pull models apart \cite{wang2020optimizing,hosseinalipour2020federated}; and (ii) heterogeneity in device resources \cite{konevcny2016federated, chen2021communication, bonawitz2019towards}. Our methodology incorporates both of these factors, and our theoretical results reveal the impact of non-IID local data distributions on convergence characteristics of model training in decentralized FL setups. In doing so, it is important for us to bound convergence directly in terms of the statistical heterogeneity of the datasets, rather than on the (sub)-gradients as is done in existing decentralized optimization works like \cite{sundhar2010distributed, nedic2009distributed}.

\subsubsection{Resource-efficient federated learning}
Several recent works in FL have investigated techniques for improving the communication and computation efficiency across devices. A popular line of research has aimed to adaptively control the FL process based on device capabilities, e.g., \cite{nishio2019client,nguyen2020fast,diao2020heterofl,wang2019adaptive,gu2021fast}. In \cite{wang2019adaptive}, the authors studied FL convergence under a total network resource budget, in which the server adapts the frequency of global aggregations. Others \cite{nishio2019client,nguyen2020fast,gu2021fast} have considered FL under partial device participation, where the communication and processing capabilities of devices are taken into account when assessing which devices will participate in each training round. \cite{diao2020heterofl} removed the necessity that every local device has to optimize the full model as the server, allowing weaker devices to take smaller subsets of the model to optimize. Furthermore, techniques such as quantization \cite{shlezinger2020federated, lan2023communication} and sparsification \cite{renggli2019sparcml} have also been studied to reduce the communication and computation overhead of the FL algorithms.

Unlike these works, we focus on novel learning topologies for decentralized FL. In this respect, some recent work \cite{savazzi2020federated,lalitha2019peer,hosseinalipour2022multi,lin2021semi} has proposed D2D communication approaches for collaborative learning over local device graphs. \cite{hosseinalipour2022multi,lin2021semi, sun2023semi} investigated a semi-decentralized FL methodology across hierarchical networks, where local model aggregations are conducted via D2D-based cooperative consensus formation to reduce the frequency of global aggregations by the coordinating node. In our work, we consider the fully decentralized setting, where a central node is not available, as in \cite{savazzi2020federated,lalitha2019peer, zehtabi2025decentralized}: along with local model updates, devices conduct consensus iterations with their neighbors in order to gradually minimize the global machine learning loss in a distributed manner. However, such techniques are not sensitive to the presence of heterogeneous communication resources across devices. Different from \cite{zehtabi2025decentralized}, our methodology incorporates asynchronous event-triggered communications, where local resource levels are factored into event thresholds to account for device heterogeneity. This introduces a key challenge solved in our analysis, as it must obtain and leverage connectivity guarantees on the information flow graph (rather than the physical D2D graph). We will see that this approach leads to substantial improvements in model convergence time compared with non-heterogeneous/non-personalized thresholding.

\subsection{Outline and Summary of Contributions}

\begin{itemize}
	\item We develop a novel methodology for fully decentralized FL, with model aggregations occurring via cooperative model consensus iterations (Sec.~\ref{sec:method}). In our methodology, communications are asynchronous and event-driven. With event thresholds defined to incorporate local model evolution and resource availability, our methodology adapts to the two salient heterogeneity dimensions in decentralized FL: limited resource availability and non-IID local dataset statistics across devices.
	
    \item We provide a detailed convergence analysis of our methodology, showing that using a diminishing step size, each device arrives at the globally optimal model over a time-varying consensus graph at an $\mathcal{O}{(\ln{k} / \sqrt{k})}$ rate (Sec.~\ref{sec:conv}). Our results are obtained based on statistical heterogeneity across local datasets, rather than a more restrictive bounded gradients assumption common in literature. Moreover, they do not impose overly restrictive connectivity requirements on the underlying D2D communication graph, so long as it satisfies a connectivity assumption over any $B$-consecutive iterations.

    \item To obtain these results, we demonstrate information flow guarantees in the presence of sporadic communications, making a distinction between physical connectivity of the underlying network graph and the information flow graph of the exchanged parameters among the devices (Proposition~\ref{proposition:conn}). Moreover, we lay out constraints on the local gradient step size to ensure a necessary spectral radius on this graph (Proposition~\ref{proposition:nonIncreasing}). This allows us to derive the convergence rate in Theorems~\ref{theorem:constant} and \ref{theorem:diminishing} for the model itself and not its cumulative average,\footnote{The cumulative average of the model is defined as $(1/T) \sum_{t=0}^{T-1}{\mathbf{w}^{(t)}}$.} contrary to the existing trend in decentralized FL.

    \item We conducted numerical experiments to compare our methodology with baselines in decentralized FL, as well as a randomized gossip algorithm using two real-world machine learning task datasets (Sec.~\ref{sec:simulation}). We show that our method is capable of reducing the model training communication time compared to decentralized FL baselines. Also, we find that the convergence rate of our method scales well with consensus graph connectivity.
\end{itemize}
This paper is an extension of our conference version of this work\cite{zehtabi2022event}. Compared to \cite{zehtabi2022event}, we make the following additional contributions: (1) connectivity of the information flow graph between devices is theoretically proven in Proposition~\ref{proposition:conn}, i.e., we guarantee that all devices will benefit from every other device in the federated learning system, despite the communications being sporadic, and the underlying physical network being time-varying, (2) the traditional assumption of bounded gradients has been replaced with two assumptions which are less strict in Assumptions~\ref{assump:smooth_convex_graddiv}: Lipschitz gradient continuity and bounded gradient diversity, (3) we obtain a non-asymptotic rate of convergence for our method, by recovering the well-known $\mathcal{O}(\ln{k} / \sqrt{k})$ sub-linear rate for distributed gradient descent like algorithms when using a diminishing learning rate in Theorem~\ref{theorem:diminishing} and (4) we conduct new experiments on an additional dataset, more graph topologies and an additional distribution for bandwidth sampling, to further validate the advantages of our proposed methodology.

\subsection{Notations}
Arguments for functions are denoted with parentheses, e.g., $f{( x )}$ implies $x$ is an argument for function $f$. The iteration index for a parameter is indicated via superscripts, e.g., $h^{(k)}$ is the value of the parameter $h$ at iteration $k$. Device indices are given via subscripts, e.g., $h_i^{(k)}$ refers to parameter belonging to device $i$. We write a graph $\mathcal{G}$ with a set of nodes (devices) $\mathcal{V}$ and a set of edges (links) $\mathcal{E}$ as $\mathcal{G} = ( \mathcal{V}, \mathcal{E} )$.

We denote vectors with lowercase boldface, e.g., $\mathbf{x}$, and matrices with uppercase boldface, e.g., $\mathbf{X}$. All vectors $\mathbf{x} \in \mathbb{R}^{d \times 1}$ are column vectors, except in certain cases where average vectors $\bar{\mathbf{x}} \in \mathbb{R}^{1 \times d}$ and optimal vectors $\mathbf{w}^\star \in \mathbb{R}^{1 \times d}$ are row vectors. $\langle \mathbf{x}, \mathbf{x}' \rangle$ and $\langle \mathbf{X}, \mathbf{X}' \rangle$ denote the inner product of two vectors $\mathbf{x}, \mathbf{x}'$ of equal dimensions and the Frobenius inner product of two matrices $\mathbf{X},\mathbf{X}'$ of equal dimensions, respectively. Moreover, ${\| \mathbf{x} \|}$ and ${\| \mathbf{X} \|}$ denote the $2$-norm of the vector $\mathbf{x}$, and the Frobenius norm of the matrix $\mathbf{X}$, respectively. The spectral norm of the matrix $\mathbf{X}$ is written as $\rho{( \mathbf{X} )}$.

Note that for brevity, all mathematical proofs have been deferred to the Appendices at the end of the manuscript.

\section{Methodology and Algorithm} \label{sec:method}
\noindent In this section, we develop our methodology for decentralized FL with event-triggered communications. After discussing preliminaries of the model in FL (Sec.~\ref{ssec:FL}), we present our cooperative consensus algorithm for distributed model aggregations (Sec.~\ref{ssec:event}).
We then present the events in our event-triggered algorithm as iterative relations, \textcolor{blue}{which enables our theoretical analysis} (Sec.~\ref{ssec:iterateRelations}).

\subsection{Device and Learning Model}
\label{ssec:FL}
We consider a network of $m$ devices/nodes, collected by set $\mathcal{M}$, $m=|\mathcal{M}|$, which are engaged in distributed training of a machine model. Under the FL framework, each device $i\in\mathcal{M}$ trains a local model $\mathbf{w}_i$ using its own generated dataset $\mathcal{D}_i$. Each data point $\xi\triangleq(\mathbf{x}_\xi, y_\xi)\in\mathcal{D}_i$ consists of a feature vector $\mathbf{x}_\xi$ and a target label $y_\xi$. The performance of the local model is measured via the
local loss $F_i{( . )}$ as
\begin{equation} \label{eqn:localloss}
	F_i{\left( \mathbf{w} \right)} = \sum_{\xi\in\mathcal{D}_i}\ell_\xi\left(\mathbf{w} \right),
\end{equation}
where $\ell_\xi(\mathbf{w} )$ is the loss of the model at the data point $\xi$ (e.g., squared prediction error) under parameter realization $\mathbf{w}\in \mathbb{R}^n$, with $n$ denoting the dimension of the target model.
The global loss is defined in terms of these local losses as
\begin{equation} \label{eqn:globalModel}
	F{\left(\mathbf{w} \right)} = \frac1m \sum_{i\in\mathcal{M}}{F_i{\left( \mathbf{w} \right)}}.
\end{equation}

The goal of the training process is to find an optimal parameter vector $\mathbf{w}^\star$ that minimizes the global loss function, that is, $\mathbf{w}^\star=\argmin_{\mathbf{w}\in\mathbb{R}^n} F{( \mathbf{w} )}$.
In the distributed setting, we desire $\mathbf{w}_1 = \cdots = \mathbf{w}_m = \mathbf{w}^*$ at the end of the training process, which requires a synchronization mechanism. In conventional FL, as discussed in Sec.~\ref{sec:intro}, synchronization is conducted periodically by a central coordinator globally aggregating the local models. However, in this work, we are interested in a fully decentralized setting where no such central node exists. Thus, in addition to using optimization techniques to minimize local loss functions, we must develop a technique to reach consensus over the parameters in a distributed manner.

To achieve this, we propose \textit{Event-triggered Federated learning with Heterogeneous Communication thresholds} ({\tt EF-HC}). In {\tt EF-HC}, devices conduct D2D communications during the model training period to synchronize their locally trained models and avoid overfitting to their local datasets. The overall {\tt EF-HC} algorithm executed by each device is given in Alg.~\ref{alg:efhc}. Two vectors of model parameters are kept on each device $i$: (i) its instantaneous \textit{main} model parameters $\mathbf{w}_i$, and (ii) the \textit{auxiliary} model parameters $\mathbf{\widehat{w}}_i$, which is the outdated version of its main parameters that had been broadcast to neighbors. Decentralized ML is conducted over the (time-varying, undirected) device graph through a sequence of four events detailed in Sec.~\ref{ssec:event}. \textcolor{blue}{Although in our distributed setup there is no physical notion of a global iteration, we introduce the universal iteration variable $k$ for analysis purposes \cite{tsitsiklis2003distributed}. In other words, event-triggering in our paper implies that not every iteration $k$ includes full participation of devices in inter-device communications, which is different from how synchronous DFL works \cite{nedic2009distributed, sundhar2010distributed, nedic2013distributed} model iterate updates.}

\subsection{Network Model and Event-Triggering}
\label{ssec:event}
We consider the \textit{physical network} graph $\mathcal{G}^{(k)} = ( \mathcal{M}, \mathcal{E}^{(k)} )$ among devices, where $\mathcal{E}^{(k)}$ is the set of edges available at iteration $k$ in the underlying time-varying communication graph. We assume that link availability varies over time according to the underlying device-to-device communication protocol in place \cite{hosseinalipour2020federated}. In each iteration, some of the edges are used for the transmission/reception of model parameters between devices. To represent this process, we define the \textit{information flow} graph $\mathcal{G}'^{(k)} = ( \mathcal{M}, \mathcal{E}'^{(k)} )$, which is a subgraph of $\mathcal{G}^{(k)}$. $\mathcal{E}'^{(k)}$ only contains the links in $\mathcal{E}^{(k)}$ that are being used at iteration $k$ to exchange parameters. Based on this, we denote the neighbors of device $i$ in iteration $k$ as $\mathcal{N}_i^{(k)} = \lbrace  j :  (i,j)\in\mathcal{E}^{(k)}\, , j \in \mathcal{M}\}$, with node degree $d_i^{(k)} = \lvert \mathcal{N}_i^{(k)} \rvert$. We also denote neighbors of $i$ that communicate directly with it in iteration $k$ as $\mathcal{N}'^{(k)}_i = \lbrace  j :  (i,j)\in\mathcal{E}'^{(k)}\, , j \in \mathcal{M}\}$. Additionally, the aggregation weights associated with the link $(i, j) \in \mathcal{E}^{(k)}$ and $(i, j) \in \mathcal{E}'^{(k)}$ are defined as $\beta_{ij}^{(k)}$ and $p_{ij}^{(k)}$, respectively, with $p_{ij}^{(k)} = \beta_{ij}^{(k)}$ if the link $(i, j)$ is used for aggregation at iteration $k$, and $p_{ij}^{(k)} = 0$ otherwise.

\begin{algorithm}[t]
	\small
	\caption{{\tt EF-HC} procedure for device $i$.}
	\label{alg:efhc}
	\textbf{Input:} $K$\\
	Initialize $k=0$, $\mathbf{w}_i^{(0)} = \mathbf{\widehat{w}}_i^{(0)}$
	\begin{algorithmic}[1]
		\While{$k\leq K$}
		
		\Comment{\textbf{Event 1. } Neighbor Connection Event}
		\If {device $j$ is connected to device $i$} \label{event:conn:begin}
		\State device $i$ appends device $j$ to its list of neighbors
		\State device $i$ sends $\mathbf{w}_i^{(k)}$ and $d_i^{(k)}$ to device $j$
		\State device $i$ receives $\mathbf{w}_j^{(k)}$ and $d_j^{(k)}$ from device $j$
		\ElsIf {device $j$ is disconnected from device $i$}
		\State device $i$ removes device $j$ from its list of neighbors
		\EndIf \label{event:conn:end}
		
		\Comment{\textbf{Event 2. } Broadcast Event}
		\If {${( 1/n )}^\frac12 {\| \mathbf{w}_i^{(k)} - \mathbf{\widehat{w}}_i^{(k)} \|}_2 \geq r \rho_i \gamma^{(k)}$} \label{event:broadcast:begin}
		\State device $i$ broadcasts $\mathbf{w}_i^{(k)}$, $d_i^{(k)}$ to all neighbors $j\in\mathcal{N}_i^{(k)}$
		\State device $i$ receives $\mathbf{w}_j^{(k)}$, $d_j^{(k)}$ from all neighbors $j\in\mathcal{N}_i^{(k)}$
		\State $\mathbf{\widehat{w}}_i^{(k+1)} = \mathbf{w}_i^{(k)}$
		\EndIf \label{event:broadcast:end}
		
		\Comment{\textbf{Event 3. } Aggregation Event}
		\If {Parameters $\mathbf{w}_j^{(k)}$, $d_j^{(k)}$ received from neighbor $j$} \label{event:agg:begin}
		\State $\mathbf{w}_i^{(k+1)} = \mathbf{w}_i^{(k)} + \sum_{j \in \mathcal{N}'^{(k)}_i}{\beta_{ij}^{(k)} ( \mathbf{w}_j^{(k)} - \mathbf{w}_i^{(k)} )} $
		\EndIf \label{event:agg:end}
		
		\Comment{\textbf{Event 4. } Gradient Descent Event}
		\State device $i$ conducts SGD iteration $\mathbf{w}_i^{(k+1)} = \mathbf{w}_i^{(k)} - \alpha^{(k)} \mathbf{g}_i^{(k)}$ \label{event:sgd:begin}
		\State $k \leftarrow k+1$ \label{event:sgd:end}
		\EndWhile
	\end{algorithmic}
\end{algorithm}

\begin{figure*}[t]
	\begin{center}
		\includegraphics[width=\textwidth]{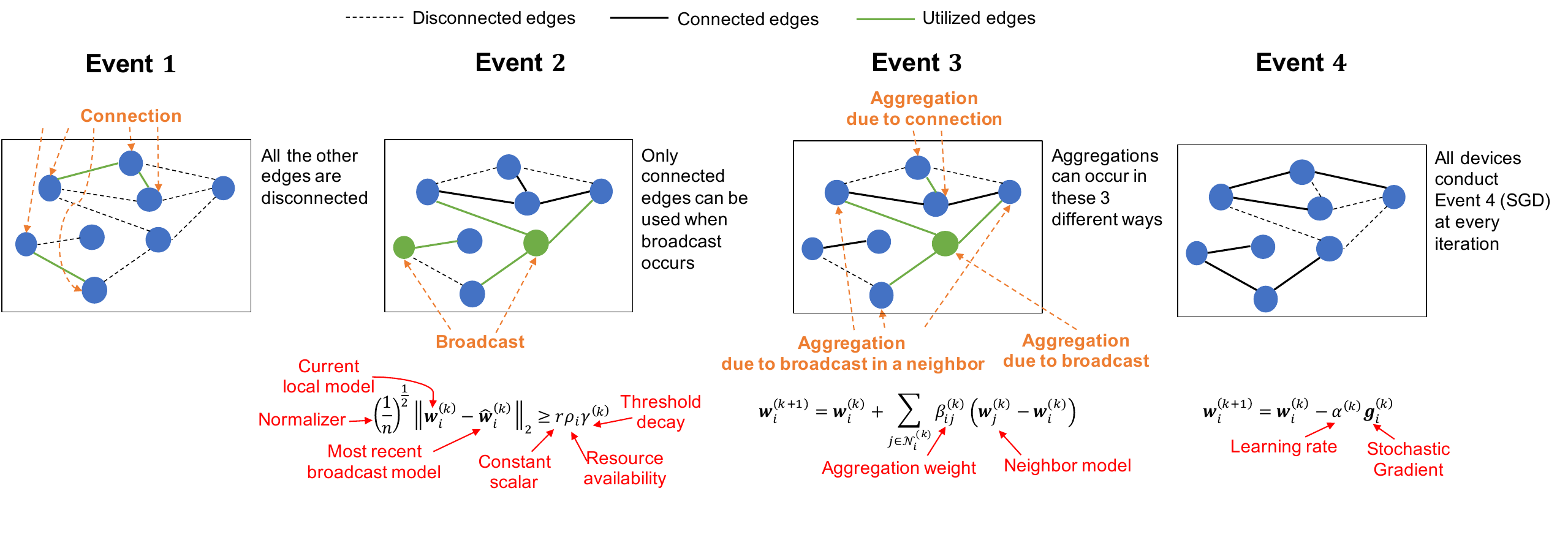}
	\end{center}
	\vspace{-10mm}
	\caption{\small{System diagram of a time-varying decentralized system, illustrating the four events of Alg.~\ref{alg:efhc}, namely (i) neighbor connection, (ii) model broadcast, (iii) model aggregation, and (iv) stochastic gradient descent.}}
	\label{fig:diag}
	\vspace{-5mm}
\end{figure*}

In {\tt EF-HC}, there are four types of events:

\noindent \underline{\textbf{Event 1: Neighbor connection}}

The first event (lines~\ref{event:conn:begin}-\ref{event:conn:end} of Alg.~\ref{alg:efhc}) is triggered at device $i$ if new devices connect to it or existing devices disconnect from it due to the time-varying nature of the graph. Model parameters $\mathbf{w}_i^{(k)}$ and the degree of the device $i$ at that time $d_i^{(k)}$ are exchanged with this new neighbor. Consequently, this results in an aggregation event (Event 3) on both devices.

\noindent \underline{\textbf{Event 2: Broadcast}}

If the normalized difference between $\mathbf{w}_i^{(k)}$ and $\mathbf{\widehat{w}}_i^{(k)}$ at device $i$ is greater than a \textit{threshold} value $r \rho_i \gamma^{(k)}$, i.e.,
\begin{equation} \label{eq:thresh}
	{\left( 1/n \right)}^\frac12 {\| \mathbf{w}_i^{(k)} - \mathbf{\widehat{w}}_i^{(k)} \|}_2 \geq r \rho_i \gamma^{(k)},
\end{equation}
then a broadcast event is triggered at that device (lines~\ref{event:broadcast:begin}-\ref{event:broadcast:end} of Alg.~\ref{alg:efhc}). In other words, communication at a device is triggered once the instantaneous local model is sufficiently different from the outdated one. When this event triggers, device $i$ broadcasts $\mathbf{w}_i^{(k)}$ and its degree $d_i^{(k)}$ to all its neighbors and receives the same information from them. Note that if a neighbor device $j$ is not available for communication with device $i$ at a certain iteration $k$, then $j$ would not be considered inside the neighbor set of $i$ in that iteration, i.e., $\mathcal{N}_i^{(k)}$.

The threshold $r \rho_i \gamma^{(k)}$ is treated as personalized/heterogeneous across devices $i\in\mathcal{M}$, to assess whether the gain from a consensus iteration on the instantaneous main models at the devices will be worth the induced utilization of network resources. Specifically, (i) $r > 0$ is a scaling hyperparameter value; (ii) $\gamma^{(k)} > 0$ is a decaying factor that accounts for smaller expected variations in local models over time, and $\lim_{k \to \infty}{\gamma^{(k)}} = 0$; and (iii) $\rho_i$ quantifies the availability of resources of device $i$. See \cite{zehtabi2022decentralized} for some remarks on ${(1/n)}^{\frac12}$, $r$ and $\gamma^{(k)}$.

The development of ${( 1/n )}^\frac12 {\| \mathbf{w}_i^{(k)} - \mathbf{\widehat{w}}_i^{(k)} \|}_2$ and the condition $r \rho_i \gamma^{(k)}$ is one of our contributions relative to existing event-triggered schemes~\cite{george2020distributed}. For example, in a bandwidth-limited environment, the transmission delay of the model transfer will be inversely proportional to the bandwidth among two devices. Thus, to decrease the latency of model training,
$\rho_i$ can be defined inversely proportional to the bandwidth, promoting a lower frequency of communication on devices with less available bandwidth. In {\tt EF-HC}, we set $\rho_i \propto \frac1{b_i}$, where $b_i$ is the average bandwidth on the outgoing links of the device $i$.

\noindent \underline{\textbf{Event 3: Aggregation}}

Following a broadcast event (Event 2) or a neighbor connection event (Event 1) on device $i$, an aggregation event (lines~\ref{event:agg:begin}-\ref{event:agg:end} of Alg.~\ref{alg:efhc}) is triggered on device $i$ and all its neighbors. This aggregation is carried out through a distributed weighted averaging consensus method~\cite{xiao2004fast} as
\begin{equation}
	\mathbf{w}_i^{(k+1)} = \mathbf{w}_i^{(k)} + \sum_{j \in \mathcal{N}'^{(k)}_i}{\beta_{ij}^{(k)} ( \mathbf{w}_j^{(k)} - \mathbf{w}_i^{(k)} )},
\end{equation}
where $\beta_{ij}^{(k)}$ is the aggregation weight that device $i$ assigns to parameters received from device $j$ in iteration $k$. The aggregation weights $\{\beta_{ij}^{(k)}\}$ for graph $\mathcal{G}^{(k)}$ can be selected based on the degree of neighbors, as will be discussed in Sec.~\ref{ssec:assumptions}.

\noindent \underline{\textbf{Event 4: Gradient descent}}

Each device $i$ conducts stochastic gradient descent (SGD) iterations for local model training (lines~\ref{event:sgd:begin}-\ref{event:sgd:end} of Alg.~\ref{alg:efhc}). Formally, device $i$ obtains $\mathbf{w}_i^{(k+1)} = \mathbf{w}_i^{(k)} - \alpha^{(k)} \mathbf{g}_i^{(k)}$, where
$\alpha^{(k)}$ is the step size, and $\mathbf{g}_i^{(k)}$ is the stochastic gradient approximation defined as $\mathbf{g}_i^{(k)} = (1 / | \mathcal{S}_i^{(k)} |) \sum_{\mathbf{\xi} \in \mathcal{S}_i^{(k)}} \nabla \ell_\xi{( \mathbf{w}_i^{(k)} )}$. Here, $\mathcal{S}_i^{(k)}$ denotes the set of data points (mini-batch) used to compute the gradient, chosen uniformly at random from the local dataset. In our analysis, we define

\begin{equation} \label{eqn:g}
	\mathbf{g}_i^{(k)} = \nabla F_i{( \mathbf{w}_i^{(k)} )} + \mathbf{\epsilon}_i^{(k)},
\end{equation}
in which $\nabla F_i{( \mathbf{w}_i^{(k)} )}$ is the gradient of $F_i$ at $\mathbf{w}_i^{(k)}$, and $\mathbf{\epsilon}_i^{(k)}$ is the error due to the stochastic gradient approximation.

\subsection{Iterate Relations} \label{ssec:iterateRelations}
We now express the model updates conducted in Alg.~\ref{alg:efhc} in an iterative format, which will be useful in our subsequent theoretical analysis. Rewriting the event-based updates of Alg.~\ref{alg:efhc} into one line of iterative model update, we get
\begin{equation} \label{eqn:w}
	\mathbf{w}_i^{(k+1)} \hspace{-1mm} = \mathbf{w}_i^{(k)} + \hspace*{-2mm} \sum_{j \in \mathcal{N}'^{(k)}_i}{\hspace*{-2mm} \beta_{ij}^{(k)} ( \mathbf{w}_j^{(k)} \hspace{-1mm} - \mathbf{w}_i^{(k)} ) v_{ij}^{(k)}} \hspace{-1mm} - \alpha^{(k)} \mathbf{g}_i^{(k)},
\end{equation}
where $v_{ij}^{(k)}$ indicates whether device $i$ aggregates its model with device $j$ at iteration $k$. Its value depends on $v_i^{(k)}$, which is an indicator of a broadcast event at device $i$ at iteration $k$ defined as

\begin{equation} \label{eqn:v}
	\begin{gathered}
		v_i^{(k)} =
		\begin{cases}
			1 & {\left( 1/n \right)}^\frac12 {\left\| \mathbf{w}_i^{(k)} - \mathbf{\widehat{w}}_i^{(k)} \right\|}_2 > r \rho_i \gamma^{(k)}
			\\
			0 & \text{o.w.}
		\end{cases},
		\\
		v_{ij}^{(k)} =
		\begin{cases}
			\max{\lbrace v_i^{(k)}, v_j^{(k)} \rbrace} & j \in \mathcal{N}_i^{(k)}
			\\
			0 & \text{o.w.}
		\end{cases},
	\end{gathered}
\end{equation}
with $\rho_i = 1 / b_i$. Also, note that the stale model parameters $\hat{\mathbf{w}}_i^{(k)}$ Rearranging the relations in \eqref{eqn:w}, we have
\begin{equation} \label{eqn:wRecursive}
	\begin{aligned}
		\mathbf{w}_i^{(k+1)} = & ( 1 - \sum_{j=1}^m{\beta_{ij}^{(k)} v_{ij}^{(k)}} ) \mathbf{w}_i^{(k)} + \sum_{j=1}^m{\beta_{ij}^{(k)} v_{ij}^{(k)} \mathbf{w}_j^{(k)}}
		\\
		& - \alpha^{(k)} \mathbf{g}_i^{(k)} = \sum_{j=1}^m{p_{ij}^{(k)} \mathbf{w}_j^{(k)}} - \alpha^{(k)} \mathbf{g}_i^{(k)},
	\end{aligned}
\end{equation}
where $p_{ij}^{(k)}$ is the transition weight that device $i$ uses to aggregate device $j$'s parameters at iteration $k$:
\begin{equation} \label{eqn:p}
	p_{ij}^{(k)} =
	\begin{cases}
		\beta_{ij}^{(k)} v_{ij}^{(k)} & \quad i \neq j
		\\
		1 - \sum_{j=1}^m{\beta_{ij}^{(k)} v_{ij}^{(k)}} & \quad i = j
	\end{cases}.
\end{equation}
Note that the aggregation and transition weights, i.e., $\beta_{ij}^{(k)}$ and $p_{ij}^{(k)}$, distinguish the two sources of time variation in the information flow graph of our method: (i) the underlying physical network being time-varying, resulting in varying number of neighbors for each device at each iteration, and (ii) the event-triggering mechanism, adding another overlay time-varying component on top of the network graph.

Next, we collect the parameter vectors of all devices that were previously introduced in matrix form as follows:
$\mathbf{W}^{(k)} = [ \mathbf{w}_1^{(k)} \quad ... \quad \mathbf{w}_m^{(k)} ]^T$, 
$\mathbf{G}^{(k)} = [\mathbf{g}_1^{(k)} \quad ... \quad \mathbf{g}_m^{(k)}]^T$,
$\mathbf{P}^{(k)} = [p_{ij}^{(k)}]_{1\leq i,j\leq m}$.
Now, we transform the recursive update rules of \eqref{eqn:wRecursive} into matrix form to obtain the following relationship
\begin{equation} \label{eqn:wRecursiveMatrix}
	\mathbf{W}^{(k+1)} = \mathbf{P}^{(k)} \mathbf{W}^{(k)} - \alpha^{(k)} \mathbf{G}^{(k)}.
\end{equation}
The recursive expression in \eqref{eqn:wRecursiveMatrix} has been investigated before \cite{nedic2009distributed, sundhar2010distributed}.
However, the existing literature on decentralized stochastic gradient descent does not account for data heterogeneity, and this motivates us to use different analytical tools to derive convergence bounds.

As a conclusion to this section on iterate relations, we introduce two quantities, which will be frequently used in our analysis. We first derive an explicit relationship of \eqref{eqn:wRecursiveMatrix}.
Starting from iteration $s$, where $s \le k$, we have
\begin{equation} \label{eqn:wExplicitMatrix}
	\begin{gathered}
		\mathbf{W}^{(k+1)} \hspace{-1mm} = \mathbf{P}^{( k:s )} \mathbf{W}^{(s)} \hspace{-1mm} - \hspace{-3mm} \sum_{r=s+1}^k{\hspace{-2mm} \alpha^{(r-1)} \mathbf{P}^{( k:r )} \mathbf{G}^{(r-1)}} \hspace{-1mm} - \alpha^{(k)} \mathbf{G}^{(k)},
		\\
		\mathbf{P}^{( k:s )} = \mathbf{P}^{(k)} \mathbf{P}^{(k-1)} \cdots \mathbf{P}^{(s+1)} \mathbf{P}^{(s)}.
	\end{gathered}
\end{equation}

Second, to analyze the consensus of local models, we define the average model as $\mathbf{\bar{w}}^{(k)} = (1/m) \sum_{i=1}^m{\mathbf{w}_i^{(k)}}$. The recursive relation for $\mathbf{\bar{w}^{(k)}}$ using \eqref{eqn:wRecursive} and the stochasticity of $\mathbf{P}^{(k)}$ is
\begin{equation} \label{eqn:wbarRecursive}
	\mathbf{\bar{w}}^{(k+1)} = \mathbf{\bar{w}}^{(k)} - (\alpha^{(k)} / m) \sum_{i=1}^m{\mathbf{g}_i^{(k)}}.
\end{equation}
Also, an explicit relationship between iteration $\mathbf{\bar{w}}^{(k+1)}$ and $\mathbf{\bar{w}}^{(s)}$, where $s \le k$, easily follows from \eqref{eqn:wbarRecursive} as

\begin{equation} \label{eqn:wbarExplicit}
	\mathbf{\bar{w}}^{(k+1)} = \mathbf{\bar{w}}^{(s)} - (1/m) \sum_{r=s}^k{\alpha^{(r)} \sum_{i=1}^m{\mathbf{g}_i^{(r)}}}.
\end{equation}


\section{Convergence Analysis}
\label{sec:conv}
\noindent \textcolor{blue}{In this section, we first detail the assumptions used in our paper (Sec.~\ref{ssec:assumptions}), and then provide the main connectivity result of our event-trigerred approach (Sec.~\ref{ssec:conn}). Finally, we present the lemmas which are used to prove our main Theorems (Sec.~\ref{ssec:lemmas}), and then present the main theoretical contributions themselves (Sec.~\ref{ssec:mainRes}).}

\subsection{\textcolor{blue}{Assumptions}} \label{ssec:assumptions}


\begin{assumption}[Transition weights] \label{assump:weights}
	Let $\{p_{ij}^{(k)}\}$ be the set of aggregation weights in the information graph $\mathcal{G}'(k)$.
	The following conditions must be met:
	\begin{enumerate}[label=(\alph*)]
		\item (Non-negative weights) $\forall i \in \mathcal{M}$, we have
		\begin{enumerate}[label=(\roman*)]
			\item $p_{ii}^{(k)} > 0$ and $p_{ij}^{(k)} > 0$ for all $k \geq 0$ and all neighboring devices $j\in\mathcal{N}'^{(k)}_i$.
			\item $p_{ij}^{(k)} = 0$, if $j\notin\mathcal{N}'^{(k)}_i$.
		\end{enumerate} \label{assump:weights:eta}
		
		\item (Doubly-stochastic weights) The rows and columns of matrix $\mathbf{P}^{(k)} = [p_{ij}^{(k)}]$ are both stochastic, i.e., $\sum_{j=1}^m{p_{ij}^{(k)}} = 1$, $\forall i$, and $\sum_{i=1}^m{p_{ij}^{(k)}} = 1$, $\forall j$. \label{assump:weights:doublystoch}
		
		\item (Symmetric weights) $p_{ij}^{(k)} = p_{ji}^{(k)}$, $\forall i,k$ and $p_{ii}^{(k)} = 1 - \sum_{j \neq i} {p_{ij}^{(k)}}$. \label{assump:weights:symmetric}
	\end{enumerate}
\end{assumption}

Taking into account the conditions mentioned in Assumption~\ref{assump:weights}, and the definition of $p_{ij}^{(k)}$ in~\eqref{eqn:p}, a choice of parameters $\beta_{ij}^{(k)}$ that satisfy these assumptions are as follows
\begin{equation}
	\beta_{ij}^{(k)} = \min{\left \lbrace \frac1{1 + d_i^{(k)}}, \frac1{1 + d_j^{(k)}} \right \rbrace},
\end{equation}
which is inspired by the Metropolis-Hastings algorithm \cite{boyd2004fastest}. Note that $p_{ij}^{(k)}$ also depends on $v_{ij}^{(k)}$, which was defined in~\eqref{eqn:v}.

\begin{assumption} [Smoothness, Strong convexity, and Data heterogeneity] \label{assump:smooth_convex_graddiv}
		The local objective function at each device $i \in \mathcal{M}$, i.e., $F_i$, satisfies the following
	\begin{enumerate}[label=(\alph*)]
		\item $L_i$-Lipschitz continuous gradients:
		
		$\left\| \nabla F_i{\left( \mathbf{w} \right)} - \nabla F_i{\left( \mathbf{w}' \right)} \right\| \le L_i \left\| \mathbf{w} - \mathbf{w}' \right\|,$ \label{assump:smoothness}
		
		\item $\mu_i$-strong convexity:
		
		$\left\langle \nabla{F_i{\left( \mathbf{w} \right)}} - \nabla{F_i{\left( \mathbf{w}' \right)}}, \mathbf{w} - \mathbf{w}' \right\rangle \geq \mu_i {\left\| \mathbf{w} - \mathbf{w}' \right\|}^2,$ \label{assump:convexity}
		
		\item The data heterogeneity across the devices is measured via $\delta_i > 0$ as
		
		$\left\| \nabla F_i{\left( \mathbf{w} \right)} - \nabla F{\left( \mathbf{w} \right)} \right\| \le \delta_i,$ \label{assump:graddiversity}
	\end{enumerate}
	$\forall ( \mathbf{{w}'}, \mathbf{w} ) \in \mathbb{R}^n \times \mathbb{R}^n$, where we also define $L = \max_{i \in \mathcal{M}}{L_i}$, $\mu = \min_{i \in \mathcal{M}}{\mu_i}$ and $\delta = \max_{i \in \mathcal{M}}{\delta_i}$.
\end{assumption}

Note that the global objective function $F{( \mathbf{w} )}$, which is a convex combination of local objective functions, will also be strongly convex, thus having a unique minimizer, denoted by $\mathbf{w}^\star = \argmin_{\mathbf{w} \in \mathbb{R}^n}{F{( \mathbf{w} )}}$. Additionally, following Assumptions~\ref{assump:smooth_convex_graddiv}-\ref{assump:smoothness} and \ref{assump:smooth_convex_graddiv}-\ref{assump:convexity}, we have $\mu \le \mu_i \le L_i \le L$, for all $i \in \mathcal{M}$. Finally, note that the assumption about data heterogeneity (Assumption \ref{assump:smooth_convex_graddiv}-\ref{assump:graddiversity}) was inspired by \cite{wang2019adaptive, lin2021semi}.

Works like \cite{nedic2009distributed, sundhar2010distributed} make a much stricter assumption than the statistical heterogeneity assumption we make in our paper: the bounded (sub)-gradients assumption, i.e., $\| \nabla{F}_i(\mathbf{w}) \| \le L_i$. The reason our Assumption \ref{assump:smooth_convex_graddiv}-\ref{assump:graddiversity} is less strict is that by letting $\mathbf{w} = \mathbf{w}^\star$, we get that local gradients should be bounded only at the optimal point, contrary to those other papers which make the assumption that they should be bounded over the full space. The implication is that the data heterogeneity assumption in our paper only requires the difference between the global gradient at a certain point and the local gradient at that same point to be upper bounded by a constant scalar. This means the norm of local and global gradients can have arbitrarily large values in our paper (unlike the bounded gradients assumption of \cite{nedic2009distributed, sundhar2010distributed}), so long as their difference is bounded for every point in $\mathbb{R}^n$.

\begin{assumption}[Gradient approximation errors] \label{assump:graderror}
	We make the following assumptions on the gradient approximation errors $\mathbf{\epsilon}_i^{(k)}$ for all $i \in \mathcal{M}$ and all $k \geq 0$:
	\begin{enumerate}[label=(\alph*)]
		\item Zero mean, i.e., $\mathbb{E}{[ \mathbf{\epsilon}_i^{(k)} ]} = 0 $ \label{assump:graderror:mean}.
		
		\item Bounded mean square, i.e., there is a scalar $\sigma_i^2$ such that $\mathbb{E}{[ {\| \mathbf{\epsilon}_i^{(k)} \|}_2^2 ]} \le \sigma_i^2 \le \sigma^2$, where $\sigma = \max_{i \in \mathcal{M}}{\sigma_i}$. \label{assump:graderror:variance}
		
		\item Each random vector $\epsilon_i^{(k)}$ is independent from $\epsilon_j^{(k)}$ for $j\neq i$. \label{assump:graderror:indep}
	\end{enumerate}
\end{assumption}

\begin{assumption}[Step sizes] \label{assump:stepsizes}
	All devices use the same step size for model training. We study the behavior of our algorithm under two policies for the step size:
	\begin{enumerate}[label=(\alph*)]
		\item Constant step size, having $\alpha^{(k)} = \alpha$ where $\alpha > 0$. \label{assump:stepsizes:constant}
		
		\item Diminishing step size, in which the step size decays over time, satisfying the following conditions
		\begin{equation*}
			\lim_{k \to \infty}{\alpha^{(k)}} = 0, \quad \sum_{k=0}^\infty{\alpha^{(k)}} = \infty, \quad \sum_{k=0}^\infty{{\left( \alpha^{(k)} \right)}}^2 < \infty.
		\end{equation*} \label{assump:stepsizes:diminishing}
	\end{enumerate}
\end{assumption}
In particular, setting $\alpha^{(k)} = \alpha^{(0)} / {( 1+ k / \eta )}^{\theta}$ meets the criteria of Assumption~\ref{assump:stepsizes}-\ref{assump:stepsizes:diminishing} for $\alpha^{(0)},\eta > 0$, and $\theta \in (0.5, 1]$.

The previous assumptions are common in the literature~\cite{wang2019adaptive,hosseinalipour2022multi}. In the next assumption, we introduce a relaxed version of graph connectivity requirements relative to existing work in distributed learning, which underscores the difference of our decentralized event-triggered FL method compared with traditional distributed optimization algorithms.

\begin{assumption}[Network graph connectivity]
	\label{assump:conn}
	The underlying communication graph satisfies the following properties:
	\begin{enumerate}[label=(\alph*)]
		\item There exists an integer $B_1 \geq 1$ such that the graph union of the physical network graph $\mathcal{G}^{(k)} = ( \mathcal{M}, \mathcal{E}^{(k)} )$ from any arbitrary iteration $k$ to $k + B_1 - 1$, i.e., $\mathcal{G}^{( k : k+B_1-1 )} = {( \mathcal{M}, \cup_{s=0}^{B_1 - 1}{\mathcal{E}^{( k+s )}} )}$, is connected for any $k \geq 0$. \label{assump:conn:physicalconn}
		
		\item There exists an integer $B_2 \geq 1$ such that for every device $i$, triggering conditions for the broadcast event occur at least once every $B_2$ consecutive iterations $\forall k \geq 0$. This is equivalent to the following condition:
		\begin{equation*}
			\exists B_2\geq 1, \forall i:	\max{\lbrace v_i^{(k)},  v_i^{(k+1)}, \cdots, v_i^{( k+B_2-1 )} \rbrace} = 1.
		\end{equation*} \label{assump:conn:boundedintercom}
	\end{enumerate}
\end{assumption}
\textcolor{blue}{Existing works in the literature either assume (i) a static physical graph ($B_1 = 1$) with sporadic communications \cite{koloskova2020unified}, or (ii) a time-varying physical graph with communications at every round ($B_2 = 1$) \cite{nedic2009distributed}. Our paper is the first to combine these two sources of time-variation (physical graph and the communication graph) for event-triggered decentralized learning methods, and provide a generalized theoretical analysis when neither of them are static.} We will use Assumption~\ref{assump:conn} in the proof of Proposition~\ref{proposition:conn} to analyze the connectivity behavior of information flow graphs, i.e., $\mathcal{G}'^{(k)}$.

\subsection{\textcolor{blue}{Main Connectivity Result}} \label{ssec:conn}

in Proposition \ref{proposition:conn} below, we provide a connectivity guarantee of the information flow graph in the presence of (i) the underlying physical network connecting the agents being time-varying and (ii) the event-triggering communication mechanism adding another layer of temporal variation for inter-device communication, i.e., as in Assumption \ref{assump:conn}. In prior works like \cite{nedic2017achieving}, only the underlying physical graph is time-varying, and consensus operations are carried out whenever possible, i.e., at every iteration. As a result, obtaining an information graph connectivity guarantee has not traditionally been a key challenge. In contrast, in our paper, consensus operations do not occur at every iteration due to the unique heterogeneity challenges in decentralized federated learning discussed in Sec.~\ref{sec:intro}, especially due to inter-device links having varying bandwidth availability. Thus, even if the underlying physical graph is connected, more elaborate considerations are required to establish connectivity results on the information flow graphs.

Additionally, note that papers like \cite{george2020distributed}, which focus on event-triggered methods, consider a simplified static underlying physical graph. Thus, the analysis in both \cite{nedic2017achieving, george2020distributed} contains only one of the two connectivity criteria we consider in our analysis.

\begin{proposition} \label{proposition:conn}
	Let Assumption~\ref{assump:conn} hold. Under the {\tt EF-HC} algorithm (Alg.~\ref{alg:efhc}), the information flow graph $\mathcal{G}'^{(k)}$ is $B$-connected, i.e., $\mathcal{G}'^{( k : k+B-1 )} = {( \mathcal{M}, \cup_{s=0}^{B-1}{\mathcal{E}'^{( k+s )}} )}$ is connected for any $k \geq 0$, where $B = ( \tilde{l}+2 ) B_1$ and $\tilde{l}$ are determined via $\tilde{l} B_1 \le B_2 \le ( \tilde{l}+1 ) B_1 - 1$. Note that $B_1$ and $B_2$ are, respectively, the connectivity bound of the physical network graph and the bound for the occurrence of communication events of Assumption~\ref{assump:conn}-\ref{assump:conn:physicalconn} and \ref{assump:conn}-\ref{assump:conn:boundedintercom}.
	
	\begin{proof}
			See Appendix~A in the supplementary material. The high-level idea behind the proof is (i) carefully keeping track of all devices that a certain agent has communicated with at every iteration, and then (ii) finding an upper bound on the number of iterations until all devices have communicated with at least one of their neighbors.
	\end{proof}
\end{proposition}

It is important to note that we use the $B$ parameter introduced in Proposition~\ref{proposition:conn} only for convergence analysis. It can have an arbitrarily large value. Therefore, we are not making strict connectivity assumptions on the underlying graph.\footnote{If our algorithms required the devices to exchange parameters with their neighbors upon disconnection, we would have $B = \max{\lbrace B_1, B_2 \rbrace}$.}

\subsection{\textcolor{blue}{Intermediate Lemmas for Convergence}} \label{ssec:lemmas}
In this section, we provide some lemmas which are useful in the proofs of Theorems~\ref{theorem:constant} and~\ref{theorem:diminishing} of Sec.~\ref{ssec:mainRes}. These lemmas also provide additional characteristics of our methodology.

Our first lemma gives a bound on the consensus error over the course of multiple iterations, i.e., $\| \mathbf{P}^{(k:s)} \mathbf{W}^{(k)} - \mathbf{1}_m \mathbf{\bar{w}}^{(k)} \|$, using the spectral norm of $\mathbf{P}^{(k:s)} - (1/m) \mathbf{1}_m \mathbf{1}_m^T$, which we show that depending on iteration $s$, this bound can be made tighter.
\begin{lemma} \label{lemma:harnessing:spectral}
	Let Assumption~\ref{assump:weights} hold, and let $B$ be the connectivity bound of Proposition~\ref{proposition:conn}. Then the following is true
	\begin{enumerate}[label=(\alph*)]
		\item From iteration $k$ to $k+B-r$, where $r = 2, \cdots, B$, we have
		\begin{equation*}
			\begin{aligned}
				& \left\| \mathbf{P}^{(k+B-r:k)} \mathbf{W}^{(k)} - \mathbf{1}_m \mathbf{\bar{w}}^{(k)} \right\|
				\\
				& \le \rho^{(k+B-r:k)} \left\| \mathbf{W}^{(k)} - \mathbf{1}_m \mathbf{\bar{w}}^{(k)} \right\|
                \\
                & \le \left\| \mathbf{W}^{(k)} - \mathbf{1}_m \mathbf{\bar{w}}^{(k)} \right\|.
			\end{aligned}
		\end{equation*}
		
		\item From iteration $k$ to $k+B-1$, we have the following
		\begin{equation*}
			\begin{aligned}
				& \left\| \mathbf{P}^{(k+B-1:k)} \mathbf{W}^{(k)} - \mathbf{1}_m \mathbf{\bar{w}}^{(k)} \right\|
				\\
				& \le \rho^{(k+B-1:k)} \left\| \mathbf{W}^{(k)} - \mathbf{1}_m \mathbf{\bar{w}}^{(k)} \right\|
                \\
                & \le \rho \left\| \mathbf{W}^{(k)} - \mathbf{1}_m \mathbf{\bar{w}}^{(k)} \right\|,
			\end{aligned}
		\end{equation*}
		in which $\rho^{(k+B-r:k)} = \rho{\left( \mathbf{P}^{(k+B-r:k)} - \frac1m \mathbf{1}_m \mathbf{1}_m^T \right)}$, $\rho = \sup_{k=0,1,\cdots}{\rho{\left( \mathbf{P}^{(k+B-1:k)} - \frac1m \mathbf{1}_m \mathbf{1}_m^T \right)}}$, and $0 < \rho < 1$.
	\end{enumerate}
	
	\begin{proof}
		Since the graph is time-varying, we can only guarantee the connectivity of $\mathbf{P}^{(k+B-1:k)}$. Therefore, $0 < \rho{\left( \mathbf{P}^{(k+B-r:k)} - \frac1m \mathbf{1}_m \mathbf{1}_m^T \right)} \le 1$ for all $r = 2, \cdots, B$, but $0 < \rho < 1$. The rest of the proof follows from Sec.~II-B of \cite{qu2017harnessing}.
	\end{proof}
\end{lemma}
Lemma~\ref{lemma:harnessing:spectral} is essential in the analysis of our method and helps us to prove the convergence under time-varying communication graphs with an arbitrary connectivity bound.

\begin{definition}
	We define the following gradient matrices: $\mathbf{\nabla}^{(k)} = [ \nabla{F_1{( \mathbf{w}_1^{(k)} )}}, \cdots, \nabla{F_m{( \mathbf{w}_m^{(k)} )}} ]^T$, $\mathbf{\bar{\nabla}}^{(k)} = \frac1m \mathbf{1}_m^T \mathbf{\nabla}^{(k)}$ and $\mathbf{\nabla}{F}^{(k)} = [ \nabla{F{( \mathbf{w}_1^{(k)} )}}, \cdots, \nabla{F{( \mathbf{w}_m^{(k)} )}} ]^T$. Furthermore, note that $\nabla{F}{( \mathbf{\bar{w}}^{(k)} )} \in \mathbb{R}^{1 \times n}$, is the gradient of global objective function evaluated at $\mathbf{\bar{w}}^{(k)}$.
\end{definition}

Using the previous definition, we next provide two inequalities which help us bound the expressions involving the gradients in any iteration $k$, via the values of the model parameters, scaled by a constant factor.
\begin{lemma} \label{lemma:harnessing:ineqs}
	Under Assumptions~\ref{assump:smooth_convex_graddiv}-\ref{assump:smoothness} and \ref{assump:smooth_convex_graddiv}-\ref{assump:convexity}, the following holds for all $k \geq 0$:
	\begin{equation*}
		\left\| \nabla{F{\left( \mathbf{\bar{w}}^{(k)} \right)}} - \mathbf{\bar{\nabla}}^{(k)} \right\| \le \frac{L}{\sqrt{m}} \left\| \mathbf{W}^{(k)} - \mathbf{1}_m \mathbf{\bar{w}}^{(k)} \right\|.
	\end{equation*}
	Also, if $\alpha^{(k)} < \frac2{\mu + L}$, then
	\begin{equation*}
		\Big\| \mathbf{\bar{w}}^{(k)} - \alpha^{(k)} \nabla{F{\left( \mathbf{\bar{w}}^{(k)} \right)}} - \mathbf{w}^\star \Big\| \le \left( 1 - \mu \alpha^{(k)} \right) \left\| \mathbf{\bar{w}}^{(k)} - \mathbf{w}^\star \right\|.
	\end{equation*}
    \begin{proof}
        Follows from Lemmas~8-(c) and 10 of \cite{qu2017harnessing}.
    \end{proof}
\end{lemma}

Next, we obtain the following bounds for the average of gradient approximation errors, which are used to obtain the results of several subsequent lemmas.
\begin{lemma} \label{lemma:epsilonbounds}
	Let Assumption~\ref{assump:graderror} hold. Provided the definitions $\mathbf{\bar{\epsilon}}^{(k)} = \frac1m \sum_{i=1}^m{\mathbf{\epsilon}_i^{(k)}}$ and $\mathbf{\epsilon}^{(k)} =
	{\begin{bmatrix}
			\mathbf{\epsilon}_1^{(k)}, \cdots, \mathbf{\epsilon}_m^{(k)}
	\end{bmatrix}}^T$, we have
	\begin{equation*}
		\begin{gathered}
			\mathbb{E}{\left[ {\left\| \mathbf{\bar{\epsilon}}^{(k)} \right\|}^2 \right]} \le \frac{\sigma^2}{m}, \quad \mathbb{E}{\left[ {\left\| \mathbf{\epsilon}^{(k)} - \mathbf{1}_m \mathbf{\bar{\epsilon}}^{(k)} \right\|}^2 \right]} \le m \sigma^2.
		\end{gathered}
	\end{equation*}
	
	\begin{proof}
		The first inequality follows from Lemma~2 of \cite{pu2021distributed}. For the proof of the second bound, see Appendix~C in the supplementary material.
	\end{proof}
\end{lemma}

Traditional analysis of distributed gradient descent involves making the assumption of bounded gradient (see \cite{nedic2009distributed}, \cite{nedic2010constrained}). However, since we have replaced such an assumption with two different but more general assumptions, namely smoothness and data heterogeneity (Assumptions~\ref{assump:smooth_convex_graddiv}-\ref{assump:smoothness} and \ref{assump:smooth_convex_graddiv}-\ref{assump:graddiversity}), our analysis is different compared to the current literature. Inspired by the gradient tracking literature in distributed learning \cite{qu2017harnessing, pu2021distributed}, in the following lemma, we look at the behavior of ${\| \mathbf{W}^{(k+1)} - \mathbf{1}_m \mathbf{\bar{w}}^{(k+1)} \|}^2$ and ${\| \mathbf{\bar{w}}^{(k+1)} - \mathbf{w}^\star \|}^2$, and bound them simultaneously via a system of inequalities.

\begin{lemma} \label{lemma:kplus1}
	Assumptions~\ref{assump:smooth_convex_graddiv} and \ref{assump:graderror} yield the following bounds:
	\begin{enumerate} [label=(\alph*)]
		\item Consensus error on local gradients:
		\begin{equation*}
			{\left\| \mathbf{\nabla}^{(k)} - \mathbf{1}_m \mathbf{\bar{\nabla}}^{(k)} \right\|}^2 \le 2 m \delta^2 + 8 L^2 {\left\| \mathbf{W}^{(k)} - \mathbf{1}_m \mathbf{\bar{w}}^{(k)} \right\|}^2.
		\end{equation*} \label{lemma:kplus1:gradbound}
		
		\item Optimization error, assuming $\alpha^{(k)} < \frac2{\mu + L}$:
		\begin{equation*}
			\begin{aligned}
				\mathbb{E} & {\left[ {\left\| \mathbf{\bar{w}}^{(k+1)} - \mathbf{w}^\star \right\|}^2 \right]} \le a_{11}^{(k)} \mathbb{E}{\left[ {\left\| \mathbf{\bar{w}}^{(k)} - \mathbf{w}^\star \right\|}^2 \right]}
				\\
				& + a_{12}^{(k)} \mathbb{E}{\left[ {\left\| \mathbf{W}^{(k)} - \mathbf{1}_m \mathbf{\bar{w}}^{(k)} \right\|}^2 \right]} + c_1^{(k)},
			\end{aligned}
		\end{equation*}
		where $a_{11}^{(k)} = 1 - \mu \alpha^{(k)}$, $a_{12}^{(k)} = ( 1 + \mu \alpha^{(k)} ) \alpha^{(k)} L^2 / (\mu m)$, $c_1^{(k)} = ( \alpha^{(k)} )^2 \sigma^2 / m$. \label{lemma:kplus1:optim}
		
		\item Let Assumption~\ref{assump:weights} also hold. The Expected consensus error of model weights is bounded as:
		\begin{equation*}
			\begin{aligned}
				\mathbb{E} & {\left[ {\left\| \mathbf{W}^{(k+1)} - \mathbf{1}_m \mathbf{\bar{w}}^{(k+1)} \right\|}^2 \right]} \le
				\\
				& a_{22}^{(k)} \mathbb{E}{\left[ {\left\| \mathbf{W}^{(k)} - \mathbf{1}_m \mathbf{\bar{w}}^{(k)} \right\|}^2 \right]} + c_2^{(k)},
			\end{aligned}
		\end{equation*}
		where we define $a_{21}^{(k)} = 0$, and obtain $a_{22}^{(k)} = {( 1 + 2\sqrt{2} \alpha^{(k)} L )}^2$ and $c_2^{(k)} = m {( \alpha^{(k)} )}^2 ( 2 (1 + 2\sqrt{2} \alpha^{(k)} L) \delta^2 / (2\sqrt{2} \alpha^{(k)} L) + \sigma^2 )$. \label{lemma:kplus1:consensus}
	\end{enumerate}
	\begin{proof}
		See Appendix~D in the supplementary material.
	\end{proof}
\end{lemma}
We make two observations from the above lemma. First, consider the term $2 m \delta^2$ in Lemma \ref{lemma:kplus1}-\ref{lemma:kplus1:gradbound}. This term reveals that even if consensus is reached among the devices, i.e., ${\| \mathbf{W}^{(k)} - \mathbf{1}_m \mathbf{\bar{w}}^{(k)} \|}^2 = 0$, the local gradients would always be different from each other due to the data heterogeneity assumption of \ref{assump:smooth_convex_graddiv}-\ref{assump:graddiversity}. Second, the system of inequalities is semi-coupled, as we can bound ${\| \mathbf{W}^{(k+1)} - \mathbf{1}_m \mathbf{\bar{w}}^{(k)} \|}^2$ at each iteration only by its own value at the previous iterations.

Next, we give the the following definition of the parameters to be analyzed:
	\begin{definition}
		The optimization error, i.e., the distance between the average model and the optimal solution, is defined as $\| \bar{\mathbf{w}}^{(k)} - \mathbf{w}^\star \|^2$ at iteration $k$. Also, the consensus error, i.e., the distance between model parameters of all devices $i \in \mathcal{M}$ with the average model, is defined as $\| \mathbf{W}^{(k)} - \mathbf{1}_m \bar{\mathbf{w}}^{(k)} \|^2$ at iteration $k$. We also define the following vector as the expected value of these error terms:
		\begin{equation}
			\mathbf{\Xi}^{(k)} =
			\begin{bmatrix}
				\mathbb{E}{\left[ {\left\| \bar{\mathbf{w}}^{(k)} - \mathbf{w}^\star \right\|}^2 \right]}
				\\
				\mathbb{E}{\left[ {\left\| \mathbf{W}^{(k)} - \mathbf{1}_m \bar{\mathbf{w}}^{(k)} \right\|}^2 \right]}
			\end{bmatrix}.
		\end{equation}
\end{definition}
Using this definition, we write the iterate relations defined in parts~\ref{lemma:kplus1:consensus}\&\ref{lemma:kplus1:optim} of Lemma~\ref{lemma:kplus1} as a system of recursive inequalities:
\begin{equation} \label{eqn:AkCkRecursive}
		\mathbf{\Xi}^{(k+1)} \le \mathbf{A}^{(k)} \mathbf{\Xi}^{(k)} + \mathbf{C}^{(k)},
\end{equation}
in which $\mathbf{A}^{(k)} = {[a_{ij}^{(k)}]}_{1 \le i,j \le 2}$ and $\mathbf{C}^{(k)} = {[c_1^{(k)}, c_2^{(k)}]}^T$, and the $a_{ij}^{(k)}$ and $c_i^{(k)}$ values were defined in Lemmas~\ref{lemma:kplus1}-\ref{lemma:kplus1:optim} and \ref{lemma:kplus1}-\ref{lemma:kplus1:consensus} for $1 \le i, j \le 2$.

Next, we derive an explicit relation for the system of inequalities of \eqref{eqn:AkCkRecursive}, starting from an arbitrary iteration $s$ as

\begin{equation} \label{eqn:AkCkExplicit}
    \mathbf{\Xi}^{(k+1)} \le \mathbf{A}^{(k:s)} \mathbf{\Xi}^{(s)} + \sum_{r=s+1}^{k}{\mathbf{A}^{(k:r)} \mathbf{C}^{(r-1)}} + \mathbf{C}^{(k)},
\end{equation}
in which $\mathbf{A}^{(k:s)} = \mathbf{A}^{(k)} \cdots \mathbf{A}^{(s)}$. Since $a_{21}^{(k)} = 0$, we can easily compute the following entries of $\mathbf{A}^{(k:s)}$ which will be frequently used in our analysis. We have
\begin{equation} \label{eqn:a_ks_Vals}
	\begin{aligned}
		& a_{11}^{(k:s)} = a_{11}^{(k)} \cdots a_{11}^{(s)}, a_{21}^{(k:s)} = 0, a_{22}^{(k:s)} = a_{22}^{(k)}. \cdots a_{22}^{(s)}.
	\end{aligned}
\end{equation}

As mentioned before Lemma~\ref{lemma:kplus1}, similar analysis to our paper is common in the gradient tracking literature. But current research has only shown convergence guarantees over static communication graphs. Next, we move on to an important lemma of our paper, which is the key to proving the convergence for time-varying graphs with arbitrary connectivity $B$.
\begin{lemma} \label{lemma:kplusB}
	Let Assumptions~\ref{assump:weights}-\ref{assump:graderror} hold. Then, using Lemma~\ref{lemma:kplus1}, we can get the following inequality on the expected consensus error of the model weights at iteration $k+B$ for any $k \geq 0$:
	\begin{equation*}
		\begin{aligned}
			& \mathbb{E}{\left[ {\left\| \mathbf{W}^{(k+B)} - \mathbf{1}_m \mathbf{\bar{w}}^{(k+B)} \right\|}^2 \right]}
			\\
			& \begin{aligned}
				\le & \phi_{22}^{(k)} \mathbb{E}{\left[ {\left\| \mathbf{W}^{(k)} - \mathbf{1}_m \mathbf{\bar{w}}^{(k)} \right\|}^2 \right]} + \psi_2^{(k)},
			\end{aligned}
		\end{aligned}
	\end{equation*}
	in which we have defined $\phi_{21}^{(k)} = 0$, and obtained $\phi_{22}^{(k)} = \frac{1+\rho^2}2 + 16 \frac{B L^2}{1-\rho^2} \sum_{r=k+1}^{k+B} {( \alpha^{(r-1)} )}^2 a_{22}^{(r-2:k)}$ and $\psi_2^{(k)} = 2B \sum_{r=kB+1}^{( k+1 ) B} {( \alpha^{(r-1)} )}^2 \lbrace \frac2{1-\rho^2} [ m \delta^2 + 4 L^2 ( \sum_{l=kB+1}^{r-2}{a_{22}^{(r-2:l)} c_2^{(l-1)}} + c_2^{(r-2)} ) ] + m \sigma^2 \rbrace$. Further note that $0 < \rho^{(k+B-1:k)} \le \rho < 1$ for any $k \geq 0$, and $B$ is the connectivity bound of Proposition~\ref{proposition:conn}.
	
	\begin{proof}
		See Appendix~E in the supplementary material.
	\end{proof}
\end{lemma}

We next derive a system of inequalities for the iterate relations of Lemmas~\ref{lemma:kplus1}-\ref{lemma:kplus1:consensus} and \ref{lemma:kplus1}-\ref{lemma:kplus1:optim}, but instead of writing a recursive relation between iteration $k+1$ and $k$ as in \eqref{eqn:AkCkRecursive}, we use Lemma~\ref{lemma:kplusB} to obtain a recursive relation between $k+B$ and $k$ as
\begin{equation} \label{eqn:PhikPsikRecursive}
		\mathbf{\Xi}^{((k+1) B)} \le \mathbf{\Phi}^{(k)} \mathbf{\Xi}^{(kB)} + \mathbf{\Psi}^{(k)},
\end{equation}
in which $\mathbf{\Phi}^{(k)} = {[\phi_{ij}^{(k)}]}_{1 \le i,j \le 2}$, and $\mathbf{\Psi}^{(k)} = {[\psi_1^{(k)}, \psi_2^{(k)}]}^T$. Next, note that while $\phi_{21}^{(k)}$, $\phi_{22}^{(k)}$ and $\psi_2^{(k)}$ come from Lemma~\ref{lemma:kplusB}, the remaining entries of these matrices are obtained as
\begin{equation} \label{eqn:phiValues}
	\phi_{11}^{(k)} = a_{11}^{(\left( k+1 \right) B-1 : k B)}, \quad \phi_{12}^{(k)} = a_{12}^{(\left( k+1 \right) B-1 : k B)},
\end{equation}
\begin{equation} \label{eqn:psiValues}
	\begin{aligned}
		& \begin{aligned}
			& \psi_1^{(k)} = \sum_{r=kB+1}^{\left( k+1 \right) B-1}{\left( \alpha^{(r-1)} \right)}^2 \Bigg[ a_{11}^{(\left( k+1 \right) B-1:r)} \frac{\sigma^2}m
			\\
			& + 2m a_{12}^{(\left( k+1 \right) B-1:r)} \left( \frac{1 + 2\sqrt{2} \alpha^{(r-1)} L}{2\sqrt{2} \alpha^{(r-1)} L} \delta^2 + 2 \sigma^2 \right) \Bigg]
			\\
			& + \frac{{\left( \alpha^{(\left( k+1 \right) B-1)} \right)}^2 \sigma^2}m.
		\end{aligned} \hspace*{-10mm}
	\end{aligned}
\end{equation}
Comparing \eqref{eqn:AkCkExplicit} and \eqref{eqn:PhikPsikRecursive}, note that we have used $\mathbf{\Phi}^{(k)} = \mathbf{A}^{(( k+1 )B-1 : kB)}$ and $\mathbf{\Psi}^{(k)} = \sum_{r=kB+1}^{( k+1 )B-1}{\mathbf{A}^{(( k+1 )B-1 : r)} \mathbf{C}^{(r-1)}} + \mathbf{C}^{(( k+1 )B-1)}$, except for two modifications, where we have replaced $\phi_{22}^{(k)}$ and $\psi_{2}^{(k)}$ with the values derived in Lemma~\ref{lemma:kplusB}.

Next, we derive an explicit equation for the system of inequalities of \eqref{eqn:PhikPsikRecursive}, starting from an arbitrary iteration $s$, as follows
\begin{equation} \label{eqn:PhikPsikExplicit}
		\mathbf{\Xi}^{((k+1) B)} \le \mathbf{\Phi}^{(k:s)} \mathbf{\Xi}^{(sB)} + \sum_{r=s+1}^{k}{\mathbf{\Phi}^{(k:r)} \mathbf{\Psi}^{(r-1)}} + \mathbf{\Psi}^{(k)},
\end{equation}
in which $\mathbf{\Phi}^{(k:s)} = \mathbf{\Phi}^{(k)} \cdots \mathbf{\Phi}^{(s)}$. Since $\phi_{21}^{(k)} = 0$ (see Lemma~\ref{lemma:kplusB}), we can easily compute the following entries of $\mathbf{\Phi}^{(k:s)}$ that will be used in our analysis
\begin{equation} \label{eqn:phi_ks_Vals}
	\phi_{11}^{(k:s)} = \phi_{11}^{(k)} \cdots \phi_{11}^{(s)}, \phi_{21}^{(k:s)} = 0, \qquad \phi_{22}^{(k:s)} = \phi_{22}^{(k)}. \cdots \phi_{22}^{(s)}.
\end{equation}

In the subsequent proposition, we build on the results of \eqref{eqn:PhikPsikExplicit}, and obtain the conditions under which the spectral norm of $\mathbf{\Phi}^{(k)}$ would be less than one. Then, we use that to bound the system of inequalities of \eqref{eqn:PhikPsikExplicit}.
\begin{proposition} \label{proposition:nonIncreasing}
	Let Assumptions~\ref{assump:weights}-\ref{assump:graderror} and \ref{assump:conn} hold, and a non-increasing step size be used such that $\alpha^{(k+1)} \le \alpha^{(k)}$ for all $k\geq0$ \footnote{This satisfies the step size policies of both Assumption~\ref{assump:stepsizes}-\ref{assump:stepsizes:constant} and \ref{assump:stepsizes}-\ref{assump:stepsizes:diminishing}.}. \textcolor{blue}{Using the definitions of $\mathbf{\Phi}^{(k:s)}$ and $\mathbf{\Psi}^{(k:s)}$ in \eqref{eqn:PhikPsikExplicit}, if the step size satisfies $\alpha^{(0)} \le (1-\rho^2) / (8 B L {( 1+\Gamma_1 )}^{B-1})$ where $\Gamma_1 = (1-\rho^2) / (2 \sqrt{2} B)$, then $\rho(\mathbf{\Phi}^{(k)}) < 1$ and the following bound holds}
	\begin{equation}
		\begin{aligned}
			\mathbf{\Xi}^{(kB)} \le \mathcal{O} & {\left( \phi_{11}^{(k-1:K)} \right)} \mathcal{O}{\left( \phi_{22}^{(K-1:0)} \right)} \mathbf{\Xi}^{(0)}
			\\
			& + \mathcal{O}{\left( \phi_{11}^{(k-1:K)} \right)} \sum_{r=1}^{K-1}{\mathcal{O}{\left( \phi_{22}^{(K-1:r)} \right)} \mathbf{\Psi}^{(r-1)}}
			\\
			& + \sum_{r=K}^{k-1}{\mathcal{O}{\left( \phi_{11}^{(k-1:r)} \right)} \mathbf{\Psi}^{(r-1)}} + \mathbf{\Psi}^{(k-1)},
		\end{aligned}
	\end{equation}
	where iteration $K$ is determined by $\phi_{11}^{(k)} < \phi_{22}^{(k)}$ for all $k=0,\cdots,K-1$, and $\phi_{11}^{(k)} \geq \phi_{22}^{(k)}$ for $k \geq K$.
	
	Furthermore, the matrix $\mathbf{\Psi}^{(k)}$ can be bounded as
	\begin{equation} \label{eqn:matEntriesPsiDiminishing}
		\mathbf{\Psi}^{(k)} \le {\left( \alpha^{(kB)} \right)}^2 B
		\begin{bmatrix}
			\frac{\sigma^2}m + \nu^{(kB)}
			\\
			2m B \left\lbrace 2 \frac{\delta^2 + 2 \mu \alpha^{(kB)} \nu^{(kB)}}{1-\rho^2} + \sigma^2 \right\rbrace
		\end{bmatrix},
	\end{equation}
	where $\nu^{(k)} = (2/\mu) ( B-1 ) L^2 {( 1+\Gamma_1 )}^{2 ( B-1 )} ( (1+\Gamma_1) \delta^2 / (\sqrt{2}L) + \alpha^{(k)} \sigma^2 )$.
	
	\begin{proof}
		See Appendix~F in the supplementary material.
	\end{proof}
\end{proposition}
Proposition \ref{proposition:nonIncreasing} lays out the constraints on the step size that have to be satisfied in order for the spectral radius of the state transition matrix, i.e., $\mathbf{\Phi}^{(k)}$ to be less than $1$. Our novel analytical approach provided in Appendix~F helps us prove that the spectral radius of the aforementioned matrix is less than $1$ despite looking at its product over $B$ iterations.
Furthermore, it implies that if the spectral norm of $\mathbf{\Phi}^{(k)}$ satisfies $\rho{( \mathbf{\Phi}^{(k)} )} = \max{\lbrace \phi_{11}^{(k)}, \phi_{22}^{(k)} \rbrace} < 1$, then we can bound the system of inequalities in terms of the spectral norm. Moreover, the matrix $\mathbf{\Phi}^{(k-1:0)}$ is separated into the product of two terms $\mathbf{\Phi}^{(k-1:K)} \mathbf{\Phi}^{(K-1:0)}$, and this is done because we have $\rho{( \mathbf{\Phi}^{(K-1:0)} )} = \phi_{22}^{(K-1:0)}$ and $\rho{( \mathbf{\Phi}^{(k-1:K-1)} )} = \phi_{11}^{(k-1:K-1)}$ (see Appendix~F for more discussion). This lemma is used directly to prove Theorems~\ref{theorem:constant} and \ref{theorem:diminishing}.

Next, we analyze the dependence of the constraint on $\alpha^{(0)}$ from Proposition~\ref{proposition:nonIncreasing} on $B$, which is the connectivity parameter of Proposition~\ref{proposition:conn}. We have
\begin{equation*}
	\begin{aligned}
		\alpha^{(0)} & \le \mathcal{O}{\left( 1 / \left[ B {\left( 1 + \Gamma_1 \right)}^{B-1} \right] \right)}
		\\
		& = \mathcal{O}{\left( 1 / \left[ B {\left( 1+ (1-\rho^2) / (2 \sqrt{2} B) \right)}^{B-1} \right] \right)}
		\\
		& = \mathcal{O}{\left( 1 / \left[ B {\left( 1 + 1 / B \right)}^B \right] \right)} = \mathcal{O}{\left( 1/B \right)}.
	\end{aligned}
\end{equation*}
We can see that, as expected, the constraint on $\alpha^{(0)}$ is inversely proportional to $B$, meaning that more frequent communications over a well-connected graph (lower $B$) allows us to choose larger step sizes.

\subsection{\textcolor{blue}{Main Convergence Results}} \label{ssec:mainRes}


We present our most central results, which obtain the convergence characteristics of {\tt EF-HC} under the step size policies of Assumption~\ref{assump:stepsizes} in Theorems~\ref{theorem:constant} and \ref{theorem:diminishing}. In Theorem~\ref{theorem:diminishing}, we reveal that using the diminishing step size of Assumption~\ref{assump:stepsizes}-\ref{assump:stepsizes:diminishing}, (a) all devices reach consensus asymptotically, i.e., each device $i$'s model $\mathbf{w}_i^{(k)}$ converges to $\mathbf{\bar{w}}^{(k)} = (1/m) \sum_{i=1}^m{\mathbf{w}_i^{(k)}}$ as $k \to \infty$, and (b) the final model across the devices (i.e., $\mathbf{\bar{w}}^{(k)}, k\rightarrow \infty$) minimizes the global loss. Using the constant step size of assumption~\ref{assump:stepsizes}-\ref{assump:stepsizes:constant}, we also show in Theorem~\ref{theorem:constant} that the same results for consensus and optimization hold but with an optimality gap which is proportional to the step size.

We further show in Theorem~\ref{theorem:diminishing} that the convergence rate with a diminishing step size is $\mathcal{O}{( \ln{k} / \sqrt{k} )}$, which is a desirable property for decentralized gradient descent algorithms \cite{lin2021semi}.

\begin{theorem} \label{theorem:constant}
	Let Assumptions~\ref{assump:weights}-\ref{assump:graderror} and \ref{assump:conn} hold, and the constant step size policy of Assumption~\ref{assump:stepsizes}-\ref{assump:stepsizes:constant} be used. Since using a constant step size will make $\mathbf{\Phi}^{(k)}$ and $\mathbf{\Psi}^{(k)}$ of \eqref{eqn:PhikPsikExplicit} time-invariant, we denote these matrices as $\mathbf{\Phi}$ and $\mathbf{\Psi}$. If the step size $\alpha$ satisfies $\alpha \le \frac{1-\rho^2}{8 B L {( 1+\Gamma_1 )}^{B-1}}$ where $\Gamma_1 = \frac{1-\rho^2}{2\sqrt{2}B}$, the following bound holds:
	
    \begin{equation} \label{eqn:conv_rate}
        \mathbf{\Xi}^{(kB)} \le \mathcal{O} {\left( {\rho{\left( \mathbf{\Phi} \right)}}^{k} \right)} \mathbf{\Xi}^{(0)} + \left( \sum_{r=1}^{k-1}{\mathcal{O}{\left( \rho{\left( \mathbf{\Phi} \right)}^{k-r-1} \right)}} + 1 \right) \mathbf{\Psi}.
    \end{equation}
    Letting $k \to \infty$, we will get
    \begin{equation} \label{eqn:optimality_gap}
        \limsup_{k \to \infty}{\mathbf{\Xi}^{(kB)}} \le \frac{\alpha^2 B}{\mathcal{O}{\left( 1 - \rho{\left( \mathbf{\Phi} \right)} \right)}}
        \begin{bmatrix}
            \frac{\sigma^2}{m} + \nu
            \\
            2mB \left\lbrace 2 \frac{\delta^2 + 2 \mu \alpha \nu}{1-\rho^2} + \sigma^2  \right\rbrace
        \end{bmatrix},
	\end{equation}
	where $\nu = (2 / \mu) (B-1) L^2 {( 1+\Gamma_1 )}^{2 (B-1)} ((1+\Gamma_1) \delta^2 / (\sqrt{2} L) + \alpha \sigma^2 ).$
	
	\begin{proof}
		See Appendix~G in the supplementary material.
	\end{proof}
\end{theorem}

\textbf{\textcolor{blue}{Discussion on Theorem~\ref{theorem:constant}.}} This theorem indicates that, using a constant step size, linear convergence is achieved due to the term $\mathcal{O}{( \rho{( \mathbf{\Phi} )}^k )}$ in Eq.~\eqref{eqn:conv_rate}. However, we observe in Eq.~\eqref{eqn:optimality_gap} that using a constant step size will result in an asymptotic optimality gap of $(\alpha^2 B / \mathcal{O}{( 1 - \rho{( \mathbf{\Phi} )} )}) {[ \psi_1, \psi_2 ]}^T$. This gap is proportional to the step size $\alpha$ and the connectivity bound of the information flow graph $B$ (see Proposition~\ref{proposition:conn}). Thus, choosing a smaller $\alpha$ and employing a strategy that conducts communication rounds more frequently (decreasing $B$), results in the optimality gap getting smaller. \textcolor{blue}{Additionally, note that the entries of the optimality gap vector in Eq.~\eqref{eqn:optimality_gap}, i.e., $\psi_1$ and $\psi_2$, depend on the data heterogeneity bound $\delta$ (formalized in Assumption \ref{assump:smooth_convex_graddiv}-\ref{assump:graddiversity}) through $\nu$, and the gradient approximation errors $\sigma$ (formalized in Assumption \ref{assump:graderror}-\ref{assump:graderror:variance}). As expected, we see that a higher $\delta$ results in a higher value for the upper bound. This also implies that in a federated learning setup where data distributions among devices are non-IID -- that is, $\delta \neq 0$ -- the optimality gap cannot be made zero when a constant step size is employed, even if full batch sizes are used for the gradient updates, i.e., $\sigma = 0$.}

Before presenting Theorem~\ref{theorem:diminishing}, we provide a supplementary lemma as a better alternative to Lemma 11 in \cite{pu2021sharp}.
	We will later use this lemma in the proof of Theorem~\ref{theorem:diminishing} in Appendix~H. This key mathematical result helps us obtain exact convergence rates on last iterates in Theorem \ref{theorem:diminishing}, when the diminishing step size policy of Assumption \ref{assump:stepsizes}-\ref{assump:stepsizes:diminishing} is used.
\begin{lemma} \label{lemma:bernoulli}
	Let ${\lbrace \zeta_r \rbrace}_{r=0}^\infty$ be a scalar sequence where $0 < \zeta_r \le 1$, $\forall r \geq 0$. For any $p \geq 1$, we have
	\begin{equation*}
		\prod_{r=s}^k{{\left( 1 - \zeta_r \right)}^p} \le \frac1{p \sum_{r=s}^k{\zeta_r}}.
	\end{equation*}
	
	\begin{proof}
		See Appendix~B in the supplementary material.
	\end{proof}
\end{lemma}

\begin{theorem} \label{theorem:diminishing}
	Let Assumptions~\ref{assump:weights}-\ref{assump:graderror} and \ref{assump:conn} hold, and the diminishing step size policy of Assumption~\ref{assump:stepsizes}-\ref{assump:stepsizes:diminishing} be used with $\alpha^{(k)} = \frac{\alpha^{(0)}}{\sqrt{1+k / \eta}}$. If the step size satisfies
    $$\alpha^{(0)} \le \frac{1-\rho^2}{4\sqrt{2} \sqrt{5-3\rho^2} BL {( 1+\Gamma_1 )}^{B-1}},$$ where $\Gamma_1 = \frac{1-\rho^2}{2\sqrt{2}B}$, the following bound holds:
	\begin{small}
        \begin{equation} \label{eqn:conv_rate_diminish}
            \begin{aligned}
                & \mathbf{\Xi}^{(kB)} \le \frac1{2 \mu \alpha^{(0)}} \mathcal{O}{\left( \frac1{\sqrt{k}} \right)} {\left( \frac{3 + \rho^2}4 \right)}^{K} \mathbf{\Xi}^{(0)}
                \\
                & \begin{aligned}
                    + \Bigg\lbrace \frac{K-1}{2 \mu \alpha^{(0)}} \mathcal{O}{\left( \frac1{\sqrt{k}} \right)} \left( \frac{1 + \rho^2}2 \right) & + \frac{\alpha^{(0)}}2 \mathcal{O}{\left( \frac{\ln{k}}{\sqrt{k}} \right)}
                    \\
                    & + {(\alpha^{(0)})}^2 \mathcal{O}{\left( \frac1k \right)} \Bigg\rbrace B
                    \begin{bmatrix}
                        \widehat{\psi}_1
                        \\
                        \widehat{\psi}_2
                    \end{bmatrix},
                \end{aligned}
            \end{aligned}
        \end{equation}
    \end{small}
    
	\noindent in which we have
	$\widehat{\psi}_1 = \sigma^2/m + \nu^{(0)}$, $\widehat{\psi}_2 = 2mB \left\lbrace 2 (\delta^2 + 2 \mu \alpha^{(0)} \nu^{(0)})/(1-\rho^2) + \sigma^2 \right\rbrace$, and
    \begin{small}$\nu^{(k)} = (2/\mu) \left( B-1 \right) L^2 {\left( 1+\Gamma_1 \right)}^{2 \left( B-1 \right)} \left( (1+\Gamma_1)\delta^2/(\sqrt{2}L) + \alpha^{(k)} \sigma^2 \right).$\end{small}
	
	Letting $k \to \infty$, we will get
	\begin{equation}
		\limsup_{k \to \infty}{\mathbf{\Xi}^{(kB)}} = 0.
	\end{equation}
	
	\begin{proof}
		See Appendix~H in the supplementary material.
	\end{proof}
\end{theorem}

\textbf{\textcolor{blue}{Discussion on Theorem~\ref{theorem:diminishing}.}} Theorem~\ref{theorem:diminishing} implies that using the diminishing step size of $\alpha^{(k)} = \frac{\alpha^{(0)}}{\sqrt{1+k / \eta}}$, a sub-linear rate of convergence $\mathcal{O}{(\ln{k} / \sqrt{k})}$ can be achieved, and that the models of all devices asymptotically converge to the global optimum point. For the more general setup that the diminishing step size is chosen at $\alpha^{(k)} = \frac{\alpha^{(0)}}{{(1+k / \eta)}^\theta}$ with $\theta \in (0.5, 1]$, see Appendix~I. \textcolor{blue}{Also, the upper bound in Eq.~\eqref{eqn:conv_rate_diminish} captures the effect of data heterogeneity level $\delta$ through the vector values $\widehat{\psi}_1$ and $\widehat{\psi}_2$ (since $\nu^{(0)}$ is a function of $\delta$), where a higher value of $\delta$ results in a larger value for the upper bound.}

\noindent \textcolor{blue}{\textbf{Effect of communication sparsity.} Note that sparse communications would affect the transition spectral radius $\rho(\mathbf{\Phi}^{(k)})$ of Eq.~\eqref{eqn:PhikPsikRecursive} through the information flow graph connectivity parameter $B$ (see Proposition~\ref{proposition:conn}, where we show how $B$ captures the effects of both physical connectivity level $B_1$ and communication interval $B_2$ introduced in Assumption~\ref{assump:conn}). This is because $\rho(\mathbf{\Phi}^{(k)}) = \max\{ \phi_{11}^{(k)}, \phi_{22}^{(k)} \}$ (see Appendix~F), and the definitions of $\phi_{11}^{(k)}$ and $\phi_{22}^{(k)}$ given in Lemmas~\ref{lemma:kplusB} and Eq.~\eqref{eqn:phiValues} illustrate the dependence of these terms on the parameter $B$. However, although a larger $B$ would result in a higher value of $\rho(\mathbf{\Phi}^{(k)})$, thus slowing the convergence rate (see Eq.~\eqref{eqn:conv_rate} in Theorem~\ref{theorem:constant}), our Proposition~\ref{proposition:conn} lays out the constraints on the step size $\alpha^{(k)}$ such that we would still get a spectral radius of $\rho(\mathbf{\Phi}^{(k)}) < 1$. We would essentially achieve this by using a smaller step size, as we see an inverse dependence of step size $\alpha^{(k)}$ on the communication level parameter $B$ in Proposition~\ref{proposition:conn}.}

\noindent \textbf{\textcolor{blue}{Significance of theoretical results.}} Importantly, note that in both Theorems~\ref{theorem:constant} and \ref{theorem:diminishing}, we analyze the convergence behavior of our methodology for the last iterates of model parameters, while seminal papers have focused on average iterates \cite{sundhar2010distributed, nedic2009distributed}, i.e., $(1/T) \sum_{t=0}^{T-1}{\mathbf{w}^{(t)}}$. We provide exact convergence rates for the last iterates of model parameters in spite of the thresholds in the event-triggering mechanism being different from device to device and the underlying physical network connecting the agents being time-varying, i.e., based on Assumption \ref{assump:conn}. Thus, this is a stronger result that complements those provided in \cite{sundhar2010distributed, nedic2009distributed}, and more generally has not been shown in the literature to date. Furthermore, while papers such as \cite{tsianos2012distributed, nedic2013distributed, koloskova2020unified} also assume strong convexity as we have done in Assumption~\ref{assump:smooth_convex_graddiv}-\ref{assump:convexity}, they are still confined to showing convergence for average iterates.

Furthermore, the analysis of existing research was based on the (sub)-gradient bound assumption \cite{zehtabi2022decentralized}, while we have replaced that with two more general assumptions, namely smoothness (Assumption~\ref{assump:smooth_convex_graddiv}-\ref{assump:smoothness}) and statistical heterogeneity of data (Assumption~\ref{assump:smooth_convex_graddiv}-\ref{assump:graddiversity}). In both Theorems~\ref{theorem:constant} and \ref{theorem:diminishing}, we show convergence for a $B$-connected time-varying graph (see Proposition \ref{proposition:conn}), in spite of not making the more restrictive bounded (sub)gradients assumption \cite{george2020distributed}.

\noindent \textbf{\textcolor{blue}{Summary of analysis.}} To summarize, the goals of Proposition \ref{proposition:nonIncreasing} and Theorems \ref{theorem:constant} and \ref{theorem:diminishing} of our paper are to (i) show that convergence to a globally optimal solution of FL can be achieved even when the event-triggering thresholds are personalized for the devices, and (ii) obtain bounds on the rate of model training convergence under such conditions, i.e., for general time-varying consensus graphs.

\section{Numerical Results}\label{sec:simulation}

\begin{figure*}[t]
	\begin{subfigure}[b]{0.5\textwidth}
		\begin{center}
			\includegraphics[width=\textwidth]{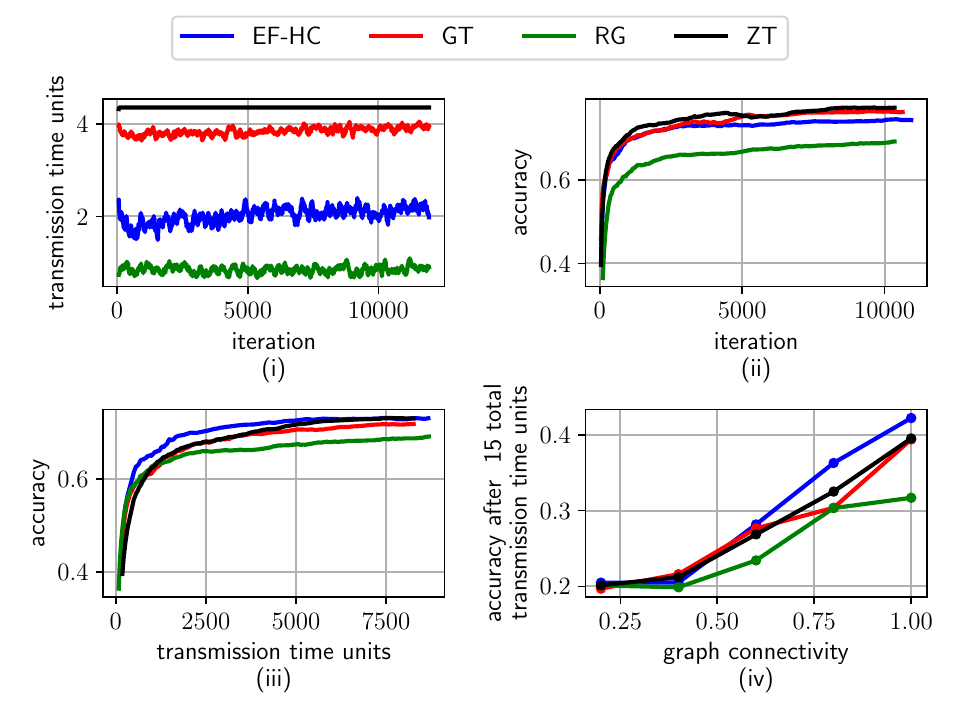}
		\end{center}
        \vspace{-2mm}
		\caption{FMNIST dataset}
		\label{fig:sim:svm_fmnist}
	\end{subfigure}
	\begin{subfigure}[b]{0.5\textwidth}
		\begin{center}
			\includegraphics[width=\textwidth]{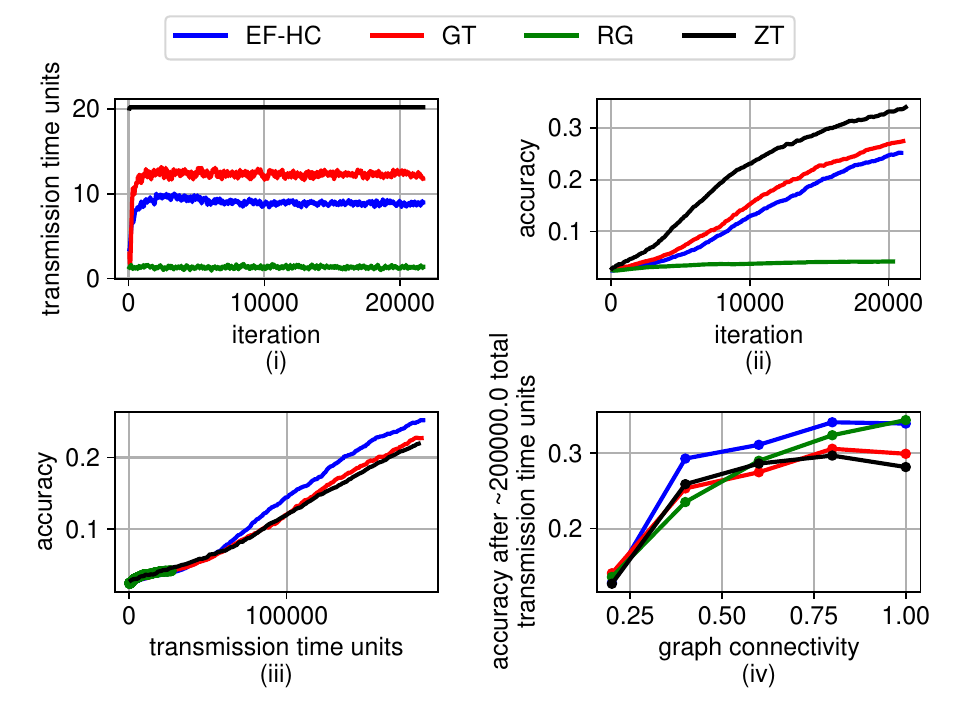}
		\end{center}
        \vspace{-2mm}
		\caption{FEMNIST dataset}
		\label{fig:sim:svm_femnist}
	\end{subfigure}
	\caption{\small{Performance comparison between our method ({\tt EF-HC}), global threshold (GT), zero threshold (ZT), and randomized gossip (RG) algorithms on (a) FMNIST and (b) FEMNIST datasets using an SVM model. The resources are allocated to devices using a uniform distribution $\mathcal{U}{\left( (1-\sigma_N)b_M, (1+\sigma_N)b_M \right)}$ under a random geometric graph. The plots show (i) transmission time per iteration, (ii) accuracy per iteration, (iii) accuracy per transmission time, and (iv) accuracy after a certain number of transmissions with respect to graph connectivity. For this figure, the link bandwidths among devices are generated using a uniform distribution $\mathcal{U}{\left( (1-\sigma_N)b_M, (1+\sigma_N)b_M \right)}$. The devices themselves are connected to each other via random geometric graph. We see how our \texttt{EF-HC} algorithm achieves higher accuracies with less transmission time passed in Figs.~\ref{fig:sim:svm_fmnist}-(iii) and \ref{fig:sim:svm_femnist}-(iii), and also how its advantage remains consistent across different graph connectivities in Figs.~\ref{fig:sim:svm_fmnist}-(iv) and \ref{fig:sim:svm_femnist}-(iv).}}
	\label{fig:sim}
    \vspace{-3mm}
\end{figure*}

\begin{figure*}[t]
	\begin{subfigure}[b]{0.49\textwidth}
		\begin{center}
			\includegraphics[width=\textwidth]{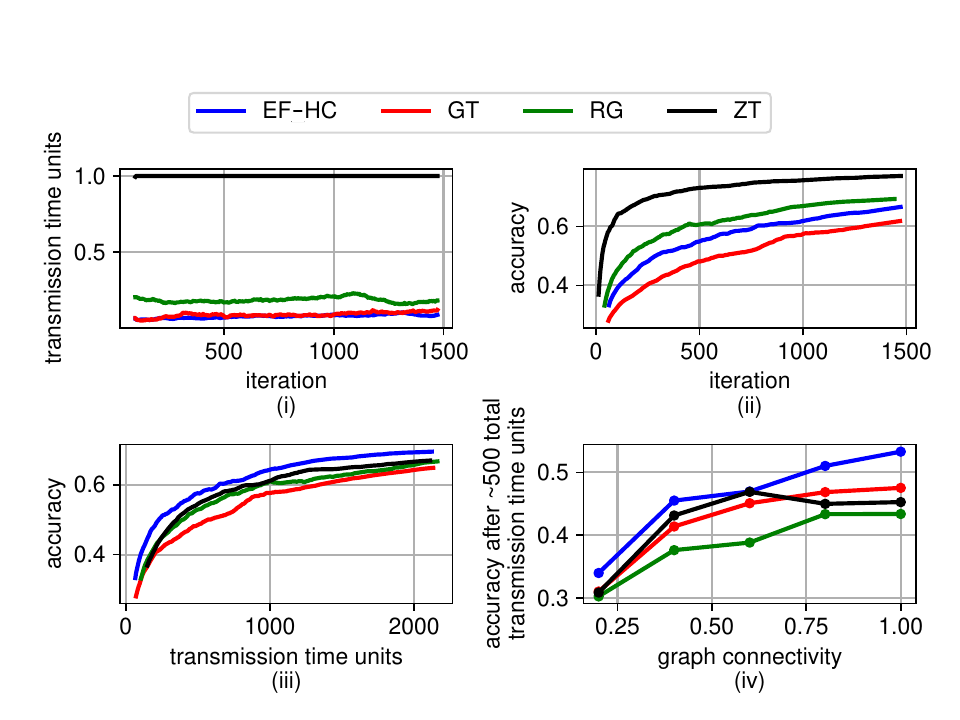}
		\end{center}
        \vspace{-5mm}
		\caption{Random geometric graph.}
		\label{fig:sim_new:svm_beta}
	\end{subfigure}
	\begin{subfigure}[b]{0.49\textwidth}
		\begin{center}
			\includegraphics[width=\textwidth]{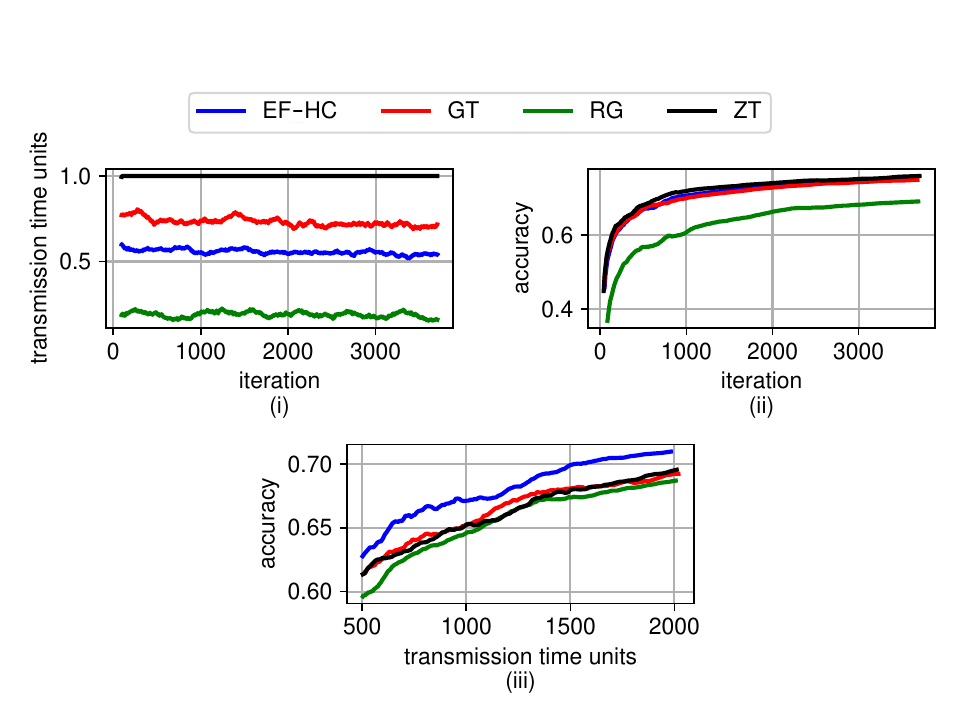}
		\end{center}
        \vspace{-5mm}
		\caption{Internet AS network graph.}
		\label{fig:sim_new:svm_internet}
	\end{subfigure}
	\caption{\small{Performance comparison between our method ({\tt EF-HC}) and the baselines using the FMNIST dataset. For this figure, the link bandwidths among devices are generated using a beta distribution $\Beta(0.5, 0.5) \cdot b_M$. The devices in Figs.~\ref{fig:sim_new:svm_beta} and \ref{fig:sim_new:svm_internet} are connected to each other via a random geometric graph and the Internet graph, respectively. We observe that regardless of the network topology, our \texttt{EF-HC} algorithm achieves higher accuracies faster in terms of the total transmission time passed. Also, comparing Fig.~\ref{fig:sim:svm_fmnist} to Fig.~\ref{fig:sim_new:svm_beta}, we observe that our proposed methodology outperforms the baselines for both uniform and beta distribution, which are used to sample the link bandwidths. (Note that the connectivity of the Internet graph is fixed and cannot be varied as for the random geometric graph.)}}
	\label{fig:sim_new}
    \vspace{-5mm}
\end{figure*}

\noindent In this section, we conduct numerical evaluations to assess the effectiveness of our methodology. We explain the setup of our experiments in Sec.~\ref{ssec:setup} and provide the results and discussion in Sec.~\ref{ssec:discussion}. For further experiments and ablation studies, please see Appendix~J.

\subsection{Simulation Setup}\label{ssec:setup}

\noindent \textbf{Datasets and models.}
We evaluate our proposed methodology using two image classification tasks: Fashion-MNIST (FMNIST) \cite{xiao2017fashion}, and Federated Extended MNIST (FEMNIST) \cite{caldas2018leaf}. Note that FMNIST contains data belonging to $10$ labels, while FEMNIST contains data points with $62$ different labels. We employ two models as classifiers, a support vector machine (SVM) and a 5-layer convolutional neural network (CNN). The loss function $\ell_{\xi}{( \mathbf{w} )}$ in \eqref{eqn:localloss} is chosen as the multi-margin loss for the SVM model, and the cross-entropy loss for the CNN. Note that SVM satisfies the convexity assumption (see~\ref{assump:smooth_convex_graddiv}-\ref{assump:convexity}), while the CNN does not; thus, we will numerically evaluate the efficacy of \texttt{EF-HC} using both convex and non-convex models, although our theoretical analysis only covers convex loss functions.

\noindent \textbf{Graph topology and data distribution.}
In the simulations for the FMNIST and FEMNIST datasets, a network of devices $m=10$ and $m=30$ is used, respectively, in which the underlying communications topology is generated according to random graphs. We conduct evaluations on two types of graphs: (i) random geometric graph, resembling local wireless networks~\cite{hmamouche2021new,hosseinalipour2022multi}, with radius $0.4$ by default; and (ii) the Internet graph of autonomous systems (AS) from \cite{elmokashfi2010scalability}. In the Internet graph, AS are categorized into four types: tier-1, mid-level, customer and content providers, which are divided into different regions to model geographical constraints \cite{elmokashfi2010scalability}. We treat each AS as a node in our system. Further, to generate non-IID data distributions across devices, each device only contains samples of the dataset from a subset of the labels. For FMNIST and FEMNIST, we consider 1 and 3 labels/device, respectively.

\noindent \textbf{Resource heterogeneity.}
Link bandwidths $b_i$ are randomly chosen for each device $i$ from a probability distribution. For completeness, we have run our experiments using two different distributions; (i) uniform distribution $\mathcal{U}{\left( (1-\sigma_N)b_M, (1+\sigma_N)b_M \right)}$, with a mean of $b_M = 5000$ and a normalized standard deviation of $\sigma_N = 0.9$, and (ii) the beta distribution $\Beta(\alpha, \beta) \cdot b_M$ with $\alpha = \beta = 0.5$ for a inverted bell-shaped distribution. For the uniform distribution, we define $\sigma_N = \sigma \sqrt{3}/b_M$, in which $\sigma$ is the standard deviation of the uniform distribution. For uniform distribution, the heterogeneity of the system resources is controlled by the standard deviation, since the value of $\sigma_N=0$ means that all devices are homogeneous in terms of resource capabilities, and $\sigma \to 1$ means choosing $b_i$ values from the range $\mathcal{U}{\left( 0, 2b_M \right)}$. After randomly choosing $b_i$ for each device, we assign the same value as the bandwidth of all outgoing links of the device $i$ for simplicity, i.e., we do not assign different values for each outgoing link of a device $i$.

In each simulation, the diminishing step size is selected as $\alpha^{(k)} = 
0.1 / \sqrt{1+k}$, and the threshold decay rate is set to $\gamma^{(k)} = \alpha^{(k)}$. Also, in~\eqref{eq:thresh}, we set the threshold $r=b_M \times 5 \times 10^{-2}$ for FMNIST, and $r=b_M \times 10^{-1}$ for FEMNIST.

\noindent \textbf{Metrics.}
At iteration $k$, we define a resource utilization score as $(1/m) \sum_{i=1}^m{\sum_{j=1}^m{(v_{ij}^{(k)}} / d_i^{(k)}) \rho_i n}$, which for our proposed method where $\rho_i = 1 / b_i$, this score is the same as the average transmission time, that is, $(1/m) \sum_{i=1}^m{(\sum_{j=1}^m{v_{ij}^{(k)}} / d_i^{(k)}) (n / b_i)}$. The term $\sum_{j=1}^m{v_{ij}^{(k)}} / d_i^{(k)}$ is the utilization of the outgoing links for device $i$, making this score the weighted average of link utilization, penalizing devices with larger $\rho_i$.

\subsection{Results and Discussion}\label{ssec:discussion}

We compare the performance of our method {\tt EF-HC} against three baselines: (1) Distributed learning with aggregations at every iteration, i.e., using zero thresholds (denoted by \textit{ZT}), which is a foundational synchronous method in contrast to our event-triggered approach; (2) Decentralized event-triggered FL, with the same global threshold $r \rho \gamma^{(k)}$ across all devices (denoted by \textit{GT}), where $\rho = 1 / b_M$ is chosen as the average of personalized thresholds of \texttt{EF-HC} for a fair comparison; (3) Randomized gossip, where each device engages in broadcast communication with probability of $1/m$ at each iteration~\cite{pu2021distributed} (denoted by \textit{RG}). The parameter $r$ in both \texttt{EF-HC} and GT as discussed in Sec. IV-A of the paper is chosen so that their frequency of communications would be comparable to RG, with the difference that RG does not conduct communications in an intelligent event-triggered way and instead does them randomly. We illustrate the performance of our method against these baselines in Fig.~\ref{fig:sim}.

\begin{figure*}[ht]
    \begin{subfigure}[b]{0.46\textwidth}
    	\begin{center}
    		\includegraphics[width=\textwidth]{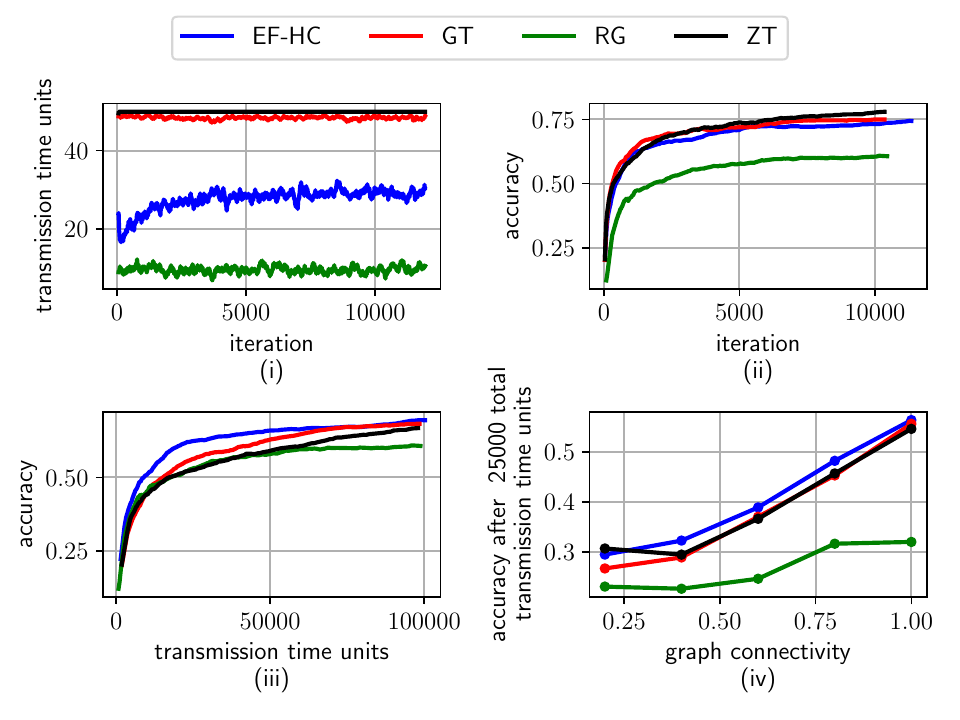}
    	\end{center}
        \vspace{-5mm}
        \caption{Uniformly distributed link bandwidths.}
        \label{fig:sim:lenet5_fmnist:eff}
    \end{subfigure}
    \hfill
    \begin{subfigure}[b]{0.52\textwidth}
		\begin{center}
			\includegraphics[width=\textwidth]{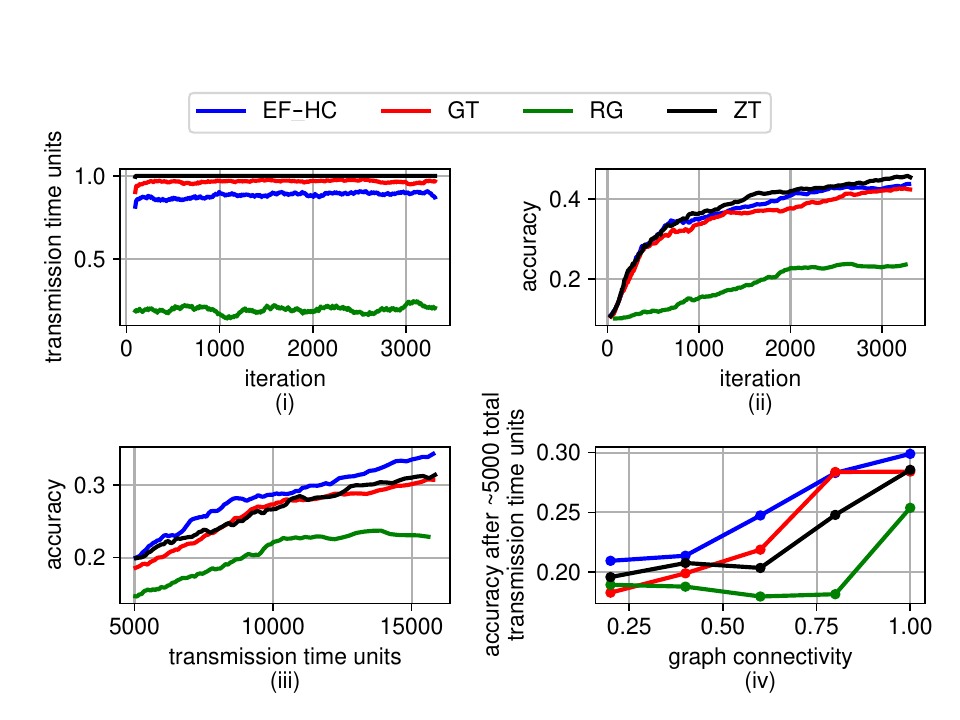}
		\end{center}
        \vspace{-5mm}
		\caption{Beta distributed link bandwidths.}
		\label{fig:sim:lenet5_fmnist:beta}
	\end{subfigure}
    \label{fig:sim:lenet5_fmnist}
    \caption{\small{Performance comparison between our method ({\tt EF-HC}) and the baselines using a CNN classifier on the FMNIST dataset. We use a random geometric graph as the network topology, and sample the link bandwidths in two different ways: (a) uniform distribution $\mathcal{U}{\left( (1-\sigma_N)b_M, (1+\sigma_N)b_M \right)}$ and (b) beta distribution $\Beta(0.5, 0.5) \cdot b_M$. We observe that the superiority of our \texttt{EF-HC} algorithm holds when using a non-convex model as well.}}
    \vspace{-5mm}
\end{figure*}

\noindent \textbf{Communication resource usage.}
We first illustrate the average transmission time units each algorithm requires per training iteration in Figs.~\ref{fig:sim:svm_fmnist}-(i), \ref{fig:sim:svm_femnist}-(i), \ref{fig:sim_new:svm_beta}-(i), \ref{fig:sim_new:svm_internet}-(i), \ref{fig:sim:lenet5_fmnist:eff}-(i) and \ref{fig:sim:lenet5_fmnist:beta}-(i). As we can see, {\tt EF-HC} results in a shorter transmission delay compared to \textit{ZT} and \textit{GT}, significantly helping to resolve the impact of stragglers by not requiring the same amount of communications from devices with less available bandwidth.
However, it is important to note that although a shorter transmission delay per iteration is beneficial for a decentralized optimization algorithm, it can negatively impact the performance of the classification task. Hence, a better comparison between multiple decentralized algorithms is to measure the accuracy reached per transmission time units. In this regard, although \textit{RG} achieves less transmission delay per iteration compared to our method in most cases, Figs.~\ref{fig:sim:svm_fmnist}-(iii), \ref{fig:sim:svm_femnist}-(iii), \ref{fig:sim_new:svm_beta}-(iii), \ref{fig:sim_new:svm_internet}-(iii), \ref{fig:sim:lenet5_fmnist:eff}-(iii) and \ref{fig:sim:lenet5_fmnist:beta}-(iii) reveal that it achieves substantially lower model performance, indicating that our method strikes an effective balance between these objectives.

\noindent \textbf{Accuracy achieved per iteration of training.}
Figs.~\ref{fig:sim:svm_fmnist}-(ii), \ref{fig:sim:svm_femnist}-(ii), \ref{fig:sim_new:svm_beta}-(ii), \ref{fig:sim_new:svm_internet}-(ii), \ref{fig:sim:lenet5_fmnist:eff}-(ii) and \ref{fig:sim:lenet5_fmnist:beta}-(ii) depict the average accuracy of the devices per iteration. These plots are indicative of processing efficiency since they evaluate the accuracy of algorithms per number of gradient descent computations. As expected, the baseline \textit{ZT} is able to achieve the highest accuracy per iteration, since it does not take into account resource efficiency and thus sacrifices network resources to achieve better accuracy. In other words, the value of $B$ explained in Proposition~\ref{proposition:conn} for \textit{ZT} has the minimum possible value compared to other algorithms, since $B_2$ of Assumption~\ref{assump:conn}-\ref{assump:conn:boundedintercom} has the value of $B_2 = 1$ for it.
In most of these plots, that is, Figs.~\ref{fig:sim:svm_fmnist}-(ii), \ref{fig:sim:svm_femnist}-(ii), \ref{fig:sim_new:svm_internet}-(ii), \ref{fig:sim:lenet5_fmnist:eff}-(ii) and \ref{fig:sim:lenet5_fmnist:beta}-(ii), we show that unlike \textit{RG}, the performance of our proposed method {\tt EF-HC} as well as \textit{GT}, which is also event-triggered, does not degrade considerably although they use less communication resources, as will be discussed next.

\noindent \textbf{Accuracy achieved per total delay.}
Figs.~\ref{fig:sim:svm_fmnist}-(iii), \ref{fig:sim:svm_femnist}-(iii), \ref{fig:sim_new:svm_beta}-(iii), \ref{fig:sim_new:svm_internet}-(iii), \ref{fig:sim:lenet5_fmnist:eff}-(iii) and \ref{fig:sim:lenet5_fmnist:beta}-(iii) are the most critical results, as they assess the accuracy vs. communication time trade-off. We see that our algorithm {\tt EF-HC} can achieve higher accuracy while using less transmission time compared to all baselines. These plots reveal that our method can adapt to non-IID data distributions across devices, which is an important characteristic of FL algorithms~\cite{kairouz2021advances}, and achieve better accuracy compared to baselines given a fixed transmission time, that is, under a fixed network resource consumption.

\noindent \textbf{Effect of graph connectivity.}
Furthermore, we evaluated the effect of network connectivity on our method and baselines in Figs.~\ref{fig:sim:svm_fmnist}-(iv), \ref{fig:sim:svm_femnist}-(iv), \ref{fig:sim_new:svm_beta}-(iv), \ref{fig:sim:lenet5_fmnist:eff}-(iv) and \ref{fig:sim:lenet5_fmnist:beta}-(iv). Since the graphs are generated randomly in our simulations, we have taken the average performance of all four algorithms over $5$ Monte Carlo instances to reduce the effect of random initialization. It can be seen that higher network connectivity improves the convergence speed of our method and most of the baselines. Importantly, however, we see that our method has the highest improvement per increase in connectivity. This becomes more pronounced in Figs.~\ref{fig:sim_new:svm_beta}-(iv) and \ref{fig:sim:lenet5_fmnist:beta}-(iv), as the degree of resource heterogeneity between devices is higher because they are sampled from a beta distribution (vs. uniform in Figs.~\ref{fig:sim:svm_fmnist}-(iv), \ref{fig:sim:svm_femnist}-(iv) and \ref{fig:sim:lenet5_fmnist:eff}-(iv)), and \texttt{EF-HC} is best suited for highly heterogeneous scenarios.

\noindent \textbf{Non-convex model.}
To generate non-IID data distributions across devices, each device only contains samples of the dataset from a subset of labels, specifically $2$ labels/device in these experiments. Note that in Figs.~\ref{fig:sim:lenet5_fmnist:eff}-(iv) and \ref{fig:sim:lenet5_fmnist:beta}-(iv), we change the simulation setup and set $r = b_M\times10^{-3}$, and let the devices have samples of only $1$ labels/device. Looking at Fig.~\ref{fig:sim:lenet5_fmnist:eff}, we can see that results similar to those of the SVM classifier (see Fig.~\ref{fig:sim} and Sec.~\ref{ssec:discussion}) can be achieved with a CNN classifier as well, i.e., the results hold with and without the model convexity assumption used in our convergence analysis. Furthermore, the gap between the accuracy achieved by \texttt{EF-HC} in Fig.~\ref{fig:sim:lenet5_fmnist:eff}-(iii) per a given delay compared to other baselines is more significant than its linear SVM counterpart.

\noindent \textbf{Summary of improvements.}
Finally, we note that Figs.~\ref{fig:sim:svm_fmnist}, \ref{fig:sim:svm_femnist}, \ref{fig:sim_new:svm_beta}, \ref{fig:sim_new:svm_internet}, \ref{fig:sim:lenet5_fmnist:eff} and \ref{fig:sim:lenet5_fmnist:beta} collectively demonstrate that our \texttt{EF-HC} algorithm's improvements hold under various settings. First, by comparing Figs. \ref{fig:sim:svm_fmnist} and \ref{fig:sim:svm_femnist} we can see that \texttt{EF-HC} outperforms all baselines under different datasets that the model is being trained, i.e., FMNIST and FEMNIST, respectively. Second, a comparison of Figs. \ref{fig:sim:svm_fmnist} and \ref{fig:sim_new:svm_beta} (or Figs.~\ref{fig:sim:lenet5_fmnist:eff} and \ref{fig:sim:lenet5_fmnist:beta}) illustrates that the improvements of \texttt{EF-HC} hold when different probability distributions are employed for sampling link bandwidths, i.e., uniform and beta distributions, respectively. Third, Figs. \ref{fig:sim:svm_fmnist} and \ref{fig:sim_new:svm_internet} show that \texttt{EF-HC} maintains its effectiveness in accuracy vs. resource utilization trade-off for different graph topologies connecting the devices together, i.e., random geometric graph and internet AS graph, respectively. Finally, we can compare Figs.~\ref{fig:sim:svm_fmnist} and \ref{fig:sim:lenet5_fmnist:eff} (or Figs.~\ref{fig:sim_new:svm_beta} and \ref{fig:sim:lenet5_fmnist:beta}) that \texttt{EF-HC} maintains its advantage gap from the baselines for both
convex and non-convex models, i.e., linear SVM and a CNN architecture, respectively.

\vspace{-2mm}
\section{Conclusion and Future Work}
\noindent In this paper, we develop a novel methodology for decentralized FL, in which model aggregations are performed through D2D communications among devices. We proposed an asynchronous, event-triggered communications mechanism in which each device decides itself when to broadcast its model parameters to its neighbors. Furthermore, to alleviate straggler effects, we developed personalized thresholds for event-triggering conditions in which each device determines its communication frequency according to its available bandwidth. Through theoretical analysis, we demonstrated that our algorithm converges to the global optimal model with a $\mathcal{O}{( \ln{k} / \sqrt{k} )}$ rate for appropriate step sizes. Our analysis holds for the last iterates under relaxed graph connectivity and data heterogeneity assumptions. To do so, we showed that the graph of information flow among devices is connected under our method, despite the fact that sporadic communications are conducted over a time-varying network graph.

Our work also gives rise to various future directions. For instance, it is promising to extend our methodology to consider event-triggering in gradient computations as well, complementing the event-triggered communications framework established in this work.

\vspace{-3mm}

\bibliographystyle{IEEEtran}
\bibliography{IEEEabrv,ton}


\vspace{-1cm}

\begin{IEEEbiography}[{\includegraphics[width=1in,height=1.25in,clip,keepaspectratio]{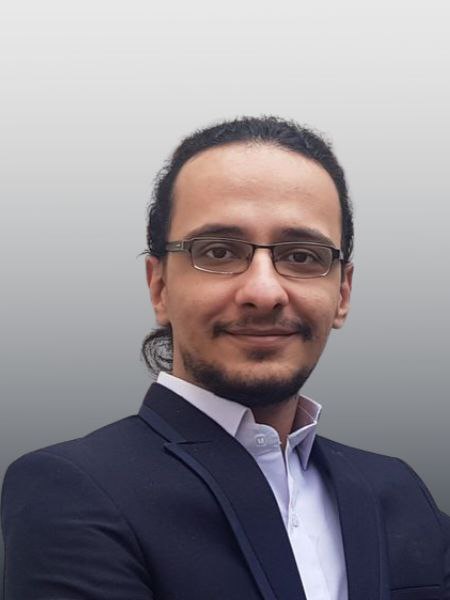}}]{Shahryar Zehtabi}
is a PhD student in ECE at Purdue University. He received his M.Sc. in ECE from Purdue University in 2024. He was the recipient of the 2024 Magoon Award for Graduate Teaching Assistants. He received a dual-degree B.Sc. in EE and CE from Amirkabir University of Technology in 2020 and 2021, respectively.
\end{IEEEbiography}

\vspace{-1cm}

\begin{IEEEbiography}[{\includegraphics[width=1in,height=1.25in,clip,keepaspectratio]{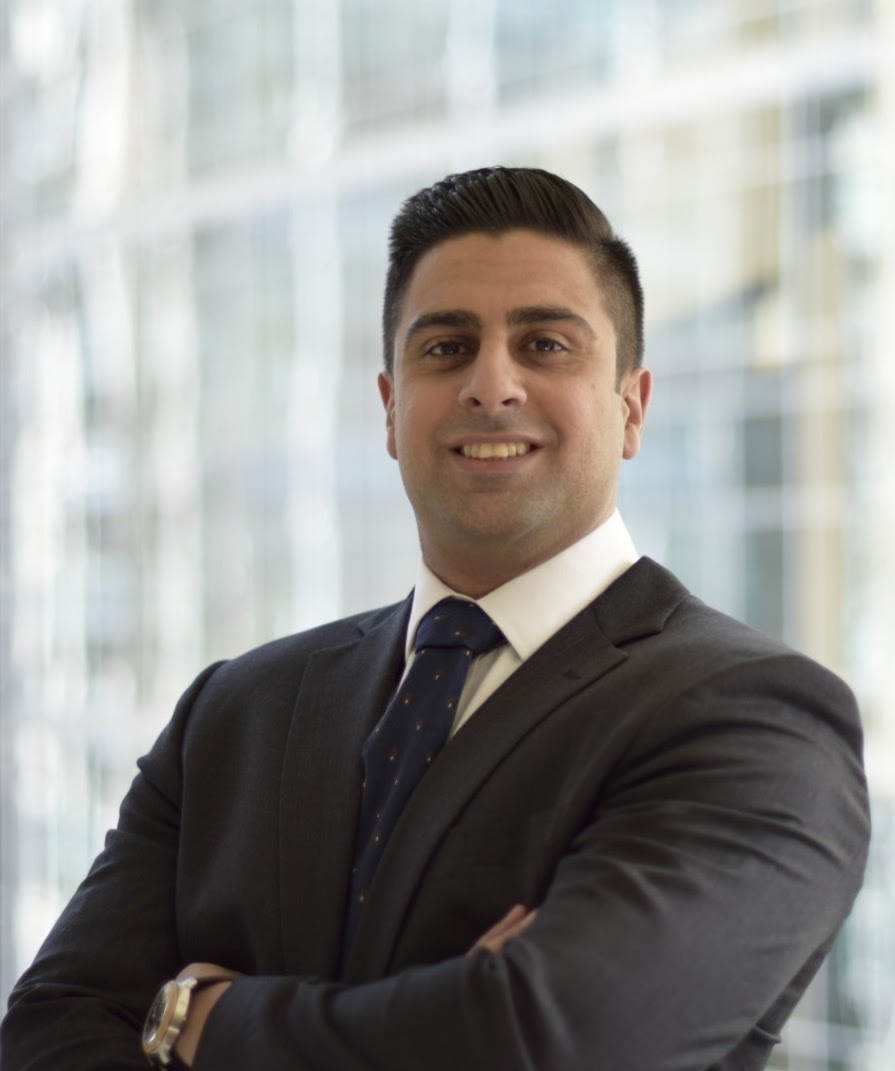}}]{Seyyedali Hosseinalipour} is currently an Assistant Professor with the EE Department at the University at Buffalo–SUNY. He received his Ph.D. and M.Sc. in EE from NCSU in 2020 and 2017, respectively, and was a postdoctoral researcher at Purdue University from 2020 to 2022. He has won the 2020 ECE Doctoral Scholar of the Year Award and 2021 ECE Distinguished Dissertation Award at NCSU.
\end{IEEEbiography}

\vspace{-1cm}

\begin{IEEEbiography}[{{\includegraphics[width=1in,height=1.25in,clip,keepaspectratio]{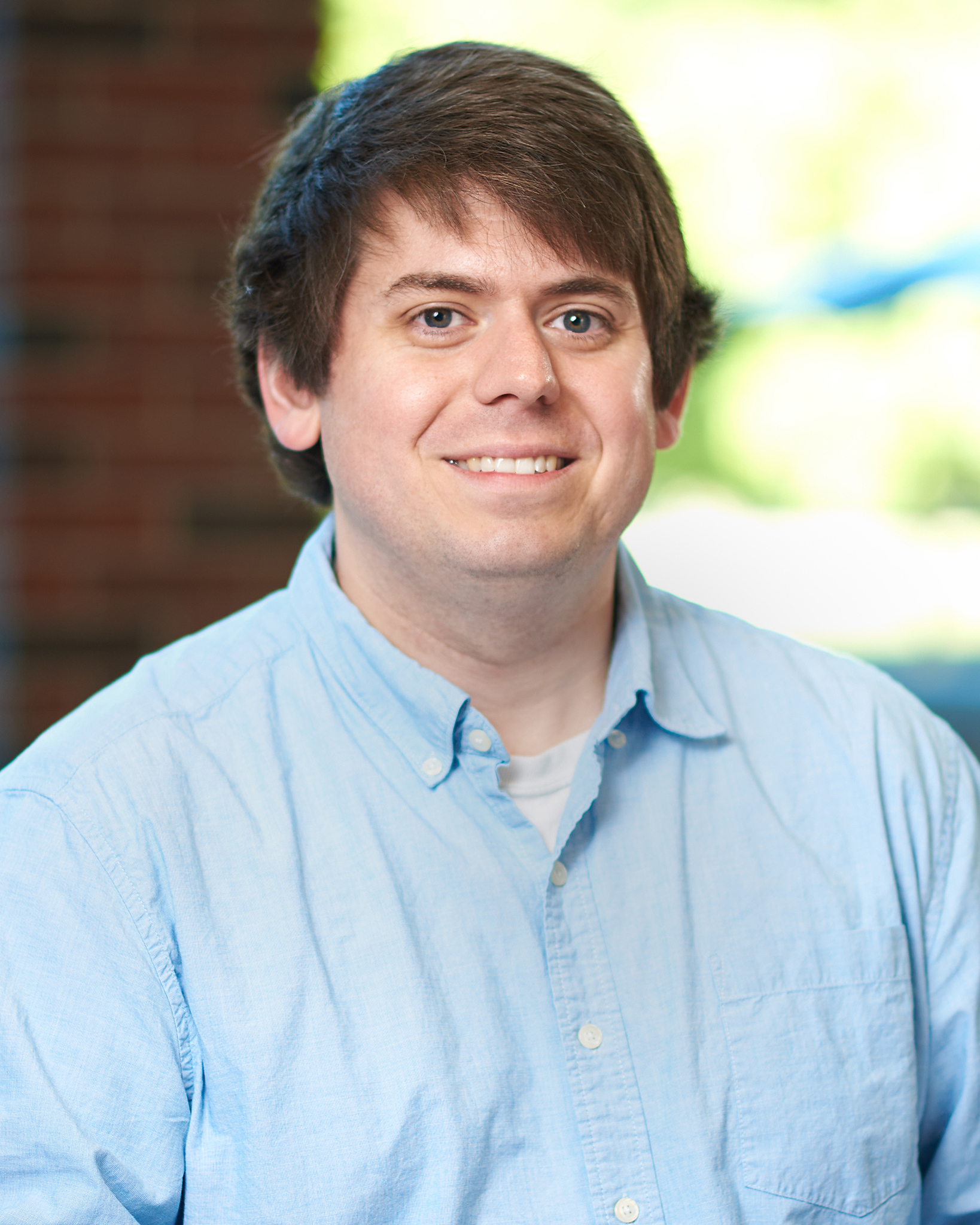}}}]{Christopher G. Brinton}
is the Elmore Associate Professor of ECE at Purdue University. He received his Ph.D. and M.Sc. in EE from Princeton University in 2016 and 2013, respectively. He is a recipient of four of the US top early career awards, from the National Science Foundation (CAREER), Office of Naval Research (YIP), Defense Advanced Research Projects Agency (YFA), and Air Force Office of Scientific Research (YIP). He is also a recipient of the Intel Rising Star Faculty Award and the Qualcomm Faculty Award.
\end{IEEEbiography}




\onecolumn
\begin{appendices}


\onecolumn

\section{Proof of Proposition~\ref{proposition:conn}} \label{appendix:proposition:conn}
\noindent We first introduce some additional definitions for our analysis.

\begin{figure*}[ht]
    \centering
    \includegraphics[width=\textwidth]{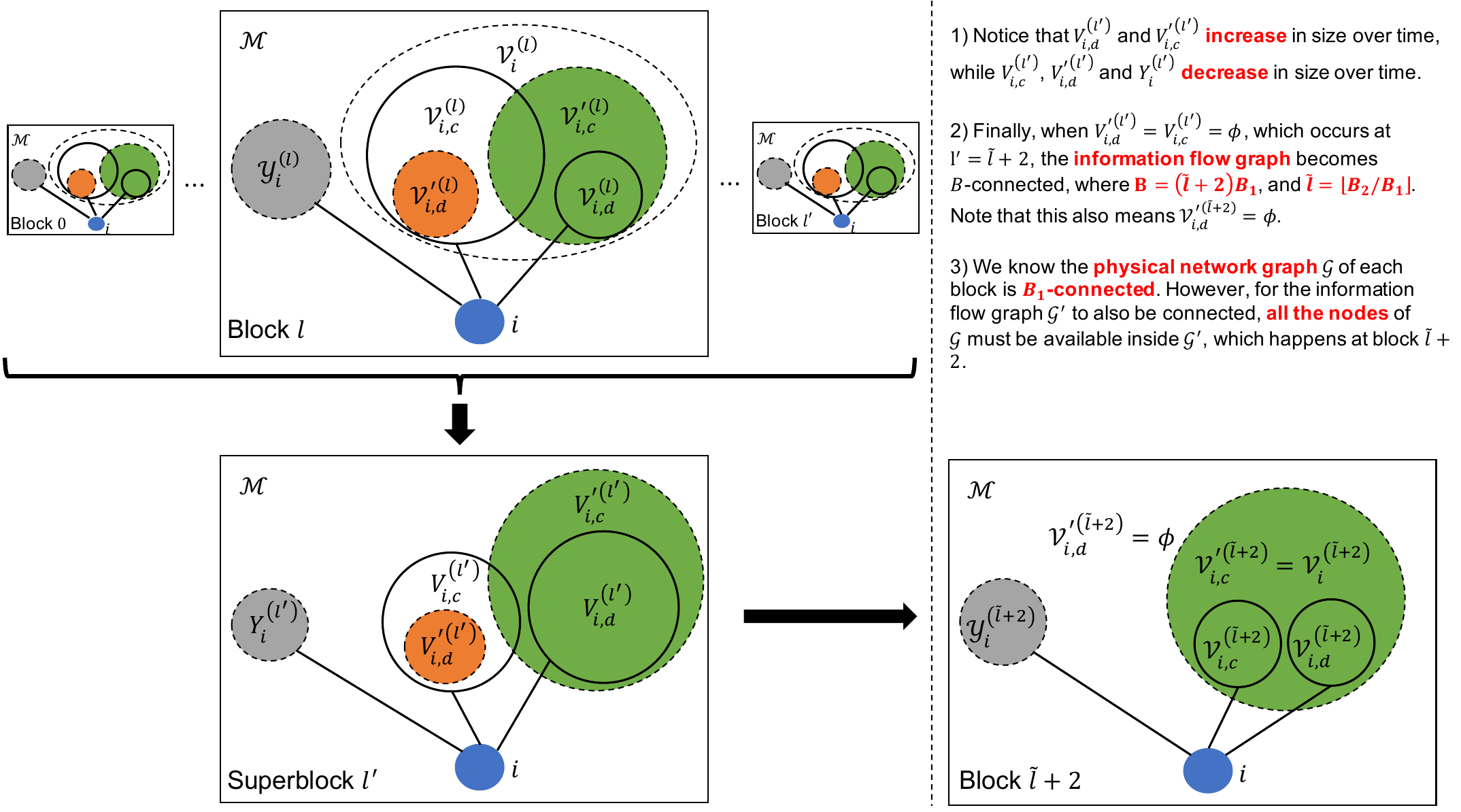}
    \caption{The illustration of definitions and lemmas given for the proof of Proposition~\ref{proposition:conn} in Appendix~\ref{appendix:proposition:conn}.}
    \label{fig:conn}
\end{figure*}

\begin{definition}
    We group the set of iterations after an arbitrary iteration $k = 0,1,\cdots$ into blocks of $B_1$ and we denote block $l$ as iterations $k + l B_1, \cdots, k + ( l+1 ) B_1 - 1$, for $l = 0, 1, \cdots$.
\end{definition}
Note that under Assumption~\ref{assump:conn}-\ref{assump:conn:physicalconn}, the graph union of the physical network graph over each of these blocks, that is, $\mathcal{G}^{(k + l B_1 : k + ( l+1 ) B_1 - 1)}$, is connected for all $k,l \geq 0$. See Fig.~\ref{fig:conn} for a visualization of blocks.

\begin{definition} \label{def:blockV}
    At each block $l \geq 0$, device $i \in \mathcal{M}$ is connected to a subset of devices $\mathcal{V}_i^{(l)} \in \mathcal{M}$, i.e., the edges $( i,j )$ for $j \in \mathcal{V}_i^{(l)}$ are present in $\cup_{s=0}^{B_1-1}{\mathcal{E}^{( k + l B_1 + s )}}$, for all $k \geq 0$. Also, we define $\mathcal{Y}_i^{(l)} = \mathcal{M} \setminus \mathcal{V}_i^{(l)} \setminus i$ as the devices that are not neighbors of device $i$ at block $l$. In Fig.~\ref{fig:conn}, we illustrate device $i$, $\mathcal{V}_i^{(l)}$, $\mathcal{Y}_i^{(l)}$ and $\mathcal{M}$ inside each block.
\end{definition}

\begin{definition} \label{def:blockConn}
    Given the time-varying network topology, we further split $\mathcal{V}_i^{(l)}$ into two disjoint subsets, $\mathcal{V}_{i,c}^{(l)}$ and $\mathcal{V}_{i,d}^{(l)}$, which contain devices connected to and disconnected from device $i$ at the beginning of block $l$, i.e., at iteration $k + l B_1$, for all $k,l \geq 0$, respectively.
\end{definition}
The devices in $\mathcal{V}_{i,d}^{(l)}$ will eventually become connected to $i \in \mathcal{M}$ during block $l$ due to Assumption~\ref{assump:conn}-\ref{assump:conn:physicalconn}, for all $l \geq 0$. The subsets $\mathcal{V}_{i,c}^{(l)}$ and $\mathcal{V}_{i,d}^{(l)}$ are illustrated in Fig.~\ref{fig:conn}.

\begin{definition} \label{def:blockInfo}
    We divide $\mathcal{V}_i^{(l)}$ into two disjoint subsets, $\mathcal{V}'^{(l)}_{i,c}$ and $\mathcal{V}'^{(l)}_{i,d}$ which, respectively, contain the devices that exchange parameters with device $i \in \mathcal{M}$ and those that do not over the course of the block $l \geq 0$. Note that the edges $( i,j )$ for $j \in \mathcal{V}'^{(l)}_{i,c}$ are present in $\cup_{s=0}^{B_1-1}{\mathcal{E}'^{( k + l B_1 + s )}}$, for all $k \geq 0$.
\end{definition}
To make the subsequent mathematical proofs more clear, we show the subsets $\mathcal{V}'^{(l)}_{i,c}$ and $\mathcal{V}'^{(l)}_{i,d}$ in Fig.~\ref{fig:conn}. It is important to make the distinction between the four subsets, $\mathcal{V}_{i,c}^{(l)}$ and $\mathcal{V}_{i,d}^{(l)}$ in one hand, $\mathcal{V}'^{(l)}_{i,c}$ and $\mathcal{V}'^{(l)}_{i,d}$ in the other.

\begin{definition}
    Starting from any iteration $k \geq 0$, multiple consecutive blocks can be further grouped to form a super block. In our analysis, we focus only on the super block which groups the first blocks $0, \cdots, l'$ together for all $l' \geq 0$. In other words, the super block consists of iterations $k, \cdots, k + l B_1, \cdots, k + ( l+1 ) B_1 -1, \cdots, k + ( l'+1 ) B_1 -1$ where $l$, $0 \le l \le l'$, indicates the $l$-th block and $k \geq 0$.
\end{definition}

\begin{definition} \label{def:superBlock}
    For the super block described above, we define the following corresponding sets:
    \begin{equation*}
        \begin{gathered}
            V_i^{(l')} = \cup_{l=0}^{l'}{\mathcal{V}_i^{(l)}}, \qquad Y_i^{(l')} = \cap_{l=0}^{l'}{\mathcal{Y}_i^{(l)}},
            \\
            V_{i,d}^{(l')} = \cup_{l=0}^{l'}{\mathcal{V}_{i,d}^{(l)}}, \qquad V_{i,c}^{(l')} = \cap_{l=0}^{l'}{\mathcal{V}_{i,c}^{(l)}},
            \\
            V'^{(l')}_{i,d} = \cap_{l=0}^{l'}{\mathcal{V}'^{(l)}_{i,d}}, \qquad V'^{(l')}_{i,c} = \cup_{l=0}^{l'}{\mathcal{V}'^{(l)}_{i,c}},
        \end{gathered}
    \end{equation*}
    for all $i \in \mathcal{M}$ and $l' \geq 0$. At the super block $l'$, $V_i^{(l')}$ is the set of all devices that are physically connected to device $i$, and $Y_i^{(l')}$ contains the devices that do not. Moreover, $V_{i,d}^{(l')}$ keeps track of all devices that were initially disconnected from device $i$ at the beginning of each block inside the super block, while $V_{i,c}^{(l')}$ only contains those devices that remain connected to device $i$ from the first block up until the last block inside the super block. Similarly, $V'^{(l')}_{i,d}$ is the set that keeps track of those devices that have not exchanged parameters with device $i$ inside the super block, and $V'^{(l')}_{i,c}$ contains all the ones that have.
\end{definition}
It can be easily shown that $V_i^{(l')} \cup Y_i^{(l')} = \mathcal{M} \setminus i$ and $V_{i,d}^{(l')} \cup V_{i,c}^{(l')} = V'^{(l')}_{i,d} \cup V'^{(l')}_{i,c} = V_i^{(l')}$. Notice that $V_{i,c}^{(l')}$ is defined as the intersection of the corresponding subsets in each block, and contains the devices that were connected to the device $i \in \mathcal{M}$ in iteration $k \geq 0$, and stay connected to it at all times until the end of super block at iteration $k + ( l'+1 ) B_1 - 1$, for all $l' \geq 0$.

Next, we provide two supplementary lemmas which will be useful in proving Proposition~\ref{proposition:conn}.
\begin{lemma} \label{lemma:block}
    The relations that relate the subsets defined in \ref{def:blockV}, \ref{def:blockConn}, \ref{def:blockInfo} and \ref{def:superBlock} together are as follows:
    \begin{enumerate}[label=(\alph*)]
        \item In each block $l \geq 0$ and for each device $i \in \mathcal{M}$, we have
        \begin{equation*}
            \mathcal{V}_{i,d}^{(l)} \in \mathcal{V}'^{(l)}_{i,c}, \qquad \mathcal{V}'^{(l)}_{i,d} \in \mathcal{V}_{i,c}^{(l)}.
        \end{equation*}
        \label{lemma:block:intraBlock}
        
        \item For super block $l' \geq 0$ and each device $i \in \mathcal{M}$, we have
        \begin{equation*}
            V_{i,d}^{(l')} \in V'^{(l')}_{i,c}, \qquad V'^{(l')}_{i,d} \in V_{i,c}^{(l')}.
        \end{equation*} \label{lemma:block:intraSuperblock}
        
        \item The set of all devices in $\mathcal{V}_{i,c}^{(l'+1)}$ can be partitioned into three disjoint subsets to get a relationship between block $l'+1$ and the super block consisting of blocks $0, \cdots, l'$, for all $i \in \mathcal{M}$ and $l' \geq 0$. We have
        \begin{equation*}
            \mathcal{V}_{i,c}^{(l'+1)} \in V_{i,d}^{(l')} \cup V_{i,c}^{(l')} \cup \left( Y_i^{(l')} \cap \mathcal{V}_{i,c}^{(l'+1)} \right).
        \end{equation*} \label{lemma:block:interBlock}
    \end{enumerate}
    
    \begin{proof}
        \ref{lemma:block:intraBlock} Given Assumption~\ref{assump:conn}-\ref{assump:conn:physicalconn}, devices in $\mathcal{V}_{i,d}^{(l)}$ will eventually become connected to $i$ in the course of this block. On the other hand, \textit{Neighbor Connection Event} in Alg.~\ref{alg:efhc} enforces that all of these devices exchange model parameters with device $i$ upon connection. Therefore, the set $\mathcal{V}'^{(l)}_{i,c}$ is a super set of $\mathcal{V}_{i,d}^{(l)}$, i.e., $\mathcal{V}_{i,d}^{(l)} \in \mathcal{V}'^{(l)}_{i,c}$, and as a result
        \begin{equation*}
            \mathcal{V}'^{(l)}_{i,d} = \mathcal{V}_i^{(l)} \setminus \mathcal{V}'^{(l)}_{i,c} \in \mathcal{V}_i^{(l)} \setminus \mathcal{V}^{(l)}_{i,d} = \mathcal{V}^{(l)}_{i,c}.
        \end{equation*}
        
        \ref{lemma:block:intraSuperblock} The proof for this follows from Part~\ref{lemma:block:intraBlock} and the definitions of the sets for the super block. Since $\mathcal{V}_{i,d}^{(l)} \in \mathcal{V}'^{(l)}_{i,c}$,
        \begin{equation*}
            V_{i,d}^{(l')} = \cup_{l=0}^{l'}{\mathcal{V}_{i,d}^{(l)}} \in \cup_{l=0}^{l'}{\mathcal{V}'^{(l)}_{i,c}} = V'^{(l')}_{i,c}.
        \end{equation*}
        On the other hand, since $\mathcal{V}'^{(l)}_{i,d} \in \mathcal{V}_{i,c}^{(l)}$, it follows that
        \begin{equation*}
            V'^{(l')}_{i,d} = \cap_{l=0}^{l'}{\mathcal{V}'^{(l)}_{i,d}} \in \cap_{l=0}^{l'}{\mathcal{V}_{i,c}^{(l)}} = V_{i,c}^{(l')}.
        \end{equation*}
        
        \ref{lemma:block:interBlock} We show that these three subsets are disjoint. For the first two, take $j \in V_{i,c}^{(l')} = \cap_{l=0}^{l'}{\mathcal{V}_{i,c}^{(l)}}$, which means $j \in \mathcal{V}_{i,c}^{(l)}$ for all $l = 0, \cdots, l'$. Hence, we have $j \notin \mathcal{V}_i^{(l)} \setminus \mathcal{V}_{i,c}^{(l)} = \mathcal{V}_{i,d}^{(l)}$ for all $l$, so $j \notin \cup_{l=0}^{l'}{\mathcal{V}_{i,d}^{(l)}} = V_{i,d}^{(l')}$.
        
        For the third set, it can easily be shown that $V_{i,d}^{(l')} \cup V_{i,c}^{(l')} = V_i^{(l')}$. We can then prove that $Y_i^{(l')}$ is disjoint from $V_i^{(l')}$, since $j \in Y_i^{(l')} = \cap_{l=0}^{l'}{\mathcal{Y}_i^{(l)}}$ indicates $j \in \mathcal{Y}_i^{(l)}$ for all $l = 0,\cdots,l'$. Hence, we have $j \notin \mathcal{M} \setminus \mathcal{Y}_i^{(l)} \setminus i = \mathcal{V}_i^{(l)}$ for all $l$, so $j \notin \cup_{l=0}^{l'}{\mathcal{V}_i^{(l)}} = V_i^{(l')}$. $Y_i^{(l')}$ and $V_i^{(l')}$ being disjoint implies $Y_i^{(l')} \cap \mathcal{V}_{i,c}^{(l'+1)}$ and $V_i^{(l')}$ are also disjoint.
    \end{proof}
\end{lemma}
Part~\ref{lemma:block:intraBlock} of the above lemma means that devices initially disconnected from device $i$ which become connected to it over the course of a block will definitely exchange parameters with it upon connection. Therefore, devices that might not exchange parameters with device $i$ inside a block will definitely be among those that were initially connected to it. Part~\ref{lemma:block:intraSuperblock} states a similar argument for a super block.

\begin{lemma} \label{lemma:connDeduction}
    The following connectivity characteristics hold for the graph union of the information flow graphs.
    
    \begin{enumerate}[label=(\alph*)]
        \item The union of the information flow graphs within each block $l \geq 0$, from iteration $k + l B_1$ to $k + ( l+1 ) B_1 - 1$, that is, $\mathcal{G}'^{(k + l B_1 : k + ( l+1 ) B_1 - 1)}$, is connected only if $\mathcal{V}_{i,c}^{(l)} \in \mathcal{V}'^{(l)}_{i,c}$, for all $k \geq 0$. \label{lemma:connDeduction:intraBlock}
        
        \item The union of the information flow graphs inside each super block $l' \geq 0$, from iteration $k$ to $k + ( l'+1 ) B_1 - 1$, that is, $\mathcal{G}'^{(k : k + ( l'+1 ) B_1 - 1)}$, is connected only if $\mathcal{V}_{i,c}^{(\tilde{l})} \in V'^{(l')}_{i,c}$ for at least one $\tilde{l}$, $0 \le \tilde{l} \le l'$ and all $k \geq 0$. \label{lemma:connDeduction:intraSuperblock}
    \end{enumerate}
    
    \begin{proof}
        \ref{lemma:connDeduction:intraBlock} First, by the definition of each block, running from iteration $k + l B_1$ to $k + ( l+1 ) B_1 - 1$ for $l=0,1,\cdots$, the physical network graph $\mathcal{G}^{(k + l B_1 : k + ( l+1 ) B_1 - 1)}$ is connected. Therefore, if $\mathcal{V}_i^{(l)} = \mathcal{V}'^{(l)}_{i,c}$, the information flow graph of this block, i.e., $\mathcal{G}'^{(k + l B_1 : k + ( l+1 ) B_1 - 1)}$, will also be connected, since these two graphs are comprised of the same set of edges.
        
        Second, note that $\mathcal{V}'^{(l)}_{i,c} \in \mathcal{V}_i^{(l)}$ is true by definition, so we only need to show $\mathcal{V}_i^{(l)} \in \mathcal{V}'^{(l)}_{i,c}$. Since $\mathcal{V}_i^{(l)} = \mathcal{V}_{i,d}^{(l)} \cup \mathcal{V}_{i,c}^{(l)}$, and as a consequence of Lemma~\ref{lemma:block}-\ref{lemma:block:intraBlock}, $\mathcal{V}_{i,d}^{(l)} \in \mathcal{V}'^{(l)}_{i,c}$, we must also have the condition $\mathcal{V}_{i,c}^{(l)} \in \mathcal{V}'^{(l)}_{i,c}$.
        
        \ref{lemma:connDeduction:intraSuperblock} We take steps similar to those in Lemma~\ref{lemma:connDeduction}-\ref{lemma:connDeduction:intraBlock} here. First, it is straightforward to show that connectivity is preserved under graph union. Therefore, if we find a block $\tilde{l}$ inside the super block $l'$, $0 \le \tilde{l} \le l'$, such that $\mathcal{V}_i^{(\tilde{l})} \in V'^{(l')}_{i,c}$, then the information flow graph of this super block, i.e., $\mathcal{G}'^{(k : k + ( l'+1 ) B_1 - 1)}$, will be connected.
        
        Second, by definition, we have $\mathcal{V}_i^{(\tilde{l})} = \mathcal{V}_{i,d}^{(\tilde{l})} \cup \mathcal{V}_{i,c}^{(\tilde{l})}$ and $V'^{(l')}_{i,c} = \cup_{l=0}^{l'}{\mathcal{V}'^{(l)}_{i,c}}$. Lemma~\ref{lemma:block}-\ref{lemma:block:intraBlock} yields $\mathcal{V}_{i,d}^{(\tilde{l})} \in \mathcal{V}'^{(\tilde{l})}_{i,c} \in \cup_{l=0}^{l'}{\mathcal{V}'^{(l)}_{i,c}}$. Thus, we must have $\mathcal{V}_{i,c}^{(\tilde{l})} \in \cup_{l=0}^{l'}{\mathcal{V}'^{(l)}_{i,c}} = V'^{(l')}_{i,c}$.
    \end{proof}
\end{lemma}

Now we are ready to prove Proposition~\ref{proposition:conn}.
\begin{proof}
    Lemma~\ref{lemma:connDeduction}-\ref{lemma:connDeduction:intraSuperblock} states that for the information flow graph $\mathcal{G}'^{(k)}$ to be $( l'+1 ) B_1$-connected, the relation $\mathcal{V}_{i,c}^{(\tilde{l})} \in V'^{(l')}_{i,c}$ must hold for at least one $\tilde{l}$, $0 \le \tilde{l} \le l'$. According to Lemma~\ref{lemma:block}-\ref{lemma:block:interBlock}, we have $\mathcal{V}_{i,c}^{(\tilde{l}+1)} = V_{i,d}^{(\tilde{l})} \cup V_{i,c}^{(\tilde{l})} \cup ( Y_i^{(\tilde{l})} \cap \mathcal{V}_{i,c}^{(\tilde{l}+1)} )$, therefore we must show $V_{i,d}^{(\tilde{l}-1)} \cup V_{i,c}^{(\tilde{l}-1)} \cup ( Y_i^{(\tilde{l}-1)} \cap \mathcal{V}_{i,c}^{(\tilde{l})} ) \in V'^{(l')}_{i,c}$ to prove connectivity. From Lemma~\ref{lemma:block}-\ref{lemma:block:intraSuperblock},
    \begin{equation*}
        V_{i,d}^{(\tilde{l}-1)} \in V'^{(\tilde{l}-1)}_{i,c} \in V'^{(\tilde{l}-1)}_{i,c} \cup \left( \cup_{l=\tilde{l}}^{l'}{\mathcal{V}'^{(l)}_{i,c}} \right) = V'^{(l')}_{i,c}.
    \end{equation*}
    $Y_i^{(\tilde{l}-1)} \cap \mathcal{V}_{i,c}^{(\tilde{l})}$ describes devices that are connected to $i$ at the beginning of block $\tilde{l}$, i.e., iteration $k + \tilde{l} B_1$, but have not been connected to $i$ in any of the previous blocks $0$ to $\tilde{l}-1$, i.e., between iterations $k$ to $k + \tilde{l} B_1 - 1$. Due to \textit{Neighbor Connection Event} in Alg.~\ref{alg:efhc}, these devices exchange parameters with device $i$ upon connection. Thus, $Y_i^{(\tilde{l}-1)} \cap \mathcal{V}_{i,c}^{(\tilde{l})} \in \mathcal{V}'^{(\tilde{l})}_{i,c} \in V'^{(l')}_{i,c}$.
    
    Next, we need to prove $V_{i,c}^{(\tilde{l}-1)} \in V'^{(l')}_{i,c}$. Recall that the set $V_{i,c}^{(\tilde{l}-1)}$ contains devices that have been connected to $i$ from the beginning of block $0$ (iteration $k$) until the beginning of block $\tilde{l}-1$ (iteration $k + ( \tilde{l} - 1 ) B_1$). Thus, for the devices in this set to exchange parameters with $i$, a \textit{Broadcast Event} (see Alg.~\ref{alg:efhc}) must occur at device $i$. Due to Assumption~\ref{assump:conn}-\ref{assump:conn:boundedintercom}, a \textit{Broadcast event} is guaranteed at least once every $B_2$ iteration on device $i$. If $\tilde{l} B_1 \le B_2 \le ( \tilde{l}+1 ) B_1 - 1$, we know that a \textit{Broadcast event} is triggered in the super block $\tilde{l}+1$, causing device $i$ to exchange parameters with its neighboring devices in this super block. As a result, we write $V_{i,c}^{(\tilde{l}+1)} \in \mathcal{V}'^{(\tilde{l}+1)}_{i,c} \in V'^{(\tilde{l}+1)}_{i,c}$.
    
    Putting everything together concludes the proof, showing that the information flow graph $\mathcal{G}'^{(k)}$ is $( l'+1 ) B_1$-connected if $l' = \tilde{l}+1$, where $\tilde{l}$ is determined by $\tilde{l} B_1 \le B_2 \le ( \tilde{l}+1 ) B_1 - 1$. Thus, $\mathcal{G}'^{(k)}$ is $( \tilde{l}+2 ) B_1$-connected.
\end{proof}

\section{Proof of Lemma~\ref{lemma:bernoulli}} \label{appendix:bernoulli}
We start by reciprocating the left-hand side of the inequality we want to prove. We have
\begin{equation*}
    \frac1{\prod_{r=s}^k{{\left( 1 - \zeta_r \right)}^p}} = \prod_{r=s}^k{\frac1{{\left( 1 - \zeta_r \right)}^p}} = \prod_{r=s}^k{{\left( 1 - \zeta_r \right)}^{-p}} \ge \prod_{r=s}^k{\left( 1 + p \zeta_r \right)} = H{\left( k,s \right)},
\end{equation*}
in which Bernoulli's inequality was used and we defined $H{( k,s )}$ as the lower bound for the reciprocate. Next, we claim that $H{( k,s )} \geq 1 + p \sum_{r=s}^k{\zeta_r}$ for $k \geq s$, and prove it using induction. We have
\begin{equation*}
    \begin{aligned}
        & H{\left( s,s \right)} = 1 + p \zeta_s \geq 1 + p \zeta_s,
        \\
        & H{\left( k+1,s \right)} = \left( 1 + p \zeta_{k+1} \right) H{\left( k,s \right)} \geq \left( 1 + p \zeta_{k+1} \right) \left( 1 + p \sum_{r=s}^k{\zeta_r} \right) = 1 + p \sum_{r=s}^{k+1}{\zeta_r} + p^2 \zeta_{k+1} \sum_{r=s}^k{\zeta_r} \geq 1 + p \sum_{r=s}^{k+1}{\zeta_r}.
    \end{aligned}
\end{equation*}
As a result, we get
\begin{equation*}
    \frac1{\prod_{r=s}^k{{\left( 1 - \zeta_r \right)}^p}} \geq 1 + p \sum_{r=s}^k{\zeta_r} \geq p \sum_{r=s}^k{\zeta_r}.
\end{equation*}
Finally, the statement of this lemma is proved by taking the reciprocal of the last line above.

\section{Proof of Lemma~\ref{lemma:epsilonbounds}} \label{appendix:lemma:epsilonbounds}
To prove the second inequality, we first bound $\mathbb{E}{[ {\| \mathbf{\epsilon}^{(k)} \|}^2 ]}$ as
\begin{equation*}
    \mathbb{E}{\left[ {\left\| \mathbf{\epsilon}^{(k)} \right\|}^2 \right]} = \mathbb{E}{\left[ \sum_{i=1}^m{{\left\| \mathbf{\epsilon}_i^{(k)} \right\|}^2} \right]} = \sum_{i=1}^m{\mathbb{E}{\left[ {\left\| \mathbf{\epsilon}_i^{(k)} \right\|}^2 \right]}} = \sum_{i=1}^m{\sigma_i^2} \le m \sigma^2.
\end{equation*}
Now, we write
\begin{equation*}
    \begin{aligned}
		\mathbb{E}{\left[ {\left\| \mathbf{\epsilon}^{(k)} - \mathbf{1}_m \mathbf{\bar{\epsilon}}^{(k)} \right\|}^2 \right]} & = \mathbb{E}{\left[ {\left\| \mathbf{\epsilon}^{(k)} \right\|}^2 - 2 \left\langle \mathbf{\epsilon}^{(k)}, \mathbf{1}_m \mathbf{\bar{\epsilon}}^{(k)} \right\rangle + {\left\| \mathbf{1}_m \mathbf{\bar{\epsilon}}^{(k)} \right\|}^2 \right]} = \mathbb{E}{\left[ {\left\| \mathbf{\epsilon}^{(k)} \right\|}^2 \right]} - 2 \sum_{i=1}^m{\mathbb{E}{\left[ \left\langle \mathbf{\epsilon}_i^{(k)}, \mathbf{\bar{\epsilon}}^{(k)} \right\rangle \right]}} + m \mathbb{E}{\left[ {\left\| \mathbf{\bar{\epsilon}}^{(k)} \right\|}^2 \right]}
		\\
		& = \mathbb{E}{\left[ {\left\| \mathbf{\epsilon}^{(k)} \right\|}^2 \right]} - \frac2m \sum_{i=1}^m{\mathbb{E}{\left[ {\left\| \mathbf{\epsilon}_i^{(k)} \right\|}^2 \right]}} - \frac2m \sum_{i=1}^m{\left\langle \mathbb{E}{\left[ \mathbf{\epsilon}_i^{(k)} \right]}, \sum_{\substack{j=1 \\ j \neq i}}^m{\mathbb{E}{\left[ \mathbf{\epsilon}_j^{(k)} \right]}} \right\rangle} + m \mathbb{E}{\left[ {\left\| \mathbf{\bar{\epsilon}}^{(k)} \right\|}^2 \right]}
		\\
		& = \left( 1 - \frac2m \right) \mathbb{E}{\left[ {\left\| \mathbf{\epsilon}^{(k)} \right\|}^2 \right]} - \frac2m \sum_{i=1}^m{\left\langle \mathbb{E}{\left[ \mathbf{\epsilon}_i^{(k)} \right]}, \sum_{\substack{j=1 \\ j \neq i}}^m{\mathbb{E}{\left[ \mathbf{\epsilon}_j^{(k)} \right]}} \right\rangle} + m \mathbb{E}{\left[ {\left\| \mathbf{\bar{\epsilon}}^{(k)} \right\|}^2 \right]} \le \left( 1 - \frac2m \right) m \sigma^2 + \sigma^2
		\\
		& = \left( m-1 \right) \sigma^2 \le m \sigma^2.
	\end{aligned}
\end{equation*}

\section{Proof of Lemma~\ref{lemma:kplus1}} \label{appendix:lemma:kplus1}

\ref{lemma:kplus1:gradbound}
Note that
\begin{equation*}
	{\left\| \mathbf{\nabla}^{(k)} - \mathbf{1}_m \mathbf{\bar{\nabla}}^{(k)} \right\|}^2 \le 2 {\left\| \mathbf{\nabla}^{(k)} - \nabla{F}^{(k)} \right\|}^2 + 2 {\left\| \mathbf{1}_m \mathbf{\bar{\nabla}}^{(k)} - \nabla{F}^{(k)} \right\|}^2.
\end{equation*}
We first bound ${\| \mathbf{\nabla}^{(k)} - \nabla{F}^{(k)} \|}^2$ as
\begin{equation*}
    {\left\| \mathbf{\nabla}^{(k)} - \nabla{F}^{(k)} \right\|}^2 = \sum_{i=1}^m{{\left\| \nabla{F_i{\left( \mathbf{w}_i^{(k)} \right)}} - \nabla{F{\left( \mathbf{w}_i^{(k)} \right)}} \right\|}^2} \le \sum_{i=1}^m{\delta_i^2} \le m \delta^2,
\end{equation*}
in which the data heterogeneity bound $\delta$ of Assumption~\ref{assump:smooth_convex_graddiv}-\ref{assump:graddiversity} was used. Next, for ${\| \mathbf{1}_m \mathbf{\bar{\nabla}}^{(k)} - \nabla{F}^{(k)} \|}^2$, we have

\begin{equation*}
    \begin{aligned}
        {\left\| \mathbf{1}_m \mathbf{\bar{\nabla}}^{(k)} - \nabla{F}^{(k)} \right\|}^2 & = \sum_{i=1}^m{{\left\| \mathbf{\bar{\nabla}}^{(k)} - \nabla{F{\left( \mathbf{w}_i^{(k)} \right)}} \right\|}^2} = \sum_{i=1}^m{{\left\| \sum_{j=1}^m{\frac{\nabla{F_j{\left( \mathbf{w}_j^{(k)} \right)}} - \nabla{F_j{\left( \mathbf{w}_i^{(k)} \right)}}}m} \right\|}^2}
        \\
        & \le \sum_{i=1}^m{m \sum_{j=1}^m{\frac{{\left\| \nabla{F_j{\left( \mathbf{w}_j^{(k)} \right)}} - \nabla{F_j{\left( \mathbf{w}_i^{(k)} \right)}} \right\|}^2}{m^2}}} \le \sum_{i=1}^m{\sum_{j=1}^m{\frac{L_j^2 {\left\| \mathbf{w}_j^{(k)} - \mathbf{w}_i^{(k)} \right\|}^2}m}}
        \\
        & \le L^2 \sum_{i=1}^m{\sum_{j=1}^m{\frac{2 {\left\| \mathbf{w}_j^{(k)} - \mathbf{\bar{w}}^{(k)} \right\|}^2 + 2 {\left\| \mathbf{w}_i^{(k)} - \mathbf{\bar{w}}^{(k)} \right\|}^2}m}} = 4 L^2 \sum_{i=1}^m{{\left\| \mathbf{w}_i^{(k)} - \mathbf{\bar{w}}^{(k)} \right\|}^2}
        \\
        & = 4 L^2 {\left\| \mathbf{W}^{(k)} - \mathbf{1}_m \mathbf{\bar{w}}^{(k)} \right\|}^2.
    \end{aligned}
\end{equation*}
where the Lipschitz constant $L$ of Assumption~\ref{assump:smooth_convex_graddiv}-\ref{assump:smoothness} was used. Therefore,
\begin{equation*}
    {\left\| \mathbf{\nabla}^{(k)} - \mathbf{1}_m \mathbf{\bar{\nabla}}^{(k)} \right\|}^2 \le 2 m \delta^2 + 8 L^2 {\left\| \mathbf{W}^{(k)} - \mathbf{1}_m \mathbf{\bar{w}}^{(k)} \right\|}^2.
\end{equation*}

\ref{lemma:kplus1:optim}
Next, we bound the difference of the average of local models Eq.~\eqref{eqn:wbarRecursive} from the global optimum $\mathbf{w}^\star$:
\begin{equation*}
    \begin{aligned}
		{\left\| \mathbf{\bar{w}}^{(k+1)} - \mathbf{w}^\star \right\|}^2 & = {\left\| \mathbf{\bar{w}}^{(k)} - \alpha^{(k)} \mathbf{\bar{g}}^{(k)} - \mathbf{w}^\star \right\|}^2
		\\
		& = {\left\| \mathbf{\bar{w}}^{(k)} - \alpha^{(k)} \mathbf{\bar{\nabla}}^{(k)} - \mathbf{w}^\star \right\|}^2 - 2 \alpha^{(k)} \left\langle \mathbf{\bar{w}}^{(k)} - \alpha^{(k)} \mathbf{\bar{\nabla}}^{(k)} - \mathbf{w}^\star, \mathbf{\bar{\epsilon}}^{(k)} \right\rangle + {\left( \alpha^{(k)} \right)}^2 {\left\| \mathbf{\bar{\epsilon}}^{(k)} \right\|}^2
		\\
		& \begin{aligned}
			\le \left( 1+H_1^{(k)} \right) {\left\| \mathbf{\bar{w}}^{(k)} - \alpha^{(k)} \nabla{F{\left( \mathbf{\bar{w}}^{(k)} \right)}} - \mathbf{w}^\star \right\|}^2 & + \left( 1+\frac1{H_1^{(k)}} \right) {\left( \alpha^{(k)} \right)}^2 {\left\| \nabla{F{\left( \mathbf{\bar{w}}^{(k)} \right)}} - \mathbf{\bar{\nabla}}^{(k)} \right\|}^2
			\\
			& - 2 \alpha^{(k)} \left\langle \mathbf{D^{(k)}}, \frac1m \mathbf{1}_m^T \mathbf{\epsilon}^{(k)} \right\rangle + {\left( \alpha^{(k)} \right)}^2 {\left\| \mathbf{\bar{\epsilon}}^{(k)} \right\|}^2
		\end{aligned}
		\\
		& \begin{aligned}
			\le \left( 1+H_1^{(k)} \right) {\left( 1 - \mu \alpha^{(k)} \right)}^2 {\left\| \mathbf{\bar{w}}^{(k)} - \mathbf{w}^\star \right\|}^2 & + \left( 1+\frac1{H_1^{(k)}} \right) \frac{{\left( \alpha^{(k)} \right)}^2 L^2}m {\left\| \mathbf{W}^{(k)} - \mathbf{1}_m \mathbf{\bar{w}}^{(k)} \right\|}^2
			\\
			& - 2 \alpha^{(k)} \left\langle \mathbf{D^{(k)}}, \frac1m \mathbf{1}_m^T \mathbf{\epsilon}^{(k)} \right\rangle + {\left( \alpha^{(k)} \right)}^2 {\left\| \mathbf{\bar{\epsilon}}^{(k)} \right\|}^2,
		\end{aligned}
	\end{aligned}
\end{equation*}
where the inequalities of Lemma~\ref{lemma:harnessing:ineqs} was used, and $\mathbf{D^{(k)}} \triangleq \mathbf{\bar{w}}^{(k)} - \alpha^{(k)} \mathbf{\bar{\nabla}}^{(k)} - \mathbf{w}^\star$. Also note that we can set $H_1^{(k)}$ to an arbitrary positive real value. Taking the expected value of this relation and setting $H_1^{(k)} = \mu \alpha^{(k)}$, while noting that $\mathbf{D^{(k)}}$ and $\epsilon^{(k)}$ are independent random variables since $\mathbf{\bar{w}}^{(k)}$ only depends on $\mathbf{\epsilon}^{(0)}$ to $\mathbf{\epsilon^{(k-1)}}$, we get

\begin{equation*}
    \begin{aligned}
        & \begin{aligned}
            \mathbb{E}{\left[ {\left\| \mathbf{\bar{w}}^{(k+1)} - \mathbf{w}^\star \right\|}^2 \right]} \le \left( 1 - \mu^2 {\left( \alpha^{(k)} \right)}^2 \right) \left( 1 - \mu \alpha^{(k)} \right) \mathbb{E}{\left[ {\left\| \mathbf{\bar{w}}^{(k)} - \mathbf{w}^\star \right\|}^2 \right]} & + \left( 1+\mu\alpha^{(k)} \right) \frac{\alpha^{(k)} L^2}{\mu m} \mathbb{E}{\left[ {\left\| \mathbf{W}^{(k)} - \mathbf{1}_m \mathbf{\bar{w}}^{(k)} \right\|}^2 \right]}
            \\
            & + \frac{{\left( \alpha^{(k)} \right)}^2 \sigma^2}m
        \end{aligned}
        \\
		& \le \left( 1 - \mu \alpha^{(k)} \right) \mathbb{E}{\left[ {\left\| \mathbf{\bar{w}}^{(k)} - \mathbf{w}^\star \right\|}^2 \right]} + \left( 1+\mu\alpha^{(k)} \right) \frac{\alpha^{(k)} L^2}{\mu m} \mathbb{E}{\left[ {\left\| \mathbf{W}^{(k)} - \mathbf{1}_m \mathbf{\bar{w}}^{(k)} \right\|}^2 \right]} + \frac{{\left( \alpha^{(k)} \right)}^2 \sigma^2}m,
	\end{aligned}
\end{equation*}
where the gradient approximation independence of Assumption~\ref{assump:graderror}-\ref{assump:graderror:indep}, and the result of Lemma~\ref{lemma:epsilonbounds} is used. Note that the reason we set $H_1^{(k)} = \mu \alpha^{(k)}$ was so that the scalar coefficient of $\mathbb{E}{[ {\| \mathbf{\bar{w}}^{(k)} - \mathbf{w}^\star \|}^2 ]}$ would become less that $1$, which is essential in our convergence analysis when proving Proposition~\ref{proposition:nonIncreasing}.

\ref{lemma:kplus1:consensus}
Finally, we bound the consensus error of the local models to their average value using Eqs.~\eqref{eqn:wRecursiveMatrix} and \eqref{eqn:wbarRecursive}.
\begin{equation} \label{eqn:wkplusone_bound}
	\begin{aligned}
		{\left\| \mathbf{W}^{(k+1)} - \mathbf{1}_m \mathbf{\bar{w}}^{(k+1)} \right\|}^2 & = {\left\| \mathbf{P}^{(k)} \mathbf{W}^{(k)} - \mathbf{1}_m \mathbf{\bar{w}}^{(k)} - \alpha^{(k)} \left( \mathbf{G}^{(k)} - \mathbf{1}_m \mathbf{\bar{g}}^{(k)} \right) \right\|}^2
		\\
		& \begin{aligned}
		    = \,\, & {\left\| \mathbf{P}^{(k)} \mathbf{W}^{(k)} - \mathbf{1}_m \mathbf{\bar{w}}^{(k)} - \alpha^{(k)} \left( \mathbf{\nabla}^{(k)} - \mathbf{1}_m \mathbf{\bar{\nabla}}^{(k)} \right) \right\|}^2
		    \\
		    & - 2 \alpha^{(k)} \left\langle \mathbf{P}^{(k)} \mathbf{W}^{(k)} - \mathbf{1}_m \mathbf{\bar{w}}^{(k)} - \alpha^{(k)} \left( \mathbf{\nabla}^{(k)} - \mathbf{1}_m \mathbf{\bar{\nabla}}^{(k)} \right), \mathbf{\epsilon}^{(k)} - \mathbf{1}_m \mathbf{\bar{\epsilon}}^{(k)} \right\rangle
		    \\
		    & + {\left( \alpha^{(k)} \right)}^2 {\left\| \mathbf{\epsilon}^{(k)} - \mathbf{1}_m \mathbf{\bar{\epsilon}}^{(k)} \right\|}^2
		\end{aligned}
		\\
		& \begin{aligned}
			\le \,\, & \left( 1+H_2^{(k)} \right) {\left\| \mathbf{P}^{(k)} \mathbf{W}^{(k)} - \mathbf{1}_m \mathbf{\bar{w}}^{(k)} \right\|}^2 + \left( 1+\frac1{H_2^{(k)}} \right) {\left( \alpha^{(k)} \right)}^2 {\left\| \mathbf{\nabla}^{(k)} - \mathbf{1}_m \mathbf{\bar{\nabla}}^{(k)} \right\|}^2
			\\
			& - 2 \alpha^{(k)} \left\langle \mathbf{D^{(k)}}, \left( \mathbf{I} - \frac1m \mathbf{1}_m \mathbf{1}_m^T \right) \mathbf{\epsilon}^{(k)} \right\rangle + {\left( \alpha^{(k)} \right)}^2 {\left\| \mathbf{\epsilon}^{(k)} - \mathbf{1}_m \bar{\mathbf{\epsilon}}^{(k)} \right\|}^2,
		\end{aligned}
	\end{aligned}
\end{equation}
where $\mathbf{D}^{(k)} \triangleq \mathbf{P}^{(k)} \mathbf{W}^{(k)} - \mathbf{1}_m \mathbf{\bar{w}}^{(k)} - \alpha^{(k)} \left( \mathbf{\nabla}^{(k)} - \mathbf{1}_m \mathbf{\bar{\nabla}}^{(k)} \right)$. Now, focusing on the first two terms in the bound, we write

\begin{equation*}
    \begin{aligned}
        & \left( 1+H_2^{(k)} \right) {\left\| \mathbf{P}^{(k)} \mathbf{W}^{(k)} - \mathbf{1}_m \mathbf{\bar{w}}^{(k)} \right\|}^2 + \left( 1+\frac1{H_2^{(k)}} \right) {\left( \alpha^{(k)} \right)}^2 {\left\| \mathbf{\nabla}^{(k)} - \mathbf{1}_m \mathbf{\bar{\nabla}}^{(k)} \right\|}^2
		\\
		& \begin{aligned}
	        \le \,\, & \left( 1+H_2^{(k)} \right) {\left( \rho^{(k)} \right)}^2 {\left\| \mathbf{W}^{(k)} - \mathbf{1}_m \mathbf{\bar{w}}^{(k)} \right\|}^2 + \left( 1+\frac1{H_2^{(k)}} \right) {\left( \alpha^{(k)} \right)}^2 \left[ 2 m \delta^2 + 8 L^2 {\left\| \mathbf{W}^{(k)} - \mathbf{1}_m \mathbf{\bar{w}}^{(k)} \right\|}^2 \right]
		\end{aligned}
        \\
	    & \le \left[ \left( 1+H_2^{(k)} \right) {\left( \rho^{(k)} \right)}^2 + 8 \left( 1+\frac1{H_2^{(k)}} \right) {\left( \alpha^{(k)} \right)}^2 L^2 \right] {\left\| \mathbf{W}^{(k)} - \mathbf{1}_m \mathbf{\bar{w}}^{(k)} \right\|}^2 + 2m \left( 1+\frac1{H_2^{(k)}} \right) {\left( \alpha^{(k)} \right)}^2 \delta^2,
	\end{aligned}
\end{equation*}
in which the results of Lemma~\ref{lemma:harnessing:spectral} and Part~\ref{lemma:kplus1:gradbound} of this lemma were used. Now, putting this inequality back in Eq.~\eqref{eqn:wkplusone_bound}, then taking its expected value with $H_2^{(k)} = \frac{2\sqrt{2} \alpha^{(k)} L}{\rho^{( k )}}$, noting that $\mathbf{D}^{(k)}$ and $\epsilon^{(k)}$ are independent random variables since $\mathbf{P}^{(k)} \mathbf{W}^{(k)} - \mathbf{1}_m \mathbf{\bar{w}}^{(k)}$ only depends on $\mathbf{\epsilon}^{(0)}$ to $\mathbf{\epsilon}^{(k-1)}$, we have
\begin{equation*}
    \mathbb{E}{\left[ {\left\| \mathbf{W}^{(k+1)} - \mathbf{1}_m \mathbf{\bar{w}}^{(k+1)} \right\|}^2 \right]} \le {\left( \rho^{(k)} + 2\sqrt{2} \alpha^{(k)} L \right)}^2 \mathbb{E}{\left[ {\left\| \mathbf{W}^{(k)} - \mathbf{1}_m \mathbf{\bar{w}}^{(k)} \right\|}^2 \right]} + m {\left( \alpha^{(k)} \right)}^2 \left[ 2 \frac{\rho^{(k)} + 2\sqrt{2} \alpha^{(k)} L}{2\sqrt{2} \alpha^{(k)} L} \delta^2 + \sigma^2 \right],
\end{equation*}
where the results of Lemma~\ref{lemma:epsilonbounds} was used. Note that $H_2^{(k)}$ was chosen such that $H_2^{(k)} = \argmin_{H_2^{(k)}}{[ ( 1+H_2^{(k)} ) {( \rho^{(k)} )}^2 + 8 ( 1+\frac1{H_2^{(k)}} ) {( \alpha^{(k)} )}^2 L^2 ]}$.

\section{Proof of Lemma~\ref{lemma:kplusB}} \label{appendix:lemma:kplusB}
\noindent We use Eqs.~\eqref{eqn:wExplicitMatrix} and \eqref{eqn:wbarExplicit} to bound the consensus error of local models from their mean value at iteration $k+B$, in which $B$ is the connectivity bound of Proposition~\ref{proposition:conn}:

\begin{equation*}
	\begin{aligned}
	    & \begin{aligned}
    		\Big\| \mathbf{W}^{(k+B)} - \mathbf{1}_m \mathbf{\bar{w}}^{(k+B)} \Big\|^2 = \Bigg\| \mathbf{P}^{(k+B-1:k)} \mathbf{W}^{(k)} - \mathbf{1}_m \mathbf{\bar{w}}^{(k)} & - \sum_{r=k+1}^{k+B-1}{\alpha^{(r-1)} \left( \mathbf{P}^{(k+B-1:r)} \mathbf{G}^{(r-1)} - \mathbf{1}_m \mathbf{\bar{g}}^{(r-1)} \right)}
    		\\
    		& - \alpha^{(k+B-1)} \left( \mathbf{G}^{(k+B-1)} - \mathbf{1}_m \mathbf{\bar{g}}^{(k+B-1)} \right) \Bigg\|^2
    	\end{aligned}
    	\\
    	& = {\left\| \mathbf{P}^{(k+B-1:k)} \mathbf{W}^{(k)} - \mathbf{1}_m \mathbf{\bar{w}}^{(k)} - \sum_{r=k+1}^{k+B}{\alpha^{(r-1)} \left( \mathbf{P}^{(k+B-1:r)} \mathbf{G}^{(r-1)} - \mathbf{1}_m \mathbf{\bar{g}}^{(r-1)} \right)} \right\|}^2
        \\
		& \begin{aligned}
            \,\, = \Bigg\| & \mathbf{P}^{(k+B-1:k)} \mathbf{W}^{(k)} - \mathbf{1}_m \mathbf{\bar{w}}^{(k)} - \sum_{r=k+1}^{k+B}{\alpha^{(r-1)} \left( \mathbf{P}^{(k+B-1:r)} \mathbf{\nabla}^{(r-1)} - \mathbf{1}_m \mathbf{\bar{\nabla}}^{(r-1)} \right)} \Bigg\|^2
		  \\
            & \begin{aligned}
                - 2 \Bigg\langle & \mathbf{P}^{(k+B-1:k)} \mathbf{W}^{(k)} - \mathbf{1}_m \mathbf{\bar{w}}^{(k)} - \sum_{r=k+1}^{k+B}{\alpha^{(r-1)} \left( \mathbf{P}^{(k+B-1:r)} \mathbf{\nabla}^{(r-1)} - \mathbf{1}_m \mathbf{\bar{\nabla}}^{(r-1)} \right)},
                \\
                & \sum_{r=k+1}^{k+B}{\alpha^{(r-1)} \left( \mathbf{P}^{(k+B-1:r)} \mathbf{\epsilon}^{(r-1)} - \mathbf{1}_m \mathbf{\bar{\epsilon}}^{(r-1)} \right)} \Bigg\rangle + {\left\| \sum_{r=k+1}^{k+B}{\alpha^{(r-1)} \left( \mathbf{P}^{(k+B-1:r)} \mathbf{\epsilon}^{(r-1)} - \mathbf{1}_m \mathbf{\bar{\epsilon}}^{(r-1)} \right)} \right\|}^2.
            \end{aligned}
		\end{aligned}
	\end{aligned}
\end{equation*}
Looking only at the inner product term in the above expression, we can write it equivalently as follows
\begin{equation*}
    \begin{aligned}
	    & -2 \left\langle \mathbf{D}^{(k)}, \sum_{r=k+1}^{k+B}{\alpha^{(r-1)} \left( \mathbf{P}^{(k+B-1:r)} - \frac1m \mathbf{1}_m \mathbf{1}_m^T \right) \mathbf{\epsilon}^{(r-1)}} \right\rangle
		\\
		& \begin{aligned}
		    \,\, + 2 \Bigg\langle \sum_{r=k+1}^{k+B}{\alpha^{(r-1)} \left( \mathbf{P}^{(k+B-1:r)} \mathbf{\nabla}^{(r-1)} - \mathbf{1}_m \mathbf{\bar{\nabla}}^{(r-1)} \right)}, \sum_{r=k+1}^{k+B}{\alpha^{(r-1)} \left( \mathbf{P}^{(k+B-1:r)} - \frac1m \mathbf{1}_m \mathbf{1}_m^T \right) \mathbf{\epsilon}^{(r-1)}} \Bigg\rangle
		\end{aligned}
		\\
		& \begin{aligned}
		    \le - 2 \left\langle \mathbf{D}^{(k)}, \sum_{r=k+1}^{k+B}{\alpha^{(r-1)} \left( \mathbf{P}^{(k+B-1:r)} - \frac1m \mathbf{1}_m \mathbf{1}_m^T \right) \mathbf{\epsilon}^{(r-1)}} \right\rangle & + {\left\| \sum_{r=k+1}^{k+B}{\alpha^{(r-1)} \left( \mathbf{P}^{(k+B-1:r)} \mathbf{\nabla}^{(r-1)} - \mathbf{1}_m \mathbf{\bar{\nabla}}^{(r-1)} \right)} \right\|}^2
		  \\
            & + {\left\| \sum_{r=k+1}^{k+B}{\alpha^{(r-1)} \left( \mathbf{P}^{(k+B-1:r)} - \frac1m \mathbf{1}_m \mathbf{1}_m^T \right) \mathbf{\epsilon}^{(r-1)}} \right\|}^2,
		\end{aligned}
    \end{aligned}
\end{equation*}
in which $\mathbf{D}^{(k)} \triangleq \mathbf{P}^{(k+B-1:k)} \mathbf{W}^{(k)} - \mathbf{1}_m \mathbf{\bar{w}}^{(k)}$. Going back to ${\left\| \mathbf{W}^{(k+B)} - \mathbf{1}_m \mathbf{\bar{w}}^{(k+B)} \right\|}^2$, we can bound it as
\begin{equation*}
    \begin{aligned}
	    \Big\| & \mathbf{W}^{(k+B)} - \mathbf{1}_m \mathbf{\bar{w}}^{(k+B)} \Big\|^2 \le \left( 1+H_3 \right) {\left\| \mathbf{P}^{(k+B-1:k)} \mathbf{W}^{(k)} - \mathbf{1}_m \mathbf{\bar{w}}^{(k)} \right\|}^2
		\\
		& + \left( 2+\frac1{H_3} \right) B \sum_{r=k+1}^{k+B} {\left( \alpha^{(r-1)} \right)}^2 {\left\| \mathbf{P}^{(k+B-1:r)} \mathbf{\nabla}^{(r-1)} - \mathbf{1}_m \mathbf{\bar{\nabla}}^{(r-1)} \right\|}^2
        \\
        & -2 \left\langle \mathbf{D}^{(k)}, \sum_{r=k+1}^{k+B}{\alpha^{(r-1)} \left( \mathbf{P}^{(k+B-1:r)} - \frac1m \mathbf{1}_m \mathbf{1}_m^T \right) \mathbf{\epsilon}^{(r-1)}} \right\rangle + 2B \sum_{r=k+1}^{k+B}{{\left( \alpha^{(r-1)} \right)}^2 {\left\| \mathbf{P}^{(k+B-1:r)} \mathbf{\epsilon}^{(r-1)} - \mathbf{1}_m \mathbf{\bar{\epsilon}}^{(r-1)}  \right\|}^2}
        \\
		& \begin{aligned}
		    \,\, \le \left( 1+H_3 \right) {\left( \rho^{(k+B-1:k)} \right)}^2 {\left\| \mathbf{W}^{(k)} - \mathbf{1}_m \mathbf{\bar{w}}^{(k)} \right\|}^2 & + \left( 2+\frac1{H_3} \right) B \sum_{r=k+1}^{k+B}{\left( \alpha^{(r-1)} \right)}^2 {\left( \rho^{(k+B-1:r)} \right)}^2 {\left\| \mathbf{\nabla}^{(r-1)} - \mathbf{1}_m \mathbf{\bar{\nabla}}^{(r-1)} \right\|}^2
		  \\
            & - 2 \sum_{r=k+1}^{k+B}{\alpha^{(r-1)} \left\langle \mathbf{D}^{(k)}, \left( \mathbf{P}^{(k+B-1:r)} - \frac1m \mathbf{1}_m \mathbf{1}_m^T \right) \mathbf{\epsilon}^{(r-1)} \right\rangle}
            \\
            & + 2B \sum_{r=k+1}^{k+B}{{\left( \alpha^{(r-1)} \right)}^2 {\left( \rho^{(k+B-1:r)} \right)}^2 {\left\| \mathbf{\epsilon}^{(r-1)} - \mathbf{1}_m \mathbf{\bar{\epsilon}}^{(r-1)}  \right\|}^2},
		\end{aligned}
	\end{aligned}
\end{equation*}
in which the results of Lemma~\ref{lemma:harnessing:spectral} is used. Next, we bound $\mathbb{E}{[ {\| \mathbf{\nabla}^{(r-1)} - \mathbf{1}_m \mathbf{\bar{\nabla}}^{(r-1)} \|}^2 ]}$ using Lemma~\ref{lemma:kplus1} and using recursion until we reach iteration $k$ in the right-hand side of the inequality as
\begin{equation*}
    \begin{aligned}
        & \mathbb{E}{\left[ {\left\| \mathbf{\nabla}^{(r-1)} - \mathbf{1}_m \mathbf{\bar{\nabla}}^{(r-1)} \right\|}^2 \right]} \le 2m \delta^2 + 8 L^2 \mathbb{E}{\left[ {\left\| \mathbf{W}^{(r-1)} - \mathbf{1}_m \mathbf{\bar{w}}^{(r-1)} \right\|}^2 \right]}
        \\
        & \begin{aligned}
            \le 2m \delta^2 + 8 L^2 \Bigg[ a_{21}^{(r-2:k)} \mathbb{E}{\left[ {\left\| \mathbf{\bar{w}}^{(k)} - \mathbf{w}^\star \right\|}^2 \right]} + a_{22}^{(r-2:k)} \mathbb{E}{\left[ {\left\| \mathbf{W}^{(k)} - \mathbf{1}_m \mathbf{\bar{w}}^{(k)} \right\|}^2 \right]} & + \sum_{l=k+1}^{r-2}{\left( a_{21}^{(r-2:l)} c_1^{(l-1)} + a_{22}^{(r-2:l)} c_2^{(l-1)} \right)}
            \\
            & + c_2^{(r-2)} \Bigg].
        \end{aligned}
    \end{aligned}
\end{equation*}
Combining the last two sets of inequalities and taking the expected value of ${\| \mathbf{W}^{(k+B)} - \mathbf{1}_m \mathbf{\bar{w}}^{(k+B)} \|}^2$, with $H_3 = \frac{1-\rho^2}{2\rho^2}$ and using the bound $\rho^{(k+B-1:k)} \le \rho$, it yields
\begin{equation*}
	\begin{aligned}
		\mathbb{E} \bigg[ \Big\| \mathbf{W}^{(k+B)} & - \mathbf{1}_m \mathbf{\bar{w}}^{(k+B)} \Big\|^2 \bigg] \le
        \\
        & \begin{aligned}
            \Bigg[ \frac{1+\rho^2}2 & + 16 \frac1{1-\rho^2} B L^2 \sum_{r=k+1}^{k+B} {\left( \alpha^{(r-1)} \right)}^2 {\left( \rho^{(k+B-1:r)} \right)}^2 a_{22}^{(r-2:k)} \Bigg] \mathbb{E}{\left[ {\left\| \mathbf{W}^{(k)} - \mathbf{1}_m \mathbf{\bar{w}}^{(k)} \right\|}^2 \right]}
    		\\
    		& + 4 \frac1{1-\rho^2} B \sum_{r=k+1}^{k+B} {\left( \alpha^{(r-1)} \right)}^2 {\left( \rho^{(k+B-1:r)} \right)}^2 \left( m \delta^2 + 4 L^2 \left[ \sum_{l=k+1}^{r-2}{a_{22}^{(r-2:l)} c_2^{(l-1)}} + c_2^{(r-2)} \right] \right)
            \\
            & + 2mB \sigma^2 \sum_{r=k+1}^{k+B}{{\left( \alpha^{(r-1)} \right)}^2 {\left( \rho^{(k+B-1:r)} \right)}^2}.
        \end{aligned}
	\end{aligned}
\end{equation*}
Note that $\mathbf{D}^{(k)}$ and $\epsilon^{(k)}$ are independent random variables, since $\mathbf{P}^{(k+B-1:k)} \mathbf{W}^{(k)} - \mathbf{1}_m \mathbf{\bar{w}}^{(k)}$ only depends on $\mathbf{\epsilon}^{(0)}$ to $\mathbf{\epsilon}^{(k-1)}$. \footnote{Although the transition matrices $\mathbf{P}^{(k)}$ are created according to the broadcast events, and the events in term depend on $\mathbf{w}^{(k)}$, here we neglect this fact for the sake of analysis and assume $\mathbf{P}^{(k)}$ are some known deterministic time-varying transition matrices. Therefore, they will be independent of $\mathbf{w}^{(k)}$, and consequently independent of $\mathbf{\epsilon}^{(k)}$. In other words, in $\mathbf{D}^{(k)} = \mathbf{P}^{(k+B-1:k)} \mathbf{W}^{(k)} - \mathbf{1}_m \mathbf{\bar{w}}^{(k)}$, only $\mathbf{W}^{(k)}$ and $\mathbf{\bar{w}}^{(k)}$ are random variables.}. Substituting $\rho^{(k+B-1:r)} < 1$ for $k+1 \le r \le k+B-1$ completes the proof. Note that $H_3 = \frac{1-\rho^2}{2\rho^2}$ was chosen such that $( 1+H_3 ) \rho^2 < 1$ would hold, which is necessary to prove Proposition~\ref{proposition:nonIncreasing}.

\section{Proof of Proposition~\ref{proposition:nonIncreasing}} \label{appendix:proposition:nonIncreasing}

We divide the proof into three steps: (i) obtaining the constraints on the step size, (ii) obtaining some bounds for the matrix entries of $\mathbf{\Psi}^{(k)}$, as defined in Lemma~\ref{lemma:kplusB} and Eq.~\eqref{eqn:psiValues}, and (iii) providing the final results.

\noindent (i) We obtain the conditions under which $\rho{( \mathbf{\Phi}^{(k)} )} < 1$. We have $\rho{( \mathbf{\Phi}^{(k)} )} = \max{\lbrace \lambda_1^{(k)}, \lambda_2^{(k)} \rbrace}$, where $\lambda_1^{(k)}$ and $\lambda_2^{(k)}$ are eigenvalues of $\mathbf{\Phi}^{(k)}$. Since $\phi_{21}^{(k)} = 0$ (see Lemma~\ref{lemma:kplusB}), we have
\begin{equation*}
    \det{\left( \lambda \mathbf{I} - \mathbf{\Phi}^{(k)} \right)} = \left( \lambda - \phi_{11}^{(k)} \right) \left( \lambda - \phi_{22}^{(k)} \right).
\end{equation*}
Consequently, we need $\phi_{11}^{(k)}, \phi_{22}^{(k)} < 1$. If $\alpha^{(0)} < \frac2{\mu + L}$, which implies $\alpha^{(k)} < \frac2{\mu + L}$, we have from Lemma~\ref{lemma:kplusB}, Eqs.~\eqref{eqn:phiValues} and \eqref{eqn:a_ks_Vals}, Lemmas~\ref{lemma:kplus1}-\ref{lemma:kplus1:optim} and \ref{lemma:kplus1}-\ref{lemma:kplus1:consensus} that
\begin{equation*}
    \phi_{11}^{(k)} = a_{11}^{(\left( k+1 \right) B - 1: kB)} = \prod_{r = kB}^{\left( k+1 \right) B - 1}{\left( 1 - \mu \alpha^{(r)} \right)} < 1 \quad \Rightarrow \quad \alpha^{(0)} < \frac1{\mu},
\end{equation*}
which trivially holds since $\mu < L$, and thus $\alpha^{(0)} < \frac2{\mu + L} < \frac2{2\mu} = \frac1{\mu}$. Next, for $\phi_{22}^{(k)} \le \frac{3 + \rho^2}4 < 1$, we must have:
\begin{equation*}
    \begin{aligned}
        & \phi_{22}^{(k)} = \frac{1+\rho^2}2 + 16 \frac1{1-\rho^2} B L^2 \sum_{r=kB+1}^{\left( k+1 \right) B} {\left( \alpha^{(r-1)} \right)}^2 a_{22}^{(r-2:k)} \quad  \Rightarrow \sum_{r=kB+1}^{\left( k+1 \right) B} {\left( \alpha^{(r-1)} \right)}^2 \prod_{l=kB}^{r-2}{{\left( 1 + 2\sqrt{2} \alpha^{(l)} L \right)}^2} \le \frac{{\left( 1-\rho^2 \right)}^2}{64 BL^2},
    \end{aligned}
\end{equation*}
in which Eq.~\eqref{eqn:a_ks_Vals}, Lemmas~\ref{lemma:kplus1}-\ref{lemma:kplus1:optim} and \ref{lemma:kplus1}-\ref{lemma:kplus1:consensus} were used. Next, to further analyze the above inequality, we introduce an extra constraint on the non-increasing step size, such that $\alpha^{(k)} \le \frac{\Gamma}{2\sqrt{2}L}$ where $\Gamma > 0$ can be any positive real value. Now we have
\begin{equation*}
    \begin{aligned}
        \sum_{r=kB+1}^{\left( k+1 \right) B} {\left( \alpha^{(r-1)} \right)}^2 \prod_{l=kB}^{r-2}{{\left( 1 + 2\sqrt{2} \alpha^{(l)} L \right)}^2} & \le \sum_{r=kB+1}^{\left( k+1 \right) B} {\left( \alpha^{(r-1)} \right)}^2 \prod_{l=kB}^{r-2}{{\left( 1 + \Gamma \right)}^2} \le \sum_{r=kB+1}^{\left( k+1 \right) B} {\left( \alpha^{(r-1)} \right)}^2 {\left( 1 + \Gamma \right)}^{2\left( r-1-kB \right)}
        \\
        & \le B {\left( \alpha^{(kB)} \right)}^2 {\left( 1 + \Gamma \right)}^{2\left( B-1 \right)}.
    \end{aligned}
\end{equation*}
As a result, we can obtain the second constraint on the upper bound for the step size as
\begin{equation*}
    B {\left( \alpha^{(kB)} \right)}^2 {\left( 1 + \Gamma \right)}^{2\left( B-1 \right)} \le \frac{{\left( 1-\rho^2 \right)}^2}{64 BL^2}, \quad \Rightarrow \alpha^{(0)} \le \frac{1-\rho^2}{8 BL {\left( 1+\Gamma \right)}^{B-1}}.
\end{equation*}
Thus, we can finally conclude the constraint
\begin{equation*}
    \alpha^{(0)} \le \min{\left\lbrace \frac{\Gamma}{2\sqrt{2}L}, \frac{1-\rho^2}{8 BL {\left( 1+\Gamma \right)}^{B-1}} \right\rbrace}.
\end{equation*}
Note that $\Gamma > 0$, and increasing it from $0$ to $\infty$ will increase the first constraint from $0$ to $\infty$, while decreasing the second one from $\frac{1-\rho^2}{8 BL}$ to $0$. Thus, we find the optimal value of $\Gamma$ next. We can say that there is a crossing point $\Gamma_0 > 0$ such that
\begin{equation*}
    \begin{cases}
        \frac{\Gamma}{2\sqrt{2}L} \le \frac{1-\rho^2}{8 BL {\left( 1+\Gamma \right)}^{B-1}} & 0 < \Gamma \le \Gamma_0
        \\
        \frac{1-\rho^2}{8 BL {\left( 1+\Gamma \right)}^{B-1}} \le \frac{\Gamma}{2\sqrt{2}L} & \Gamma \geq \Gamma_0
    \end{cases}.
\end{equation*}
So, by introducing an auxiliary function $H{( \cdot )}$ as follows,
\begin{equation*}
    H{\left( x \right)} = x{\left( 1+x \right)}^{B-1} - \frac{1-\rho^2}{2\sqrt{2}B},
\end{equation*}
we will have $H{\left( \Gamma_0 \right)} = 0$.

However, we are not interested in calculating the exact value of $\Gamma_0$, but instead we obtain the following bounds for it
\begin{equation*}
    H{\left( 0 \right)} = -\frac{1-\rho^2}{2\sqrt{2}B} < 0, \qquad H{\left( \frac{1-\rho^2}{2\sqrt{2}B} \right)} = \frac{1-\rho^2}{2\sqrt{2}B} \left[ {\left( 1 + \frac{1-\rho^2}{2\sqrt{2}B} \right)}^{B-1} - 1 \right] \geq 0, \qquad \Rightarrow \Gamma_0 \in \Bigg( 0, \frac{1-\rho^2}{2\sqrt{2}B} \Bigg].
\end{equation*}
Note that we have $\Gamma_1 = \frac{1-\rho^2}{2\sqrt{2}B} \le \frac1{2\sqrt{2}B} \le \frac1{2\sqrt{2}}$, and $\Gamma_1 \geq \Gamma_0$. Thus, setting $\Gamma = \Gamma_1$ will yield the following constraint
\begin{equation*}
    \alpha^{(0)} \le \frac{1-\rho^2}{8 BL {\left( 1+\Gamma_1 \right)}^{B-1}}.
\end{equation*}
In order to build an intuition to understand this bound, we can substitute the upper-bound of $\Gamma_1$ to get the following \footnote{Note that replacing the upper-bound of $\Gamma_1$ further restricts the step size. Moreover, see the explanation given for Proposition~\ref{proposition:nonIncreasing} on how $\alpha^{(0)}$ is actually related to $B$.}
\begin{equation*}
    \alpha^{(0)} \le \frac{1-\rho^2}{8 BL {\left( 1+\frac1{2\sqrt{2}} \right)}^{B-1}}.
\end{equation*}
Finally, we show that the above constraint is a tighter constraint than $\alpha^{(k)} < \frac2{\mu + L}$, which was the assumption we initially started from. Since $\mu < L$ for a $\mu$-convex and $L$-smooth objective function, we have
\begin{equation*}
    \frac{1-\rho^2}{8 BL {\left( 1 + \frac1{2\sqrt{2}} \right)}^{B-1}} \le \frac{1-\rho^2}{8 BL {\left( 1+\Gamma_1 \right)}^{B-1}} \le \frac{\Gamma_1}{2\sqrt{2}L} = \frac{\Gamma_1}{2\sqrt{2}} \frac2{2L} \le \frac18 \frac2{\mu+L} \le \frac2{\mu+L}.
\end{equation*}

(ii) Next, we derive the following bounds for Lemma \ref{lemma:kplus1}-\ref{lemma:kplus1:consensus}:
\begin{equation*}
    a_{22}^{(k)} \le {\left( 1+\Gamma \right)}^2, \qquad c_2^{(k)} = m {\left( \alpha^{(k)} \right)}^2 \left( \frac{1+\Gamma}{\sqrt{2} \alpha^{(k)} L} \delta^2 + \sigma^2 \right).
\end{equation*}
Second, for Lemma~\ref{lemma:kplusB} and Eq.~\eqref{eqn:phiValues}:
\begin{equation*}
    \phi_{11}^{(k)} \geq {\left( 1 - \mu \alpha^{(kB+1)} \right)}^B, \qquad \phi_{22}^{(k)} \le \frac{1+\rho^2}2 + 16 \frac1{1-\rho^2} B^2 L^2 {\left( \alpha^{(kB)} \right)}^2 {\left( 1+\Gamma \right)}^{2 \left( B-1 \right)}.
\end{equation*}
Similarly for~\eqref{eqn:a_ks_Vals}:
\begin{equation*}
    \begin{aligned}
		a_{12}^{(k+B-1:k)} & = a_{11}^{(k+B-1)} a_{12}^{(k+B-2:k)} + a_{12}^{(k+B-1)} a_{22}^{(k+B-2:k)} = a_{11}^{(k+B-1:k+1)} a_{12}^{(k)} + \sum_{r=k+2}^{k+B}{a_{11}^{(k+B-1:r)} a_{12}^{(r-1)} a_{22}^{(r-2:k)}}
		\\
		& = \sum_{r=k+1}^{k+B}{a_{11}^{(k+B-1:r)} a_{12}^{(r-1)} a_{22}^{(r-2:k)}} \le \sum_{r=k+1}^{k+B} {\left( a_{11}^{(k+B-1)} \right)}^{k+B-r} a_{12}^{(r-1)} \left( a_{22}^{(r-2)} \right)^{r-1-k} \le B a_{12}^{(k)} {\left( a_{22}^{(k)} \right)}^{B-1}
		\\
		& \le B \left( 1 + \mu \alpha^{(k)} \right) \frac{\alpha^{(k)} L^2}{\mu m} {\left( 1+\Gamma \right)}^{2 \left( B-1 \right)}.
	\end{aligned}
\end{equation*}
Finally, we find an upper-bound for Lemma~\ref{lemma:kplusB} and Eq.~\eqref{eqn:psiValues}. We have from part (i) that $0 < \phi_{11}^{(k)}, \phi_{22}^{(k)} < 1$. Next, we obtain bounds for $\psi_1^{(k)}$ and $\psi_2^{(k)}$, but note that we only need them to be bounded constants
\begin{equation*}
    \begin{aligned}
        & \begin{aligned}
            \psi_1^{(k)} \le {\left( \alpha^{(kB)} \right)}^2 \Bigg\lbrace \left( B-1 \right) \Bigg[ & \left( 1 - \mu \alpha^{(\left( k+1 \right) B - 1)} \right) \frac{\sigma^2}m
            \\
            & + \left( B-1 \right) \left( 1 + \mu \alpha^{(kB+1)} \right) \frac{\alpha^{(kB+1)} L^2}{\mu} {\left( 1+\Gamma \right)}^{2 \left( B-2 \right)} \left( \frac{1 + \Gamma}{\sqrt{2} \alpha^{(kB)} L} \delta^2 + \sigma^2 \right) \Bigg] + \frac{\sigma^2}m \Bigg\rbrace
        \end{aligned}
        \\
        & \le {\left( \alpha^{(kB)} \right)}^2 B \left\lbrace \frac{\sigma^2}m + \nu^{(kB)} \right\rbrace,
    \end{aligned}
\end{equation*}
\begin{equation*}
    \begin{aligned}
	    \psi_2^{(k)} & = 2B^2 {\left( \alpha^{(kB)} \right)}^2 \left\lbrace 2 \frac1{1-\rho^2} \left[ m \delta^2 + 4m L^2 {\left( \alpha^{(k)} \right)}^2 \left( \frac{1+\Gamma}{\sqrt{2} \alpha^{(k)} L} \delta^2 + \sigma^2 \right) \left( B-1 \right) {\left( 1+\Gamma \right)}^{2 \left( B-2 \right)} \right] + m \sigma^2 \right\rbrace
	    \\
	    & \le 2m {\left( \alpha^{(kB)} \right)}^2 B^2 \left\lbrace 2 \frac{\delta^2 + 2 \mu \alpha^{(kB)} \nu^{(kB)}}{1-\rho^2} + \sigma^2 \right\rbrace,
	\end{aligned}
\end{equation*}
where we have defined
\begin{equation*}
    \nu^{(k)} = \frac2{\mu} \left( B-1 \right) L^2 {\left( 1+\Gamma \right)}^{2 \left( B-1 \right)} \left( \frac{1+\Gamma}{\sqrt{2}L} \delta^2 + \alpha^{(k)} \sigma^2 \right).
\end{equation*}

(iii) It can be seen that as $k$ goes from $0$ to $\infty$, $\phi_{11}^{(k)}$ increases from ${(1 - \mu \alpha^{(1)})}^B$ to $1$, and $\phi_{22}^{(k)}$ decreases from $\frac{1+\rho^2}2 + 16 \frac1{1-\rho^2} B^2 L^2 {(\alpha^{(0)})}^2 {(1+\Gamma)}^{2 (B-1)}$ to $\frac{1+\rho^2}2$. Therefore, since $\frac{1+\rho^2}2 < 1$, there is an iteration $K \geq 0$ where the inequality $\phi_{11}^{(K)} \geq \phi_{22}^{(K)}$ holds for the first time, i.e., $\phi_{11}^{(k)} \le \phi_{22}^{(k)}$ for all $0 \le k < K$. Consequently, we have
\begin{equation*}
    \rho{\left( \mathbf{\Phi}^{(k)} \right)} =
    \begin{cases}
        \phi_{22}^{(k)} & 0 \le k < K
        \\
        \phi_{11}^{(k)} & k \geq K
    \end{cases}.
\end{equation*}
Note that $K = 0$ is also possible, which happens if ${(1 - \mu \alpha^{(1)})}^B \ge \frac{1+\rho^2}2 + 16 \frac1{1-\rho^2} B^2 L^2 {(\alpha^{(0)})}^2 {(1+\Gamma)}^{2 (B-1)}$. As a result, we write
\begin{equation*}
	\begin{aligned}
		\mathbf{\Xi}^{(kB)} & \le \mathbf{\Phi}^{(k-1:K)} \mathbf{\Phi}^{(K-1:0)} \mathbf{\Xi}^{(0)} + \sum_{r=1}^{k-1}{\mathbf{\Phi}^{(k-1:K)} \mathbf{\Phi}^{(K-1:r)} \mathbf{\Psi}^{(r-1)}} + \mathbf{\Psi}^{(k-1)}
		\\
		& \le \mathcal{O}{\left( \phi_{11}^{(k-1:K)} \right)} \mathcal{O}{\left( \phi_{22}^{(K-1:0)} \right)} \mathbf{\Xi}^{(0)} + \mathcal{O}{\left( \phi_{11}^{(k-1:K)} \right)} \sum_{r=1}^{K-1}{\mathcal{O}{\left( \phi_{22}^{(K-1:r)} \right)} \mathbf{\Psi}^{(r-1)}} + \sum_{r=K}^{k-1}{\mathcal{O}{\left( \phi_{11}^{(k-1:r)} \right)} \mathbf{\Psi}^{(r-1)}} + \mathbf{\Psi}^{(k-1)}.
	\end{aligned}
\end{equation*}

\section{Proof of Theorem~\ref{theorem:constant}} \label{appendix:theorem:constant}

\noindent If a constant step size $\alpha > 0$ is used, i.e., $\alpha^{(k)} = \alpha$ for all $k \geq 0$, the values in Lemmas~\ref{lemma:kplus1}-\ref{lemma:kplus1:optim} and \ref{lemma:kplus1}-\ref{lemma:kplus1:consensus}, Eq.~\eqref{eqn:a_ks_Vals}, Lemma~\ref{lemma:kplusB}, Eqs.~\eqref{eqn:phiValues}, \eqref{eqn:psiValues} and \eqref{eqn:phi_ks_Vals} will be time-invariant. Therefore, the inequalities in Eq.~\eqref{eqn:PhikPsikExplicit} become
\begin{equation}
	\mathbf{\Xi}^{((k+1) B)} \le \mathbf{\Phi}^{k-s+1} \mathbf{\Xi}^{(s)} + \left( \sum_{r=s+1}^{k}{\mathbf{\Phi}^{k-r+1}} + 1 \right) \mathbf{\Psi}.
\end{equation}
Therefore, Eq.~\eqref{eqn:matEntriesPsiDiminishing} in the results of Proposition~\ref{proposition:nonIncreasing} imply that the following condition must hold to have $\rho{( \mathbf{\Phi} )} < 1$:
\begin{equation*}
    \alpha \le \frac{1-\rho^2}{8 BL {\left( 1+\Gamma_1 \right)}^{B-1}}.
\end{equation*}

Finally, we employ~\eqref{eqn:matEntriesPsiDiminishing} to obtain the bounds for $\psi_1$ and $\psi_2$ as:
\begin{equation} \label{eqn:matEntriesConstant}
    \begin{gathered}
        \psi_1 \le \alpha^2 B \left\lbrace \frac{\sigma^2}m + \nu \right\rbrace, \qquad \psi_2 \le 2m \alpha^2 B^2 \left\lbrace 2 \frac{\delta^2 + 2 \mu \alpha \nu}{1-\rho^2} + \sigma^2 \right\rbrace,
        \\
        \nu = \frac2{\mu} \left( B-1 \right) L^2 {\left( 1+\Gamma \right)}^{2 \left( B-1 \right)} \left( \frac{1+\Gamma}{\sqrt{2} L} \delta^2 + \alpha \sigma^2 \right).
    \end{gathered}
\end{equation}

\section{Proof of Theorem~\ref{theorem:diminishing}} \label{appendix:theorem:diminishing}
\noindent Noting that $\phi_{11}^{(k:s)}$ and $\phi_{22}^{(k:s)}$ are used in the results of Proposition~\ref{proposition:nonIncreasing}, first we derive bounds for these quantities separately using relations Eq.~\eqref{eqn:phi_ks_Vals}, Lemma~\ref{lemma:kplusB}, Eqs.~\eqref{eqn:phiValues} and \eqref{eqn:a_ks_Vals} and Lemmas~\ref{lemma:kplus1}-\ref{lemma:kplus1:optim} and \ref{lemma:kplus1}-\ref{lemma:kplus1:consensus} as follows:
\begin{equation} \label{eqn:phiBounds}
    \phi_{11}^{(k:s)} = \prod_{r=sB}^{\left( k+1 \right) B - 1}{\left( 1 - \mu \alpha^{(r)} \right)} \le \frac1{\mu \sum_{r=sB}^{\left( k+1 \right) B - 1}{\alpha^{(r)}}}, \qquad \phi_{22}^{(k:s)} \le \prod_{r=s}^k{\frac{3 + \rho^2}4} = {\left( \frac{3 + \rho^2}4 \right)}^{k-s+1},
\end{equation}
where Lemma~\ref{lemma:bernoulli} was used to prove the first line, and by satisfying the constraint that $\alpha^{(0)} \le \frac{1-\rho^2}{8 BL {( 1+\Gamma_1 )}^{B-1}}$ where $\Gamma_1 = \frac{1-\rho^2}{2\sqrt{2}B}$, the results of Appendix~\ref{appendix:proposition:nonIncreasing} were used to prove the second line.

The bound in Proposition~\ref{proposition:nonIncreasing} consists of four terms: (i) $\mathcal{O}{( \phi_{11}^{(k-1:K)} )} \mathcal{O}{( \phi_{22}^{(K-1:0)} )}$ and (ii) $\mathcal{O}{( \phi_{11}^{(k-1:K)} )} \sum_{r=1}^{K-1}{\mathcal{O}{( \phi_{22}^{(K-1:r)} )} \Psi^{(r-1)}}$, both of which are the product of a constant and a diminishing term, and thus will converge to $0$ as $k \to \infty$; (iii) $\sum_{r=K}^{k-1}{\mathcal{O}{( \phi_{11}^{(k-1:r)} )} \mathbf{\Psi}^{(r-1)}}$ which is the product of an increasing and a decreasing term, and (iv) $\mathbf{\Psi}^{(k-1)}$ which diminishes over time. Terms (i) and (ii) can be easily determined using the values of Eq.~\eqref{eqn:phiBounds}, and (iv) is obtained via Eq.~\eqref{eqn:matEntriesPsiDiminishing}, but bounding (iii) needs one extra step as follows
\begin{equation} \label{eqn:phiPsiTerm}
    \sum_{r=K}^{k-1} {\phi_{11}^{(k-1:r)} \mathbf{\Psi}^{(r-1)}} \le \sum_{r=K}^{k-1}{\frac{{\left( \alpha^{(\left( r-1 \right) B)} \right)}^2 B}{\mu \sum_{l=rB}^{kB-1}{\alpha^{(l)}}}
    \begin{bmatrix}
        \left\lbrace \frac{\sigma^2}m + \nu^{(0)} \right\rbrace
        \\
        2m B \left\lbrace 2 \frac{\delta^2 + 2 \mu \alpha^{(0)} \nu^{(0)}}{1-\rho^2} + \sigma^2 \right\rbrace
    \end{bmatrix}},
\end{equation}
where we have defined
\begin{equation*}
    \nu^{(k)} = \frac2{\mu} \left( B-1 \right) L^2 {\left( 1+\Gamma_1 \right)}^{2 \left( B-1 \right)} \left( \frac{1+\Gamma_1}{\sqrt{2}L} \delta^2 + \alpha^{(k)} \sigma^2 \right).
\end{equation*}

Noting the summations in Eqs.~\eqref{eqn:phiBounds} and \eqref{eqn:phiPsiTerm}, we first provide the following supplementary lemma for their values, substituting the diminishing step size with the specific choice of $\alpha^{(k)} = \frac{\alpha^{(0)}}{{( 1 + k / \eta )}^{\theta}}$ with $\theta = 0.5$.

\begin{lemma} \label{lemma:summations}
    If the diminishing step size is chosen as $\alpha^{(k)} = \frac{\alpha^{(0)}}{ \sqrt{1 + k / \eta}}$, the following inequalities hold
    \begin{enumerate}[label=(\alph*)]
        \item
        \begin{equation*}
            \sum_{r=sB}^{\left( k+1 \right) B -  1}{\alpha^{(r)}} \geq 2 \alpha^{(0)} \left( \sqrt{1 + \frac{k+1}{\eta / B}} - \sqrt{1 + \frac{s}{\eta / B}} \right).
        \end{equation*} \label{lemma:summations:sum}
        
        \item
        \begin{equation*}
            \sum_{r=s}^k{\frac{{\left( \alpha^{(\left( r-1 \right) B)} \right)}^2}{\sum_{l=rB}^{\left( k+1 \right) B - 1}{\alpha^{(l)}}}} \le \frac{\frac{\alpha^{(0)}}2 \left( 1+\frac1{\eta / B} \right)}{ \sqrt{1 + \frac{k+1}{\eta / B}} - \sqrt{1 + \frac{s}{\eta / B}} } + \frac{\alpha^{(0)} \left( 1+\eta / B \right)}{\sqrt{1 + \frac{k+1}{\eta / B}}} \left[ \ln{\sqrt{1 + \frac{k+1}{\eta / B}}} + \ln{\frac1{\sqrt{1+\frac1{\eta / B+k}} - 1}} \right].
        \end{equation*} \label{lemma:summations:sumfracsum}
    \end{enumerate}
    
    \begin{proof}
        \ref{lemma:summations:sum} We can find a lower-bound for the summation by changing it to an integral as
        \begin{equation*}
            \sum_{r=sB}^{\left( k+1 \right) B -  1}{\alpha^{(r)}} \geq \int_{sB}^{\left( k+1 \right) B}{\alpha^{(x)} \, dx} = \int_{sB}^{\left( k+1 \right) B}{\frac{\alpha^{(0)}}{\sqrt{1+ x / \eta}} \, dx} = 2 \alpha^{(0)} \left( \sqrt{1 + \frac{k+1}{\eta / B}} - \sqrt{1 + \frac{s}{\eta / B}} \right).
        \end{equation*}
        
        \ref{lemma:summations:sumfracsum} Similarly, we can find an upper-bound for the summation by changing it to an integral. But first we substitute the bound obtained in Part~\ref{lemma:summations:sum} of this lemma to get
        
        \begin{equation*}
            \begin{aligned}
                \sum_{r=s}^k{\frac{{\left( \alpha^{(\left( r-1 \right) B)} \right)}^2}{\sum_{l=rB}^{\left( k+1 \right) B - 1}{\alpha^{(l)}}}} & \le \sum_{r=s}^k{\frac{\alpha^{(0)}}{2 \left( 1 + \frac{r-1}{\eta / B} \right) \left( \sqrt{1 + \frac{k+1}{\eta / B}} - \sqrt{1 + \frac{r}{\eta / B}} \right)}} \le \sum_{r=s}^k{\frac{\frac{\alpha^{(0)}}2 \left( 1 + \frac1{\eta / B} \right)}{\left( 1 + \frac{r}{\eta / B} \right) \left( \sqrt{1 + \frac{k+1}{\eta / B}} - \sqrt{1 + \frac{r}{\eta / B}} \right)}}
                \\
                & \le \frac{\frac{\alpha^{(0)}}2 \left( 1 + \frac1{\eta / B} \right)}{\left( 1 + \frac{s}{\eta / B} \right) \left( \sqrt{1 + \frac{k+1}{\eta / B}} - \sqrt{1 + \frac{s}{\eta / B}} \right)} + \int_{s}^k{\frac{\frac{\alpha^{(0)}}2 \left( 1 + \frac1{\eta / B} \right) \, dx}{\left( 1 + \frac{x}{\eta / B} \right) \left( \sqrt{1 + \frac{k+1}{\eta / B}} - \sqrt{1 + \frac{x}{\eta / B}} \right)}}.
            \end{aligned}
        \end{equation*}
        Focusing only on the integral term in the above expression without its constant coefficient, and changing the variable to $u{( x )} = \sqrt{1 + \frac{x}{\eta / B}}$, the integral becomes equal to
        \begin{equation*}
            \begin{aligned}
                \int_{s}^k{\frac{2 \left( \eta / B \right) \, d{u{\left( x \right)}}}{u{\left( x \right)} \left[ u{\left( k+1 \right)} - u{\left( x \right)} \right]}} & = \frac{2 \left( \eta / B \right)}{u{\left( k+1 \right)}} \left[ \int_{s}^k{\frac{d{u{\left( x \right)}}}{u{\left( x \right)}}} + \int_{s}^k{\frac{d{u{\left( x \right)}}}{u{\left( k+1 \right)} - u{\left( x \right)}}}\right] = \frac{2 \left( \eta / B \right)}{u{\left( k+1 \right)}} \left[ \ln{\frac{u{\left( k \right)}}{u{\left( s \right)}}} + \ln{\frac{u{\left( k+1 \right)} - u{\left( s \right)}}{u{\left( k+1 \right)} - u{\left( k \right)}}} \right]
                \\
                & \le \frac{2 \left( \eta / B \right)}{u{\left( k+1 \right)}}\ln{\frac{u{\left( k \right)} u{\left( k+1 \right)}}{u{\left( k+1 \right)} - u{\left( k \right)}}} = \frac{2 \left( \eta / B \right)}{u{\left( k+1 \right)}} \left[ \ln{u{\left( k+1 \right)}} + \ln{\frac1{\frac{u{\left( k+1 \right)}}{u{\left( k \right)}} - 1}} \right]
                \\
                & = \frac{2 \left( \eta / B \right)}{\sqrt{1 + \frac{k+1}{\eta / B}}} \left[ \ln{\sqrt{1 + \frac{k+1}{\eta / B}}} + \ln{\frac1{\sqrt{1+\frac1{\eta / B+k}} - 1}} \right].
            \end{aligned}
        \end{equation*}
        Putting all the terms together proves the statement of the lemma.
    \end{proof}
\end{lemma}

Finally, we substitute the results of Lemma~\ref{lemma:summations} and the bounds of Eq.~\eqref{eqn:phiBounds} accordingly in Proposition~\ref{proposition:nonIncreasing}, to conclude the proof of Theorem~\ref{theorem:diminishing}. We have

\begin{equation}
    \begin{aligned}
    	& \begin{aligned}
    	    \mathbf{\Xi}^{(kB)} \le \mathcal{O}{\left( \phi_{11}^{(k-1:K)} \right)} \mathcal{O}{\left( \phi_{22}^{(K-1:0)} \right)} \mathbf{\Xi}^{(0)} + \mathcal{O}{\left( \phi_{11}^{(k-1:K)} \right)} \sum_{r=1}^{K-1}{\mathcal{O}{\left( \phi_{22}^{(K-1:r)} \right)} \mathbf{\Psi}^{(r-1)}} & + \sum_{r=K}^{k-1}{\mathcal{O}{\left( \phi_{11}^{(k-1:r)} \right)} \mathbf{\Psi}^{(r-1)}}
            \\
            & + \mathbf{\Psi}^{(k-1)}
    	\end{aligned}
    	\\
    	& \begin{aligned}
    	    \le \frac1{2 \mu \alpha^{(0)}} \mathcal{O}{\left( \frac1{\sqrt{k}} \right)} & {\left( \frac{3 + \rho^2}4 \right)}^{K} \mathbf{\Xi}^{(0)}
            \\
            & + \left\lbrace \frac{K-1}{2 \mu \alpha^{(0)}} \mathcal{O}{\left( \frac1{\sqrt{k}} \right)} \left( \frac{1 + \rho^2}2 \right) + \frac{\alpha^{(0)}}2 \mathcal{O}{\left( \frac{\ln{k}}{\sqrt{k}} \right)} + {(\alpha^{(0)})}^2 \mathcal{O}{\left( \frac1k \right)} \right\rbrace B
            \begin{bmatrix}
                \widehat{\psi}_1
                \\
                \widehat{\psi}_2
            \end{bmatrix},
    	\end{aligned}
    \end{aligned}
\end{equation}
where
\begin{equation*}
    \widehat{\psi}_1 = \frac{\sigma^2}m + \nu^{(0)}, \quad \widehat{\psi}_2 = 2m B \left\lbrace 2 \frac{\delta^2 + 2 \mu \alpha^{(0)} \nu^{(0)}}{1-\rho^2} + \sigma^2 \right\rbrace, \quad \nu^{(k)} = \frac2{\mu} \left( B-1 \right) L^2 {\left( 1+\Gamma_1 \right)}^{2 \left( B-1 \right)} \left( \frac{1+\Gamma_1}{\sqrt{2}L} \delta^2 + \alpha^{(k)} \sigma^2 \right).
\end{equation*}
Note that $\ln{\frac1{\sqrt{1+\frac1{\eta / B+k}} - 1}} \le \mathcal{O}{\left( \ln{k} \right)}$.

\section{Convergence Results for the General Diminishing Step Size Policy} \label{appendix:generalTheta}
We mentioned in Assumption~\ref{assump:stepsizes}-\ref{assump:stepsizes:diminishing} that choosing a decreasing step size policy in the form of $\alpha^{(k)} = \frac{\alpha^{(0)}}{{(1+k / \eta)}^\theta}$ satisfies that assumption for $\theta \in (0.5, 1]$. We then provided the convergence results for the case of setting $\theta = 0.5$ in Theorem~\ref{theorem:diminishing}, which gives us a $\frac{\ln{k}}{\sqrt{k}}$ convergence rate. In this appendix, we touch on similar convergence results for all values of $\theta$.

We first provide a lemma similar to Lemma~\ref{lemma:summations}, but for all values of $\theta$. The main difference in the following lemma is that the lower-bound of and the upper-bound of Lemma~\ref{lemma:summations}-\ref{lemma:summations:sum} and \ref{lemma:summations}-\ref{lemma:summations:sumfracsum}, respectively, will be different.

\begin{lemma} \label{lemma:generalTheta}
    If the diminishing step size is chosen as $\alpha^{(k)} = \frac{\alpha^{(0)}}{{(1 + k / \eta)}^\theta}$ with $\theta = \frac{Q}{Q+1}$ for all integers $Q \geq 2$, the following inequalities hold
    \begin{enumerate}[label=(\alph*)]
        \item
        \begin{equation*}
            \sum_{r=sB}^{\left( k+1 \right) B -  1}{\alpha^{(r)}} \geq \frac{\alpha^{(0)}}{1-\theta} \left( {\left( 1 + \frac{k+1}{\eta / B} \right)}^{1-\theta} - {\left( 1 + \frac{s}{\eta / B} \right)}^{1-\theta} \right),
        \end{equation*} \label{lemma:generalTheta:sum}
        
        \item
        \begin{equation*}
            \begin{aligned}
                \sum_{r=s}^k{\frac{{\left( \alpha^{(\left( r-1 \right) B)} \right)}^2}{\sum_{l=rB}^{\left( k+1 \right) B - 1}{\alpha^{(l)}}}} \le \,\, & \frac{{\left( 1+\frac{1}{\eta / B} \right)}^{2\theta} \alpha^{(0)} \left( 1-\theta \right)}{{\left( 1+\frac{k+1}{\eta / B} \right)}^{1-\theta} - {\left( 1+\frac{s}{\eta / B} \right)}^{1-\theta}}
                \\
                & + \frac{\alpha^{(0)} \left(\eta / B\right) {\left( 1+\frac1{\eta / B} \right)}^{2\theta}}{{\left( 1+\frac{k+1}{\eta / B} \right)}^\theta} \left[ \ln{{\left( 1+\frac{k+1}{\eta / B} \right)}^{1-\theta}} - \ln{\frac1{{\left(1 + \frac1{\eta / B + k} \right)}^{1-\theta} - 1}} \right]
                \\
                & + \sum_{q=2}^{Q}{\frac{\alpha^{(0)} \left(\eta / B\right) {\left( 1+\frac1{\eta / B} \right)}^{2\theta}}{q-1} \frac1{{\left( 1+\frac{k+1}{\eta / B} \right)}^{1 - \frac{q}{Q+1}}}}.
            \end{aligned}
        \end{equation*} \label{lemma:generalTheta:sumfracsum}
    \end{enumerate}
    
    \begin{proof}
        \ref{lemma:generalTheta:sum} The proof is similar to the proof of Lemma~\ref{lemma:summations}-\ref{lemma:summations:sum}, replacing general $\theta$, where $0.5 < \theta < 1$.
        
        \ref{lemma:generalTheta:sumfracsum}
        The proof is similar to the proof of Lemma~\ref{lemma:summations}-\ref{lemma:summations:sumfracsum}, replacing general $\theta$, where $0.5 < \theta < 1$. The major difference here is the integral part $\int_{s}^k{\frac{\left( 1-\theta \right) \alpha^{(0)} \, dx}{{\left( 1 + \frac{x}{\eta / B} \right)}^{2\theta} \left( {\left( 1 + \frac{k+1}{\eta / B} \right)}^{1-\theta} - {\left( 1 + \frac{x}{\eta / B} \right)}^{1-\theta} \right)}}$, as we will elaborate next. Using a change of variable as $u{(x)} = {(1 + \frac{x}{\eta / B})}^{1-\theta} = {(1 + \frac{x}{\eta / B})}^{\frac1{Q+1}}$, the integral (not including its constant coefficients) becomes equal to
        \begin{equation*}
            \begin{aligned}
                & \int_{s}^k{\frac{\frac{\eta / B}{1-\theta} \, d{u{\left( x \right)}}}{{u{\left( x \right)}}^Q \left[ u{\left( k+1 \right)} - u{\left( x \right)} \right]}} = \frac{\eta / B}{1-\theta} \int_s^k{\left[ \sum_{q=1}^{Q}{\frac{\frac1{u{\left( k+1 \right)}^{Q+1-q}}}{u{\left( x \right)}^{q}}} + \frac{\frac1{u{\left( k+1 \right)}^{Q}}}{u{\left( k+1 \right)} - u{\left( x \right)}} \right] \, d{u{\left( x \right)}}}
                \\
                & = \frac{\eta / B}{1-\theta} \left[ \frac{\ln{\left( \frac{u{\left( k \right)}}{u{\left( s \right)}} \right)}}{u{\left( k+1 \right)}^{Q}} + \sum_{q=2}^{Q}{\frac{\frac1{u{\left( s \right)}^{q-1}} - \frac1{u{\left( k \right)}^{q-1}}}{\left( q-1 \right) u{\left( k+1 \right)}^{Q+1-q}}} + \frac{\ln{\left( \frac{u{\left( k+1 \right)} - u{\left( s \right)}}{u{\left( k+1 \right)} - u{\left( k \right)}} \right)}}{u{\left( k+1 \right)}^{Q}} \right]
                \\
                & \le \frac{\eta / B}{1-\theta} \left[ \frac{\ln{\left( \frac{u{\left( k \right)} u{\left( k+1 \right)}}{u{\left( k+1 \right)} - u{\left( k \right)}} \right)}}{u{\left( k+1 \right)}^{Q}} + \sum_{q=2}^{Q}{\frac1{\left( q-1 \right) u{\left( k+1 \right)}^{Q+1-q}}} \right]
                \\
                & = \frac{\frac{\eta / B}{1-\theta}}{u{\left( k+1 \right)}^{Q}} \left[ \ln{u{\left( k+1 \right)}} + \ln{\frac1{\frac{u{\left( k+1 \right)}}{u{\left( k \right)}} - 1}} \right] + \sum_{q=2}^{Q}{\frac{\frac{\eta / B}{1-\theta}}{q-1} \frac1{u{\left( k+1 \right)}^{Q+1-q}}}.
            \end{aligned}
        \end{equation*}
    \end{proof}
\end{lemma}
Note that although we started this appendix trying to prove the results for any real value of $\theta \in (0.5, 1]$, where $\theta$ is constant inside the step size policy of $\alpha^{(k)} = \frac{\alpha^{(0)}}{{\left( 1 + \frac{k}{\eta / B} \right)}^\theta}$, in Lemma~\ref{lemma:generalTheta} we were able to derive an upper bound in part~\ref{lemma:generalTheta:sumfracsum} only for the choices of $\theta = \frac{Q}{Q+1}$, where $Q \geq 1$ is an integer \footnote{Note that the case $Q=1$ indicates $\theta = 0.5$, which we studied separately in Lemma~\ref{lemma:summations}}.

Next, we provide another supplementary lemma for the case $\theta = 1$.

\begin{lemma} \label{lemma:theta1}
    If the diminishing step size is chosen as $\alpha^{(k)} = \frac{\alpha^{(0)}}{1 + k / \eta}$, the following inequalities hold
    \begin{enumerate}[label=(\alph*)]
        \item
        \begin{equation*}
            \sum_{r=sB}^{\left( k+1 \right) B -  1}{\alpha^{(r)}} \geq \alpha^{(0)} \ln{\left( \frac{1 + \frac{k+1}{\eta / B}}{1 + \frac{s}{\eta / B}} \right)},
        \end{equation*} \label{lemma:theta1:sum}
        
        \item
        \begin{equation*}
            \sum_{r=s}^k{\frac{{\left( \alpha^{(\left( r-1 \right) B)} \right)}^2}{\sum_{l=rB}^{\left( k+1 \right) B - 1}{\alpha^{(l)}}}} \le \frac{{\left( 1+\frac{1}{\eta / B} \right)}^2 \alpha^{(0)}}{\ln{\left( \frac{1+\frac{k+1}{\eta / B}}{1+\frac{s}{\eta / B}} \right)}} + \frac{\alpha^{(0)} \left( 1+\eta / B \right) \left( 1+\frac1{\eta / B} \right)}{1 + \frac{k+1}{\eta / B}} \left[ \li{\left( 1+\frac{k+1}{\eta / B} \right)} + \left\lvert \li{\left( 1+\frac1{\eta / B + k} \right)} \right\rvert \right].
        \end{equation*} \label{lemma:theta1:sumfracsum}
    \end{enumerate}
    
    \begin{proof}
        \ref{lemma:theta1:sum}
        The proof is similar to the proof of Lemma~\ref{lemma:summations}-\ref{lemma:summations:sum}, replacing $\theta = 1$.
        
        \ref{lemma:theta1:sumfracsum}
        The proof is similar to the proof of Lemma~\ref{lemma:summations}-\ref{lemma:summations:sumfracsum}, replacing $\theta = 1$. The major difference here is the integral part $\int_{s}^k{\frac{\alpha^{(0)} \, dx}{{\left( 1 + \frac{x}{\eta / B} \right)}^2 \ln{\left( \frac{1 + \frac{k+1}{\eta / B}}{1 + \frac{s}{\eta / B}} \right)}}}$, as we will elaborate next. Using a change of variable as $u{(x)} = \frac{1 + \frac{k+1}{\eta / B}}{1 + \frac{x}{\eta / B}}$, the integral (not including its constant coefficients) becomes equal to
        \begin{equation*}
            \int_{s}^k{\frac{ \frac{-\eta / B}{1+\frac{k+1}{\eta / B}} \, d{u{\left( x \right)}}}{\ln{\left( u{\left( x \right)} \right)}}} = \frac{\eta / B}{1+\frac{k+1}{\eta / B}} \left[ \li{\left( u{\left( s \right)} \right)} - \li{\left( u{\left( k \right)} \right)} \right] \le \frac{\eta / B}{1+\frac{k+1}{\eta / B}} \left[ \li{\left( u{\left( s \right)} \right)} + \left\lvert \li{\left( u{\left( k \right)} \right)} \right\rvert \right],
        \end{equation*}
        in which $\li{(x)} = \int_0^x{\frac{dt}{\ln{(t)}}}$ is the logarithmic integral function.
    \end{proof}
\end{lemma}
In order to obtain an expression for the upper-bound in Lemma~\ref{lemma:theta1}-\ref{lemma:theta1:sumfracsum} which is easier to understand, we look at the properties of the logarithmic integral function. We have
\begin{equation*}
    \begin{cases}
        \li(x) \le 0 & x \in [0,1) \cup (1, \mu]
        \\
        \li(x) \ge 0 & x \ge \mu
    \end{cases},
\end{equation*}
in which $\mu \approx 1.45$ is the zero-crossing point of $\li(x)$, and is known as the Ramanujan-Soldner constant. Furthermore, the asymptotic behavior of this function is as follows
\begin{equation*}
    \lim_{x \to \infty}{\frac{\li(x)}{\frac{x}{\ln{(x)}}}} = 1, \qquad \lim_{x \to 1^+}{\frac{\li(x)}{\ln{(x-1)}}} = 1.
\end{equation*}
Therefore,

\begin{equation*}
    \li{\left( 1+\frac{k+1}{\eta / B} \right)} + \left\lvert \li{\left( 1+\frac1{\eta / B + k} \right)} \right\rvert \le \mathcal{O}{\left( \frac{k}{\ln{k}} \right)} + \mathcal{O}{\left( \ln{k} \right)} \le \mathcal{O}{\left( \frac{k}{\ln{k}} \right)}.
\end{equation*}

Ultimately, by combining Lemma~\ref{lemma:generalTheta} and Lemma~\ref{lemma:theta1}, we conclude that utilizing a diminishing step size policy $\alpha^{(k)} = \frac{\alpha^{(0)}}{{\left( 1+k / \eta \right)}^\theta}$ with a general $\theta \in (0.5, 1]$ changes the results of Theorem~\ref{theorem:diminishing} in the following manner
\begin{equation*}
    \begin{bmatrix}
			\mathbb{E}{\left[ {\left\| \mathbf{\bar{w}}^{(k B)} - \mathbf{w}^\star \right\|}^2 \right]}
			\\
			\mathbb{E}{\left[ {\left\| \mathbf{W}^{(k B)} - \mathbf{1}_m \mathbf{\bar{w}}^{(k B)} \right\|}^2 \right]}
		\end{bmatrix}
		\le
		\begin{cases}
		    \mathcal{O}{\left( \frac{\ln{k}}{\sqrt{k}} \right)} & \theta = 0.5
		    \\
		    \mathcal{O}{\left( \frac1{k^{1-\theta}} \right)} & 0.5 < \theta < 1
		    \\
		    \mathcal{O}{\left( \frac1{\ln{k}} \right)} & \theta = 1
		\end{cases}.
\end{equation*}

\color{blue}

\section{Further Numerical Experiments} \label{appendix:experiments}

\begin{figure}[t]
    \begin{minipage}{0.49\textwidth}
        \begin{minipage}{\textwidth}
            \begin{subfigure}[b]{0.49\textwidth}
                \centering
                \includegraphics[width=\textwidth]{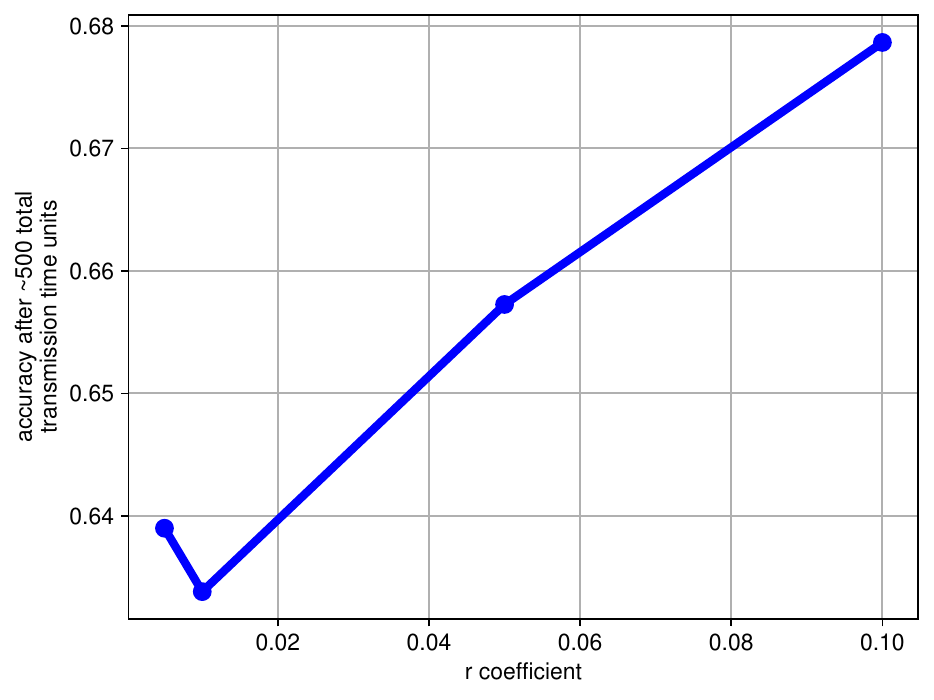}
        		\caption{\textcolor{blue}{IID setup.}}
        		\label{fig:rev:r_coef:iid}
        	\end{subfigure}
        	\begin{subfigure}[b]{0.49\textwidth}
        		\centering
                \includegraphics[width=\textwidth]{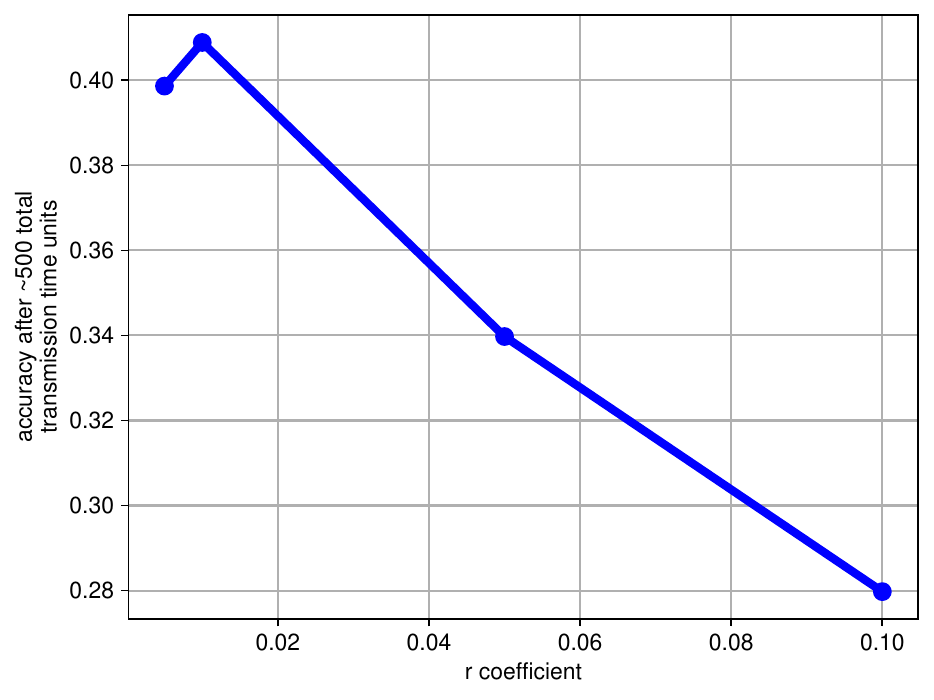}
        		\caption{\textcolor{blue}{Non-IID setup.}}
        		\label{fig:rev:r_coef:noniid}
        	\end{subfigure}
        	\caption{\small{\textcolor{blue}{Varying the threshold coefficient $r$ for our \texttt{EF-HC} algorithm, when using an SVM model on the FMNIST dataset in a network of $m=10$ devices.}}}
        	\label{fig:rev:r_coef}
        \end{minipage}
        
        \begin{minipage}{\textwidth}
            \begin{subfigure}[b]{0.49\textwidth}
        		\centering
                \includegraphics[width=\textwidth]{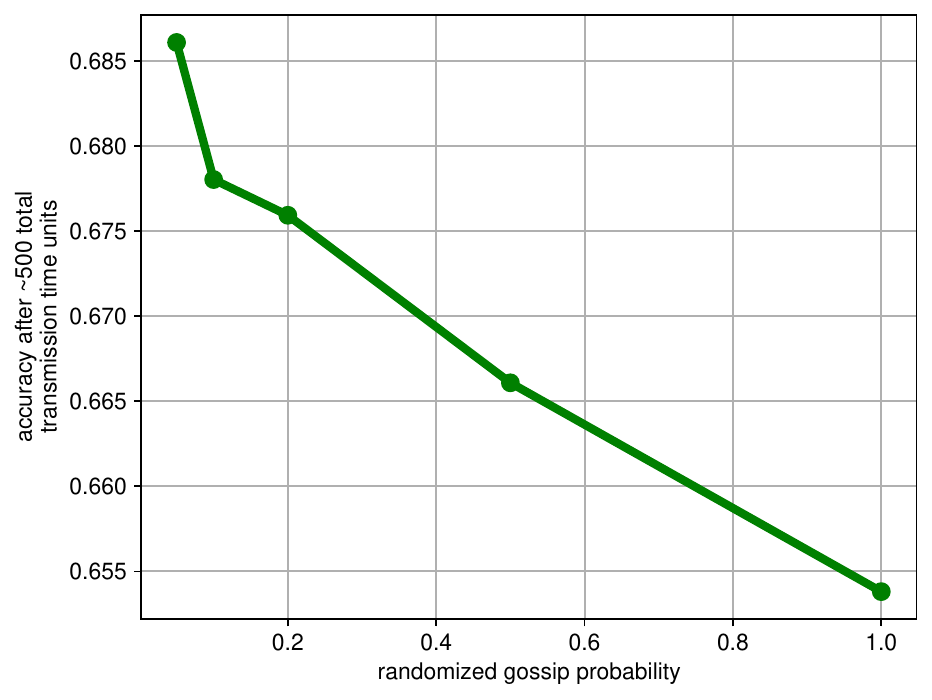}
        		\caption{\textcolor{blue}{IID setup.}}
        		\label{fig:rev:rg_prob:iid}
        	\end{subfigure}
        	\begin{subfigure}[b]{0.49\textwidth}
        		\centering
                \includegraphics[width=\textwidth]{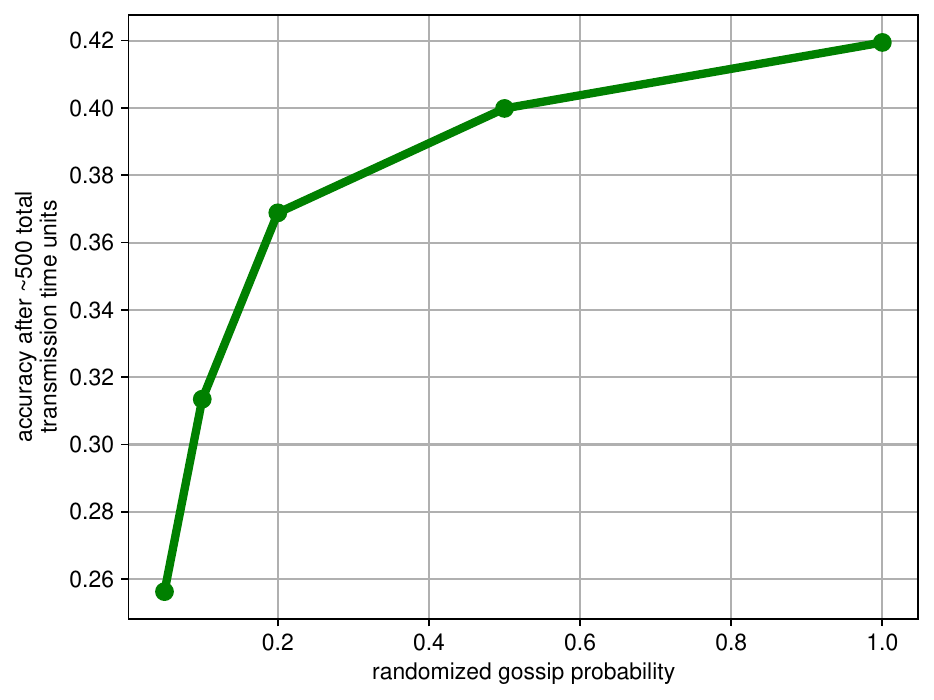}
        		\caption{\textcolor{blue}{Non-IID setup.}}
        		\label{fig:rev:rg_prob:noniid}
        	\end{subfigure}
    	   \caption{\small{\textcolor{blue}{Varying the probability for the randomized gossip (RG) algorithm, when using an SVM model on the FMNIST dataset in a network of $m=10$ devices.}}}
    	   \label{fig:rev:rg_prob}
        \end{minipage}
    \end{minipage}
    \begin{minipage}{0.49\textwidth}
        \centering
        \includegraphics[width=\textwidth]{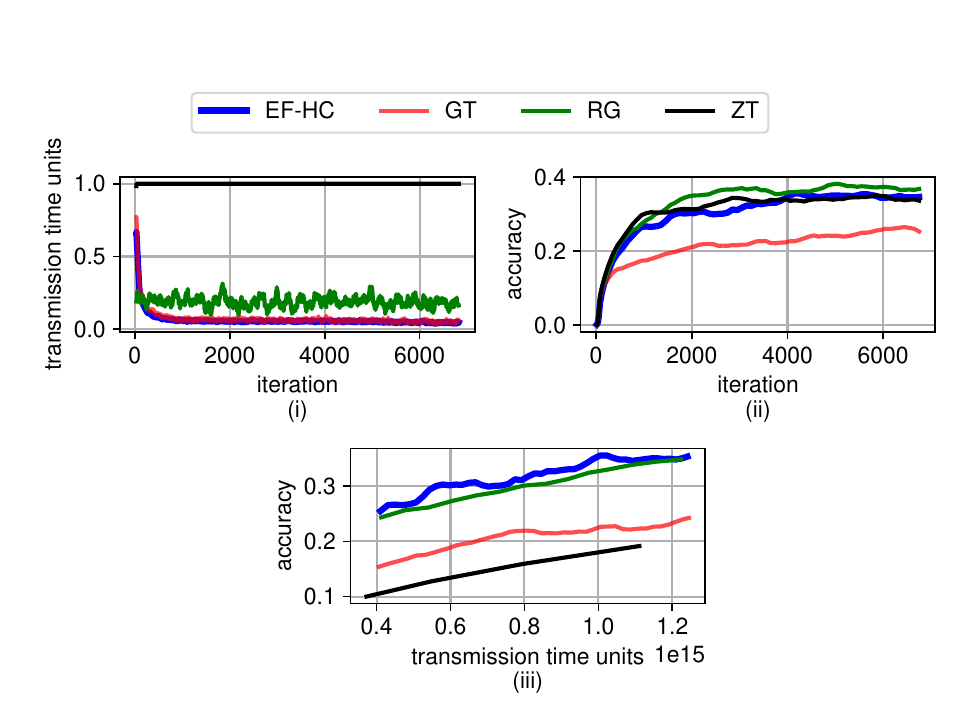}
        \caption{\small{\textcolor{blue}{Performance comparison between our method ({\tt EF-HC}) and the baselines, using a ResNet18 model on the CIFAR10 dataset. We observe that the superiority of our \texttt{EF-HC} algorithm holds (Fig.~\ref{fig:rev:resnet}-(iii)) when using a large deep ML model on a more challenging colored image classification dataset.}}}
        \label{fig:rev:resnet}
    \end{minipage}
    \vspace*{-5mm}
\end{figure}

In this section, we provide the results of additional numerical experiments. In Figs.~\ref{fig:rev:r_coef} and \ref{fig:rev:rg_prob}, we run ablation studies of our \texttt{EF-HC} algorithm and the randomized gossip (RG) algorithm, respectively. Moreover, in Fig.~\ref{fig:rev:resnet} we provide a comparison of our \texttt{EF-HC} algorithm against its event-triggered counterpart with non-heterogeneous global threshold (GT), when using a ResNet18 model on the CIFAR10 dataset.

\textbf{Threshold coefficient $r$.} In Fig.~\ref{fig:rev:r_coef}, we do an ablation study on the threshold coefficient $r$ defined in Eq.~\eqref{eq:thresh}. We observe that the optimal value of $r$ depends highly on the heterogeneity level of the underlying data distribution, as increasing it results in better performance in the IID setup (Fig.~\ref{fig:rev:r_coef:iid}) while the performance degrades by increasing it in the non-IID setup (Fig.~\ref{fig:rev:r_coef:noniid}). This is an expected observation as a lower value of $r$ corresponds to a smaller threshold value, which results in more frequent communications among the devices. Therefore, since in a non-IID setup each device in a decentralized network has access to a limited portion of the data distribution, it can significantly benefit from frequent communications with its neighbors to get exposed to the full data distribution.

On the other hand, the opposite is true in the IID setup. Since each device in the network has access to the full distribution, it does not need to communicate frequently with its neighbors. Therefore, it can save time by communicating only when needed, i.e., avoiding the overhead caused by transmission delays, and instead spend that time on local SGD steps.

\textbf{Randomized gossip probability.} In conventional randomized gossip methods, the probability of inter-device communications is set to the standard value of $1/m$ for a network with $m$ devices. However, we investigate this choice further in this section by doing an ablation study on the probability of conducting communications for the randomized gossip (RG) baseline in Fig.~\ref{fig:rev:rg_prob}. We observe that the optimal value of this probability depends highly on the heterogeneity level of the underlying data distribution, as increasing it results in a performance degradation in the IID setup (Fig.~\ref{fig:rev:rg_prob:iid}) while the performance improved by increasing it in the non-IID setup (Fig.~\ref{fig:rev:rg_prob:noniid}). This is an expected observation as a higher probability corresponds to more frequent communications among the devices. Therefore, since in a non-IID setup each device in a decentralized network has access to a limited portion of the data distribution, it can significantly benefit from frequent communications with its neighbors to get exposed to the full data distribution.

On the other hand, the opposite is true in the IID setup. Since each device in the network has access to the full distribution, it does not need to communicate frequently with its neighbors. Therefore, it can save time by communicating only when needed, i.e., avoiding the overhead caused by transmission delays, and instead spend that time on local SGD steps.

\textbf{ResNet18 on CIFAR10.} To investigate the effectiveness of our \texttt{EF-HC} algorithm on large ML models, we conduct an experiment to train a ResNet18 model on CIFAR10, which is a widely-used colored image classification dataset. We compare our method against all the baselines introduced in Sec.~\ref{ssec:setup} in the IID setup. We run our experiment in a network of $10$ devices, which are fully-connected to each other with heterogeneous bandwidths sampled from a beta distribution $\Beta(0.1, 0.1)$. Our results which are given in Fig.~\ref{fig:rev:resnet} show that \texttt{EF-HC} outperforms all of the baselines in terms of achieved accuracy per transmission delay (Fig.~\ref{fig:rev:resnet}-(iii)). Therefore, our \texttt{EF-HC} algorithm maintains its superiority even when using large ML models. Also, it can be seen that our \texttt{EF-HC} method outperforms GT (which is the non-heterogeneous counterpart of our event-triggered approach) in terms of achieved accuracy per iteration (Fig.~\ref{fig:rev:resnet}-(ii)), illustrating the important role of heterogeneous thresholds in \texttt{EF-HC}. 

\section{Convergence Results for Non-Convex Loss Functions}
\textbf{Outline.} In this section, we conduct the convergence analysis of our methodology for general non-convex loss functions. We start by obtaining some upper bounds for terms involving the local and global gradients in Lemmas~\ref{lemma:grads} and \ref{lemma:graddiv}. Next, in Lemmas~\ref{lemma:nonconvex_lossdec} and \ref{lemma:conconvex_consensus}, we provide the one-step loss and consensus update bounds, respectively. We then extend these results to analyze the consensus error of our methods over a block of iterations of size $B$, i.e., a $B$-step update, in Lemmas~\ref{lemma:B} and \ref{lemma:Bblock}. Ultimately, we prove the main proposition of our work for non-convex loss functions in Proposition~\ref{proposition:nonconvex}, where we obtain conditions for general non-increasing step sizes under which our method is guaranteed to converge to a stationary point. We finally use this result to prove Theorem~\ref{theorem:nonconvex}, which illustrates the convergence rate and stationarity gap of our \texttt{EF-HC} method for general non-convex loss functions using a constant step size.

\begin{lemma} \label{lemma:grads}
    Let Assumption~\ref{assump:smooth_convex_graddiv}-\ref{assump:smoothness} hold. Then the following upper bounds related to the gradients of the global loss function can be obtained in terms of the consensus error $\mathbb{E}{\left[ {\| \mathbf{W}^{(k)} - \mathbf{1}_m \bar{\mathbf{w}}^{(k)} \|}^2 \right]}$ and the global gradient norm $\mathbb{E}{\left[ {\left\| \nabla{F}(\bar{\mathbf{w}}^{(k)}) \right\|}^2 \right]}$.
    \begin{enumerate}[label=(\alph*)]
        \item ${\| \nabla{F}(\bar{\mathbf{w}}^{(k)}) - \bar{\mathbf{\nabla}}^{(k)} \|}^2 \le \frac{L^2}m {\| \mathbf{W}^{(k)} - \mathbf{1}_m \bar{\mathbf{w}}^{(k)} \|}^2.$ \label{lemma:grads:div}

        \item $- \left\langle \nabla{F}(\bar{\mathbf{w}}^{(k)}), \bar{\mathbf{\nabla}}^{(k)} \right\rangle \le -\frac12 {\left\| \nabla{F}(\bar{\mathbf{w}}^{(k)}) \right\|}^2 + \frac{L^2}{2m} {\left\| \mathbf{W}^{(k)} - \mathbf{1}_m \bar{\mathbf{w}}^{(k)} \right\|}^2.$ \label{lemma:grads:inner}

        \item $\frac12 {\left\| \bar{\mathbf{\nabla}}^{(k)} \right\|}^2 \le {\left\| \nabla{F}(\bar{\mathbf{w}}^{(k)}) \right\|}^2 + \frac{L^2}m {\left\| \mathbf{W}^{(k)} - \mathbf{1}_m \bar{\mathbf{w}}^{(k)} \right\|}^2.$ \label{lemma:grads:norm}
    \end{enumerate}

    \begin{proof}
        \ref{lemma:grads:div} We write out the definition of the deviation using Eq.~\eqref{eqn:globalModel} and then use triangle inequality. Finally we invoke Assumption~\ref{assump:smooth_convex_graddiv}-\ref{assump:smoothness} to get
        \begin{equation*}
        	\begin{aligned}
        		{\left\| \nabla{F}(\bar{\mathbf{w}}^{(k)}) - \bar{\mathbf{\nabla}}^{(k)} \right\|}^2 & = {\left\| \frac1m \sum_{i=1}^m{\left( \nabla{F}_i(\bar{\mathbf{w}}^{(k)}) - \nabla{F}_i(\mathbf{w}_i^{(k)}) \right)} \right\|}^2 \le \frac1m \sum_{i=1}^m{{\left\| \nabla{F}_i(\bar{\mathbf{w}}^{(k)}) - \nabla{F}_i(\mathbf{w}_i^{(k)}) \right\|}^2}
        		\\
        		& \le \frac1m \sum_{i=1}^m{L_i^2 {\left\| \bar{\mathbf{w}}^{(k)} - \mathbf{w}_i^{(k)} \right\|}^2} \le \frac{L^2}m {\left\| \mathbf{W}^{(k)} - \mathbf{1}_m \bar{\mathbf{w}}^{(k)} \right\|}^2.
        	\end{aligned}
        \end{equation*}

        \noindent \ref{lemma:grads:inner} We add and subtract the global gradient term $\nabla{F}(\bar{\mathbf{w}}^{(k)})$ within the inner product, and then expand it. Then, Young's inequality gives us
        \begin{equation*}
        	\begin{aligned}
        		- \left\langle \nabla{F}(\bar{\mathbf{w}}^{(k)}), \bar{\mathbf{\nabla}}^{(k)} \right\rangle & = - \left\langle \nabla{F}(\bar{\mathbf{w}}^{(k)}), \bar{\mathbf{\nabla}}^{(k)} - \nabla{F}(\bar{\mathbf{w}}^{(k)}) + \nabla{F}(\bar{\mathbf{w}}^{(k)}) \right\rangle
        		\\
        		& = - {\left\| \nabla{F}(\bar{\mathbf{w}}^{(k)}) \right\|}^2 + \left\langle \nabla{F}(\bar{\mathbf{w}}^{(k)}), \nabla{F}(\bar{\mathbf{w}}^{(k)}) - \bar{\mathbf{\nabla}}^{(k)} \right\rangle
        		\\
        		& \le \frac{-1}2 {\left\| \nabla{F}(\bar{\mathbf{w}}^{(k)}) \right\|}^2 + \frac12 {\left\| \nabla{F}(\bar{\mathbf{w}}^{(k)}) - \bar{\mathbf{\nabla}}^{(k)} \right\|}^2.
        	\end{aligned}
        \end{equation*}
        Using the results of Lemma~\ref{lemma:grads}-\ref{lemma:grads:div} concludes the proof.

        \noindent \ref{lemma:grads:norm} We add and subtract the global gradient term $\nabla{F}(\bar{\mathbf{w}}^{(k)})$ within the norm, and then use triangle inequality to get
        \begin{equation*}
        	\frac12 {\left\| \bar{\mathbf{\nabla}}^{(k)} \right\|}^2 = \frac12 {\left\| \bar{\mathbf{\nabla}}^{(k)} - \nabla{F}(\bar{\mathbf{w}}^{(k)}) + \nabla{F}(\bar{\mathbf{w}}^{(k)}) \right\|}^2 \le {\left\| \nabla{F}(\bar{\mathbf{w}}^{(k)}) \right\|}^2 + {\left\| \nabla{F}(\bar{\mathbf{w}}^{(k)}) - \bar{\mathbf{\nabla}}^{(k)} \right\|}^2.
        \end{equation*}
        Using the results of Lemma~\ref{lemma:grads}-\ref{lemma:grads:div} concludes the proof.
    \end{proof}
\end{lemma}

\begin{lemma} \label{lemma:graddiv}
    Let Assumptions~\ref{assump:smooth_convex_graddiv}-\ref{assump:smoothness} and \ref{assump:smooth_convex_graddiv}-\ref{assump:graddiversity} hold. Then the difference of local gradients from their average is bounded as
    \begin{equation*}
        {\left\| \mathbf{\nabla}^{(k)} - \mathbf{1}_m \bar{\mathbf{\nabla}}^{(k)} \right\|}^2 \le 8 L^2 {\left\| \mathbf{W}^{(k)} - \mathbf{1}_m \bar{\mathbf{w}}^{(k)} \right\|}^2 + 8m \delta^2.
    \end{equation*}
    
    \begin{proof}
        We start by writing out the definition of the local gradients matrix and its average using Eq.~\eqref{eqn:globalModel}. We then use triangle inequality, followed by adding and subtracting $\nabla{F}_i(\bar{w}^{(k)})$, $\nabla{F}(\bar{w}^{(k)})$ and $\nabla{F}_j(\bar{w}^{(k)})$. Ultimately, we use Assumptions~\ref{assump:smooth_convex_graddiv}-\ref{assump:smoothness} and \ref{assump:smooth_convex_graddiv}-\ref{assump:graddiversity} to conclude the proof. We have
        \begin{equation*}
        	\begin{aligned}
        		{\left\| \mathbf{\nabla}^{(k)} - \mathbf{1}_m \bar{\mathbf{\nabla}}^{(k)} \right\|}^2 & = \sum_{i=1}^m{{\left\| \nabla{F}_i(w_i^{(k)}) - \bar{\mathbf{\nabla}}^{(k)} \right\|}^2} = \sum_{i=1}^m{{\left\| \frac1m \sum_{j=1}^m{\left( \nabla{F}_i(w_i^{(k)}) - \nabla{F}_j(w_j^{(k)}) \right)} \right\|}^2}
        		\\
        		& \le \frac1m \sum_{i=1}^m{\sum_{j=1}^m{{\left\| \nabla{F}_i(w_i^{(k)}) - \nabla{F}_j(w_j^{(k)}) \right\|}^2}}
        		\\
        		& \begin{aligned}
        			\,\, \le \frac1m \sum_{i=1}^m \sum_{j=1}^m \Big\| \nabla{F}_i(w_i^{(k)}) - \nabla{F}_i(\bar{w}^{(k)}) & + \nabla{F}_i(\bar{w}^{(k)}) - \nabla{F}(\bar{w}^{(k)}) + \nabla{F}(\bar{w}^{(k)})
        			\\
        			& - \nabla{F}_j(\bar{w}^{(k)}) + \nabla{F}_j(\bar{w}^{(k)}) - \nabla{F}_j(w_j^{(k)}) \Big\|^2
        		\end{aligned}
        		\\
        		& \begin{aligned}
        			\,\, \le \frac4m \sum_{i=1}^m \sum_{j=1}^m \Bigg( & {\left\| \nabla{F}_i(w_i^{(k)}) - \nabla{F}_i(\bar{w}^{(k)}) \right\|}^2 + {\left\| \nabla{F}_i(\bar{w}^{(k)}) - \nabla{F}(\bar{w}^{(k)}) \right\|}^2
        			\\
        			& + {\left\| \nabla{F}(\bar{w}^{(k)}) - \nabla{F}_j(\bar{w}^{(k)}) \right\|}^2 + {\left\| \nabla{F}_j(\bar{w}^{(k)}) - \nabla{F}_j(w_j^{(k)}) \right\|}^2 \Bigg)
        		\end{aligned}
                \\
        		& \le \frac4m \sum_{i=1}^m \sum_{j=1}^m L_i^2 {\left\| \mathbf{w}_i^{(k)} - \bar{\mathbf{w}}^{(k)} \right\|}^2 + \delta_i^2 + \delta_j^2 + L_j^2 {\left\| \bar{\mathbf{w}}^{(k)} - \mathbf{w}_j^{(k)} \right\|}^2
        		\\
        		& = 4 \left[ \sum_{i=1}^m{L_i^2 {\left\| \mathbf{w}_i^{(k)} - \bar{\mathbf{w}}^{(k)} \right\|}^2} + \sum_{i=1}^m{\delta_i^2} + \sum_{j=1}^m{\delta_j^2} + \sum_{j=1}^m{L_j^2 {\left\| \bar{\mathbf{w}}^{(k)} - \mathbf{w}_j^{(k)} \right\|}^2} \right]
        		\\
        		& = 8 \sum_{i=1}^m{\left[ L_i^2 {\left\| \mathbf{w}_i^{(k)} - \bar{\mathbf{w}}^{(k)} \right\|}^2 + \delta_i^2 \right]} \le 8 L^2 {\left\| \mathbf{W}^{(k)} - \mathbf{1}_m \bar{\mathbf{w}}^{(k)} \right\|}^2 + 8m \delta^2.
        	\end{aligned}
        \end{equation*}
    \end{proof}
\end{lemma}

\begin{lemma} \label{lemma:nonconvex_lossdec}
    Let Assumptions~\ref{assump:smooth_convex_graddiv}-\ref{assump:smoothness} and \ref{assump:graderror} hold. Then the expected global loss function evaluated on the average model is bounded as
    \begin{equation*}
        \mathbb{E}{\left[ F(\bar{\mathbf{w}}^{(k+1)}) \right]} \le \mathbb{E} {\left[ F(\bar{\mathbf{w}}^{(k)}) \right]} - a_{11}^{(k)} \mathbb{E}{\left[ {\left\| \nabla{F}(\bar{\mathbf{w}}^{(k)}) \right\|}^2 \right]} + a_{12}^{(k)} \mathbb{E}{\left[ {\left\| \mathbf{W}^{(k)} - \mathbf{1}_m \bar{\mathbf{w}}^{(k)} \right\|}^2 \right]} + c_1^{(k)},
    \end{equation*}
    where $a_{11}^{(k)} = \alpha^{(k)} ( \frac12 - L \alpha^{(k)} )$, $a_{12}^{(k)} = \frac{L^2}{2m} \alpha^{(k)} ( 1 + 2 L \alpha^{(k)} )$ and $c_1^{(k)} = \frac{L}2 {( \alpha^{(k)} )}^2 \frac{\sigma^2}m$.

    \begin{proof}
        We know that Assumption~\ref{assump:smooth_convex_graddiv}-\ref{assump:smoothness} for smooth loss functions have the property that $F_i(\mathbf{w}) \le F_i(\mathbf{w}') + \langle \nabla{F}_i(\mathbf{w}), \mathbf{w} - \mathbf{w}' \rangle + \frac12 L_i \| \mathbf{w} - \mathbf{w}' \|$. Using this property for the smooth global loss function, and evaluating it at $\bar{\mathbf{w}}^{(k+1)}$ and $\bar{\mathbf{w}}^{(k)}$ gives us
        \begin{equation*}
        	\begin{aligned}
        		F(\bar{\mathbf{w}}^{(k+1)}) & \le F(\bar{\mathbf{w}}^{(k)}) + \left\langle \nabla{F}(\bar{\mathbf{w}}^{(k)}), \bar{\mathbf{w}}^{(k+1)} - \bar{\mathbf{w}}^{(k)} \right\rangle + \frac{L}2 {\left\| \bar{\mathbf{w}}^{(k)} - \bar{\mathbf{w}}^{(k+1)} \right\|}^2
        		\\
        		& = F(\bar{\mathbf{w}}^{(k)}) + \left\langle \nabla{F}(\bar{\mathbf{w}}^{(k)}), - \alpha^{(k)} \bar{\mathbf{g}}^{(k)} \right\rangle + \frac{L}2 {\left\| \alpha^{(k)} \bar{\mathbf{g}}^{(k)} \right\|}^2
        		\\
        		& \begin{aligned}
        			\,\, = F(\bar{\mathbf{w}}^{(k)}) & - \alpha^{(k)} \left\langle \nabla{F}(\bar{\mathbf{w}}^{(k)}), \bar{\mathbf{\nabla}}^{(k)} \right\rangle - \alpha^{(k)} \left\langle \nabla{F}( \bar{\mathbf{w}}^{(k)}), \bar{\mathbf{\epsilon}}^{(k)} \right\rangle
        			\\
        			& + \frac{L}2 {\left( \alpha^{(k)} \right)}^2 {\left\| \bar{\mathbf{\nabla}}^{(k)} \right\|}^2 + \frac{L}2 {\left( \alpha^{(k)} \right)}^2 {\left\| \bar{\mathbf{\epsilon}}^{(k)} \right\|}^2 + L {\left( \alpha^{(k)} \right)}^2 \left\langle \bar{\mathbf{\nabla}}^{(k)}, \bar{\mathbf{\epsilon}}^{(k)} \right\rangle,
        		\end{aligned}
        	\end{aligned}
        \end{equation*}
        where we used Eq.~\eqref{eqn:wbarRecursive} for the definition of $\bar{\mathbf{w}}^{(k)}$, and Eq.~\eqref{eqn:g} for the definition of $\mathbf{g}_i^{(k)}$. We next take the expected value of the above inequality to get
        
        \begin{equation*}
        	\begin{aligned}
        		& \begin{aligned}
        			\mathbb{E}{\left[ F(\bar{\mathbf{w}}^{(k+1)}) \right]} \le \mathbb{E} {\left[ F(\bar{\mathbf{w}}^{(k)}) \right]} & - \frac12 \alpha^{(k)} \mathbb{E}{\left[ {\left\| \nabla{F}(\bar{\mathbf{w}}^{(k)}) \right\|}^2 \right]} + \frac{L^2}{2m} \alpha^{(k)} \mathbb{E}{\left[ {\left\| \mathbf{W}^{(k)} - \mathbf{1}_m \bar{\mathbf{w}}^{(k)} \right\|}^2 \right]}
        			\\
        			& + L {\left( \alpha^{(k)} \right)}^2 \mathbb{E}{\left[ {\left\| \nabla{F}(\bar{\mathbf{w}}^{(k)}) \right\|}^2 \right]} + \frac{L^3}m {\left( \alpha^{(k)} \right)}^2 \mathbb{E}{\left[ {\left\| \mathbf{W}^{(k)} - \mathbf{1}_m \bar{\mathbf{w}}^{(k)} \right\|}^2 \right]} + \frac{L}2 {\left( \alpha^{(k)} \right)}^2 \frac{\sigma^2}m
        		\end{aligned}
        		\\
        		& \begin{aligned}
        			= \mathbb{E} {\left[ F(\bar{\mathbf{w}}^{(k)}) \right]} & - \alpha^{(k)} \left[ \frac12 - L \alpha^{(k)} \right] \mathbb{E}{\left[ {\left\| \nabla{F}(\bar{\mathbf{w}}^{(k)}) \right\|}^2 \right]} + \frac{L^2}{2m} \alpha^{(k)} \left( 1 + 2 L \alpha^{(k)} \right) \mathbb{E}{\left[ {\left\| \mathbf{W}^{(k)} - \mathbf{1}_m \bar{\mathbf{w}}^{(k)} \right\|}^2 \right]}
                    \\
                    & + \frac{L}2 {\left( \alpha^{(k)} \right)}^2 \frac{\sigma^2}m,
        		\end{aligned}
        	\end{aligned}
        \end{equation*}
        where we used Assumption~\ref{assump:graderror}-\ref{assump:graderror:mean} and Lemma~\ref{lemma:epsilonbounds} to conclude the proof.
    \end{proof}
\end{lemma}

\begin{lemma} \label{lemma:conconvex_consensus}
    Let Assumptions~\ref{assump:weights}, \ref{assump:smooth_convex_graddiv}-\ref{assump:smoothness}, \ref{assump:smooth_convex_graddiv}-\ref{assump:graddiversity} and \ref{assump:graderror} holds. Then, the consensus error is upper bounded as
    \begin{equation*}
        \mathbb{E} {\left[ {\left\| \mathbf{W}^{(k+1)} - \mathbf{1}_m \bar{\mathbf{w}}^{(k+1)} \right\|}^2 \right]} \le a_{22}^{(k)} \mathbb{E}{\left[ {\left\| \mathbf{W}^{(k)} - \mathbf{1}_m \bar{\mathbf{w}}^{(k)} \right\|}^2 \right]} + c_2^{(k)},
    \end{equation*}
    where $a_{21}^{(k)} = 0$, $a_{22}^{(k)} = 2 [ 1 + 8 {\left( \alpha^{(k)} \right)}^2 L^2 ]$ and $c_2^{(k)} = m {\left( \alpha^{(k)} \right)}^2 ( 16 \delta^2 + \sigma^2 )$

    \begin{proof}
        We start the proof by writing out the definition of model parameters given in Eq.~\eqref{eqn:wRecursiveMatrix}. We then use the definition of stochastic gradients given in Eq.~\eqref{eqn:g}, and continue with the triangle inequality to get
        \begin{equation*}
        	\begin{aligned}
        		\left\| \mathbf{W}^{(k+1)} - \mathbf{1}_m \bar{\mathbf{w}}^{(k+1)} \right\|^2 & = {\left\| \mathbf{P}^{(k)} \mathbf{W}^{(k)} - \mathbf{1}_m \bar{\mathbf{w}}^{(k)} - \alpha^{(k)} \left( \mathbf{G}^{(k)} - \mathbf{1}_m \bar{\mathbf{g}}^{(k)} \right) \right\|}^2
        		\\
        		& \begin{aligned}
        			\,\, \le \Big\| \mathbf{P}^{(k)} & \mathbf{W}^{(k)} - \mathbf{1}_m \bar{\mathbf{w}}^{(k)} - \alpha^{(k)} \left( \mathbf{\nabla}^{(k)} - \mathbf{1}_m \bar{\mathbf{\nabla}}^{(k)} \right) \Big\|^2
        			\\
        			& - 2 \alpha^{(k)} \, \Big\langle \mathbf{P}^{(k)} \mathbf{W}^{(k)} - \mathbf{1}_m \bar{\mathbf{w}}^{(k)} - \alpha^{(k)} \left( \mathbf{\nabla}^{(k)} - \mathbf{1}_m \bar{\mathbf{\nabla}}^{(k)} \right), \mathbf{\epsilon}^{(k)} - \mathbf{1}_m \bar{\mathbf{\epsilon}}^{(k)} \Big\rangle
        			\\
        			& + {\left( \alpha^{(k)} \right)}^2 {\left\| \mathbf{\epsilon}^{(k)} - \mathbf{1}_m \bar{\mathbf{\epsilon}}^{(k)} \right\|}^2
        		\end{aligned}
        		\\
        		& \begin{aligned}
        			\,\, \le 2 & {\left\| \mathbf{P}^{(k)} \mathbf{W}^{(k)} - \mathbf{1}_m \bar{\mathbf{w}}^{(k)} \right\|}^2 + 2 {\left( \alpha^{(k)} \right)}^2 {\left\|\mathbf{\nabla}^{(k)} - \mathbf{1}_m \bar{\mathbf{\nabla}}^{(k)} \right\|}^2
        			\\
        			& - 2 \alpha^{(k)} \bigg\langle\mathbf{P}^{(k)} \mathbf{W}^{(k)} - \mathbf{1}_m \bar{\mathbf{w}}^{(k)} - \alpha^{(k)} \left( \mathbf{\nabla}^{(k)} - \mathbf{1}_m \bar{\mathbf{\nabla}}^{(k)} \right), \left( \mathbf{I}_m - \frac1m \mathbf{1}_m \mathbf{1}_m^T \right) \mathbf{\epsilon}^{(k)} \bigg\rangle
        			\\
        			& + {\left( \alpha^{(k)} \right)}^2 {\left\| \mathbf{\epsilon}^{(k)} - \mathbf{1}_m \bar{\mathbf{\epsilon}}^{(k)} \right\|}^2.
        		\end{aligned}
        	\end{aligned}
        \end{equation*}
        We now take the expected value of the above inequality. Finally, we use Assumption~\ref{assump:graderror}-\ref{assump:graderror:mean} and the results of Lemmas~\ref{lemma:epsilonbounds} and \ref{lemma:graddiv} to get
        \begin{equation*}
        	\begin{aligned}
        	    & \begin{aligned}
            		\mathbb{E} {\left[ {\left\| \mathbf{W}^{(k+1)} - \mathbf{1}_m \bar{\mathbf{w}}^{(k+1)} \right\|}^2 \right]} \le 2 \mathbb{E}{\left[ {\left\| \mathbf{W}^{(k)} - \mathbf{1}_m \bar{\mathbf{w}}^{(k)} \right\|}^2 \right]} & + 16 {\left( \alpha^{(k)} \right)}^2 \bigg( m \delta^2 + L^2 \mathbb{E}{\left[ {\left\| \mathbf{W}^{(k)} - \mathbf{1}_m \bar{\mathbf{w}}^{(k)} \right\|}^2 \right]} \bigg)
            		\\
            		& + m {\left( \alpha^{(k)} \right)}^2 \sigma^2
            	\end{aligned}
                \\
            	& \le 2 \Bigg[ 1 + 8 {\left( \alpha^{(k)} \right)}^2 L^2 \Bigg] \mathbb{E}{\left[ {\left\| \mathbf{W}^{(k)} - \mathbf{1}_m \bar{\mathbf{w}}^{(k)} \right\|}^2 \right]} + m {\left( \alpha^{(k)} \right)}^2 \left( 16 \delta^2 + \sigma^2 \right).
        	\end{aligned}
        \end{equation*}
    \end{proof}
\end{lemma}

\begin{lemma} \label{lemma:B}
    Let Assumptions~\ref{assump:weights}, \ref{assump:smooth_convex_graddiv}-\ref{assump:smoothness}, \ref{assump:smooth_convex_graddiv}-\ref{assump:graddiversity}  and \ref{assump:graderror} hold.
    \begin{enumerate}[label=(\alph*)]
        \item For any iteration $r > k$, it holds that
        \begin{equation*}
        	\mathbb{E}{\left[ {\left\| \mathbf{\nabla}^{(r-1)} - \mathbf{1}_m \bar{\mathbf{\nabla}}^{(r-1)} \right\|}^2 \right]} \le 8m \delta^2 + 8 L^2 \left[ a_{22}^{(r-2:k)} \mathbb{E}{\left[ {\left\| \mathbf{W}^{(k)} - \mathbf{1}_m \mathbf{\bar{w}}^{(k)} \right\|}^2 \right]} + \sum_{l=k+1}^{r-2}{a_{22}^{(r-2:l)} c_2^{(l-1)}} + c_2^{(r-2)} \right],
        \end{equation*}
        in which $a_{22}^{(k)}$ and $c_2^{(k)}$ were defined in Lemma~\ref{lemma:conconvex_consensus}.
        \label{lemma:B:grad}

        \item Let Assumption~\ref{assump:conn} to hold as well. Then the consensus error is bounded as
        \begin{equation*}
            \mathbb{E}{\left[ {\left\| \mathbf{W}^{(k+B)} - \mathbf{1}_m \mathbf{\bar{w}}^{(k+B)} \right\|}^2 \right]} \le \phi_{22}^{(k)} \mathbb{E}{\left[ {\left\| \mathbf{W}^{(k)} - \mathbf{1}_m \mathbf{\bar{w}}^{(k)} \right\|}^2 \right]} + \psi_2^{(k)},
        \end{equation*}
        where $\phi_{21}^{(k)} = 0$, $\phi_{22}^{(k)} = \frac{1+\rho^2}2 + 8 \frac{1+\rho^2}{1-\rho^2} B L^2 \sum_{r=k+1}^{k+B} {\left( \alpha^{(r-1)} \right)}^2 {\left( \rho^{(k+B-1:r)} \right)}^2 a_{22}^{(r-2:k)}$ and $\psi_2^{(k)} = 8 \frac{1+\rho^2}{1-\rho^2} B \sum_{r=k+1}^{k+B} {\left( \alpha^{(r-1)} \right)}^2 {\left( \rho^{(k+B-1:r)} \right)}^2 \left( m \delta^2 + L^2 \left[ \sum_{l=k+1}^{r-2}{a_{22}^{(r-2:l)} c_2^{(l-1)}} + c_2^{(r-2)} \right] \right) + mB \sigma^2 \sum_{r=k+1}^{k+B}{{\left( \alpha^{(r-1)} \right)}^2 {\left( \rho^{(k+B-1:r)} \right)}^2}$. Note that $\rho$ and $B$ are derived in Lemma~\ref{lemma:harnessing:spectral} and Proposition~\ref{proposition:conn}, respectively.
        \label{lemma:B:consensus}
    \end{enumerate}

    \begin{proof}
        \ref{lemma:B:grad} We bound ${\| \mathbf{\nabla}^{(r-1)} - \mathbf{1}_m \mathbf{\bar{\nabla}}^{(r-1)} \|}^2$ using Lemma~\ref{lemma:kplus1}. Then, we use recursion of $\mathbb{E}{\left[ {\left\| \mathbf{W}^{(r-1)} - \mathbf{1}_m \mathbf{\bar{w}}^{(r-1)} \right\|}^2 \right]}$ using Lemma~\ref{lemma:conconvex_consensus} until we reach iteration $k$ in the right-hand side of the inequality as
        \begin{equation*}
        	\begin{aligned}
        		& \mathbb{E}{\left[ {\left\| \mathbf{\nabla}^{(r-1)} - \mathbf{1}_m \bar{\mathbf{\nabla}}^{(r-1)} \right\|}^2 \right]} \le 8m \delta^2 + 8 L^2 \mathbb{E}{\left[ {\left\| \mathbf{W}^{(r-1)} - \mathbf{1}_m \mathbf{\bar{w}}^{(r-1)} \right\|}^2 \right]}
        		\\
        		& \le 8m \delta^2 + 8 L^2 \Bigg[ a_{22}^{(r-2:k)} \mathbb{E}{\left[ {\left\| \mathbf{W}^{(k)} - \mathbf{1}_m \mathbf{\bar{w}}^{(k)} \right\|}^2 \right]} + \sum_{l=k+1}^{r-2}{a_{22}^{(r-2:l)} c_2^{(l-1)}} + c_2^{(r-2)} \Bigg].
        	\end{aligned}
        \end{equation*}

        \noindent \ref{lemma:B:consensus}
        We start by using the definition of the consensus error as given in Eqs.~\eqref{eqn:wExplicitMatrix} and \eqref{eqn:wbarExplicit}. We have
        \begin{equation*}
        	\begin{aligned}
        		& \begin{aligned}
        			\Big\| \mathbf{W}^{(k+B)} - \mathbf{1}_m \mathbf{\bar{w}}^{(k+B)} \Big\|^2 = \Bigg\| \mathbf{P}^{(k+B-1:k)} \mathbf{W}^{(k)} - \mathbf{1}_m \mathbf{\bar{w}}^{(k)} & - \sum_{r=k+1}^{k+B-1}{\alpha^{(r-1)} \left( \mathbf{P}^{(k+B-1:r)} \mathbf{G}^{(r-1)} - \mathbf{1}_m \mathbf{\bar{g}}^{(r-1)} \right)}
        			\\
        			& - \alpha^{(k+B-1)} \left( \mathbf{G}^{(k+B-1)} - \mathbf{1}_m \mathbf{\bar{g}}^{(k+B-1)} \right) \Bigg\|^2
        		\end{aligned}
        		\\
        		& = {\left\| \mathbf{P}^{(k+B-1:k)} \mathbf{W}^{(k)} - \mathbf{1}_m \mathbf{\bar{w}}^{(k)} - \sum_{r=k+1}^{k+B}{\alpha^{(r-1)} \left( \mathbf{P}^{(k+B-1:r)} \mathbf{G}^{(r-1)} - \mathbf{1}_m \mathbf{\bar{g}}^{(r-1)} \right)} \right\|}^2
        		\\
        		& \begin{aligned}
        			\,\, = \Bigg\| & \mathbf{P}^{(k+B-1:k)} \mathbf{W}^{(k)} - \mathbf{1}_m \mathbf{\bar{w}}^{(k)} - \sum_{r=k+1}^{k+B}{\alpha^{(r-1)} \left( \mathbf{P}^{(k+B-1:r)} \mathbf{\nabla}^{(r-1)} - \mathbf{1}_m \mathbf{\bar{\nabla}}^{(r-1)} \right)} \Bigg\|^2
        			\\
        			& \begin{aligned}
        				- 2 \Bigg\langle & \mathbf{P}^{(k+B-1:k)} \mathbf{W}^{(k)} - \mathbf{1}_m \mathbf{\bar{w}}^{(k)} - \sum_{r=k+1}^{k+B}{\alpha^{(r-1)} \left( \mathbf{P}^{(k+B-1:r)} \mathbf{\nabla}^{(r-1)} - \mathbf{1}_m \mathbf{\bar{\nabla}}^{(r-1)} \right)},
        				\\
        				& \sum_{r=k+1}^{k+B}{\alpha^{(r-1)} \left( \mathbf{P}^{(k+B-1:r)} \mathbf{\epsilon}^{(r-1)} - \mathbf{1}_m \mathbf{\bar{\epsilon}}^{(r-1)} \right)} \Bigg\rangle + {\left\| \sum_{r=k+1}^{k+B}{\alpha^{(r-1)} \left( \mathbf{P}^{(k+B-1:r)} \mathbf{\epsilon}^{(r-1)} - \mathbf{1}_m \mathbf{\bar{\epsilon}}^{(r-1)} \right)} \right\|}^2
        			\end{aligned}
        		\end{aligned}
            \end{aligned}
        \end{equation*}
        \begin{equation*}
            \begin{aligned}
                & \begin{aligned}
            		\,\, \le \left( 1 + \frac{1-\rho^2}{2\rho^2} \right) & {\left\| \mathbf{P}^{(k+B-1:k)} \mathbf{W}^{(k)} - \mathbf{1}_m \mathbf{\bar{w}}^{(k)} \right\|}^2
            		\\
            		& + \left( 1 + \frac{2\rho^2}{1-\rho^2} \right) B \sum_{r=k+1}^{k+B} {\left( \alpha^{(r-1)} \right)}^2 {\left\| \mathbf{P}^{(k+B-1:r)} \mathbf{\nabla}^{(r-1)} - \mathbf{1}_m \mathbf{\bar{\nabla}}^{(r-1)} \right\|}^2
            		\\
            		& \begin{aligned}
            		    -2 \Bigg\langle \mathbf{P}^{(k+B-1:k)} & \mathbf{W}^{(k)} - \mathbf{1}_m \mathbf{\bar{w}}^{(k)} - \sum_{r=k+1}^{k+B}{\alpha^{(r-1)} \left( \mathbf{P}^{(k+B-1:r)} \mathbf{\nabla}^{(r-1)} - \mathbf{1}_m \mathbf{\bar{\nabla}}^{(r-1)} \right)},
                        \\
                        & \sum_{r=k+1}^{k+B}{\alpha^{(r-1)} \left( \mathbf{P}^{(k+B-1:r)} - \frac1m \mathbf{1}_m \mathbf{1}_m^T \right) \mathbf{\epsilon}^{(r-1)}} \Bigg\rangle
            		\end{aligned}
            		\\
            		& + B \sum_{r=k+1}^{k+B}{{\left( \alpha^{(r-1)} \right)}^2 {\left\| \mathbf{P}^{(k+B-1:r)} \mathbf{\epsilon}^{(r-1)} - \mathbf{1}_m \mathbf{\bar{\epsilon}}^{(r-1)}  \right\|}^2}
        		\end{aligned}
                \\
                & \begin{aligned}
                    \le \frac{1+\rho^2}{2\rho^2} & {\left( \rho^{(k+B-1:k)} \right)}^2 {\left\| \mathbf{W}^{(k)} - \mathbf{1}_m \mathbf{\bar{w}}^{(k)} \right\|}^2 + \frac{1+\rho^2}{1-\rho^2} B \sum_{r=k+1}^{k+B}{\left( \alpha^{(r-1)} \right)}^2 {\left( \rho^{(k+B-1:r)} \right)}^2 {\left\| \mathbf{\nabla}^{(r-1)} - \mathbf{1}_m \mathbf{\bar{\nabla}}^{(r-1)} \right\|}^2
                    \\
                    & \begin{aligned}
                        - 2 \sum_{r=k+1}^{k+B} \alpha^{(r-1)} \Bigg\langle \mathbf{P}^{(k+B-1:k)} & \mathbf{W}^{(k)} - \mathbf{1}_m \mathbf{\bar{w}}^{(k)} - \sum_{r=k+1}^{k+B}{\alpha^{(r-1)} \left( \mathbf{P}^{(k+B-1:r)} \mathbf{\nabla}^{(r-1)} - \mathbf{1}_m \mathbf{\bar{\nabla}}^{(r-1)} \right)},
                        \\
                        & \left( \mathbf{P}^{(k+B-1:r)} - \frac1m \mathbf{1}_m \mathbf{1}_m^T \right) \mathbf{\epsilon}^{(r-1)} \Bigg\rangle
                    \end{aligned}
                    \\
                    & + B \sum_{r=k+1}^{k+B}{{\left( \alpha^{(r-1)} \right)}^2 {\left( \rho^{(k+B-1:r)} \right)}^2 {\left\| \mathbf{\epsilon}^{(r-1)} - \mathbf{1}_m \mathbf{\bar{\epsilon}}^{(r-1)}  \right\|}^2},
                \end{aligned}
            \end{aligned}
        \end{equation*}
        where we used Young's inequality first, and then invoked Lemma~\ref{lemma:harnessing:spectral}. We next take the expected value of the above expression, and employ the results of Lemmas~\ref{lemma:B}-\ref{lemma:B:grad} and \ref{lemma:epsilonbounds} to get
        \begin{equation*}
        	\begin{aligned}
        		\mathbb{E} \bigg[ \Big\| \mathbf{W}^{(k+B)} & - \mathbf{1}_m \mathbf{\bar{w}}^{(k+B)} \Big\|^2 \bigg] \le
        		\\
        		& \begin{aligned}
        			& \Bigg[ \frac{1+\rho^2}2 + 8 \frac{1+\rho^2}{1-\rho^2} B L^2 \sum_{r=k+1}^{k+B} {\left( \alpha^{(r-1)} \right)}^2 {\left( \rho^{(k+B-1:r)} \right)}^2 a_{22}^{(r-2:k)} \Bigg] \mathbb{E}{\left[ {\left\| \mathbf{W}^{(k)} - \mathbf{1}_m \mathbf{\bar{w}}^{(k)} \right\|}^2 \right]}
        			\\
        			& + 8 \frac{1+\rho^2}{1-\rho^2} B \sum_{r=k+1}^{k+B} {\left( \alpha^{(r-1)} \right)}^2 {\left( \rho^{(k+B-1:r)} \right)}^2 \left( m \delta^2 + L^2 \left[ \sum_{l=k+1}^{r-2}{a_{22}^{(r-2:l)} c_2^{(l-1)}} + c_2^{(r-2)} \right] \right)
        			\\
        			& + mB \sigma^2 \sum_{r=k+1}^{k+B}{{\left( \alpha^{(r-1)} \right)}^2 {\left( \rho^{(k+B-1:r)} \right)}^2}.
        		\end{aligned}
        	\end{aligned}
        \end{equation*}
    \end{proof}
\end{lemma}

\begin{lemma} \label{lemma:Bblock}
    Let Assumptions~\ref{assump:weights}, \ref{assump:smooth_convex_graddiv}-\ref{assump:smoothness}, \ref{assump:smooth_convex_graddiv}-\ref{assump:graddiversity}, \ref{assump:graderror} and \ref{assump:conn} hold.
    Then, we can use Lemma~\ref{lemma:B}-\ref{lemma:B:consensus} to explicitly bound the consensus error as
    \begin{equation*}
        \mathbb{E}{\left[ {\left\| \mathbf{W}^{((k+1)B)} - \mathbf{1}_m \bar{\mathbf{w}}^{((k+1)B)} \right\|}^2 \right]} \le \phi_{22}^{(k:0)} {\left\| \mathbf{W}^{(0)} - \mathbf{1}_m \bar{\mathbf{w}}^{(0)} \right\|}^2 + \sum_{r=0}^k{\phi_{22}^{(k:r+1)} \psi_2^{(r)}}.
    \end{equation*}
        
    \begin{proof}
        We split the training iterations from $0$ to $(k+1)B$ into chunks of size $B$. Then, recursively expanding the consensus error from Lemma~\ref{lemma:B}-\ref{lemma:B:consensus}, we get
        \begin{equation*}
            \mathbb{E}{\left[ {\left\| \mathbf{W}^{((k+1)B)} - \mathbf{1}_m \bar{\mathbf{w}}^{((k+1)B)} \right\|}^2 \right]} \le \left( \prod_{r=0}^k{\phi_{22}^{(r)}} \right) {\left\| \mathbf{W}^{(0)} - \mathbf{1}_m \bar{\mathbf{w}}^{(0)} \right\|}^2 + \sum_{r=0}^k{\left( \prod_{s=r+1}^k{\phi_{22}^{(s)}} \right) \psi_2^{(r)}}.
        \end{equation*}
    \end{proof}
\end{lemma}

\begin{proposition} \label{proposition:nonconvex}
    Let Assumptions~\ref{assump:weights}, \ref{assump:smooth_convex_graddiv}-\ref{assump:smoothness}, \ref{assump:smooth_convex_graddiv}-\ref{assump:graddiversity}, \ref{assump:graderror} and \ref{assump:conn} hold.
    Then, we can use Lemmas~\ref{lemma:nonconvex_lossdec} and \ref{lemma:Bblock} to obtain the following bounds:
    \begin{enumerate}[label=(\alph*)]
        \item The global loss function after $K$ iteration blocks of size $B$ can be bounded as
        \begin{equation*}
            \begin{aligned}
                \mathbb{E}{\left[ F(\bar{\mathbf{w}}^{(KB+1)}) \right]} \le F(\bar{\mathbf{w}}^{(0)}) & - \sum_{k=0}^{KB}{a_{11}^{(k)} \mathbb{E}{\left[ {\left\| \nabla{F}(\bar{\mathbf{w}}^{(k)}) \right\|}^2 \right]}} + {\left\| \mathbf{W}^{(0)} - \mathbf{1}_m \bar{\mathbf{w}}^{(0)} \right\|}^2 \sum_{k=0}^K{\phi_{22}^{(k-1 : 0)} \sum_{b=0}^B{a_{12}^{(kB+b)}}}
                \\
                & + \sum_{r=0}^{K-1}{\psi_2^{(r)} \sum_{k=r+1}^K{\phi_{22}^{(k-1 : r+1)}} \sum_{b=0}^{B-1}{a_{12}^{(kB+b)}}} + \sum_{k=0}^{KB}{c_1^{(k)}}.
            \end{aligned}
        \end{equation*}
        \label{proposition:nonconvex:loss}

        \item If the step size satisfies
        \begin{equation*}
            0 < \alpha^{(kB)} < \frac1{4\sqrt{2} B L} \frac1{{[ 2(1+\Gamma_1) ]}^{B/2}} \frac{1 - \rho^2}{\sqrt{1+\rho^2}},
        \end{equation*}
        with $\Gamma_1 = \frac{1 - \rho^2}{2\sqrt{2} B 2^{B/2} \sqrt{1 + \rho^2}}$,
        then the global loss function given in Proposition~\ref{proposition:nonconvex}-\ref{proposition:nonconvex:loss} can be derived explicitly, which is given in Proposition~\ref{proposition:nonconvex}-\ref{proposition:nonconvex:dec}.
        \label{proposition:nonconvex:stepsize}
        
        \item Let the step size be non-increasing, i.e., $\alpha^{(k+1)} \le \alpha^{(k)}$. Then, using the results of Propositions~\ref{proposition:nonconvex}-\ref{proposition:nonconvex:loss} and \ref{proposition:nonconvex}-\ref{proposition:nonconvex:stepsize}, we get
        \begin{equation*}
            \mathbb{E}{\left[ F(\bar{\mathbf{w}}^{(KB+1)}) \right]} \le F(\bar{\mathbf{w}}^{(0)}) - \sum_{k=0}^{KB}{a_{11}^{(k)} \mathbb{E}{\left[ {\left\| \nabla{F}(\bar{\mathbf{w}}^{(k)}) \right\|}^2 \right]}} + \frac{{\left\| \mathbf{W}^{(0)} - \mathbf{1}_m \bar{\mathbf{w}}^{(0)} \right\|}^2 B a_{12}^{(0)}}{1 - \phi_{22}^{(0)}} + B \sum_{k=0}^{K-1}{\frac{\psi_2^{(k)} a_{12}^{(kB)}}{1 - \phi_{22}^{(k)}}} + \sum_{k=0}^{KB}{c_1^{(k)}}.
        \end{equation*}
        \label{proposition:nonconvex:dec}
    \end{enumerate}

    \begin{proof}
        \ref{proposition:nonconvex:loss} We prove this part of the proposition by recursively expanding Lemma~\ref{lemma:nonconvex_lossdec}, and using algebraic techniques to simplify the bound.
        \begin{equation*}
        	\begin{aligned}
        		& \begin{aligned}
        		    \mathbb{E}{\left[ F(\bar{\mathbf{w}}^{(kB+1)}) \right]} \le \mathbb{E}{\left[ F(\bar{\mathbf{w}}^{(kB)}) \right]} & - a_{11}^{(kB)} \mathbb{E}{\left[ {\left\| \nabla{F}(\bar{\mathbf{w}}^{(kB)}) \right\|}^2 \right]}
        		      \\
                    & + a_{12}^{(kB)} \left[ \phi_{22}^{(k-1 : 0)} {\left\| \mathbf{W}^{(0)} - \mathbf{1}_m \bar{\mathbf{w}}^{(0)} \right\|}^2 + \sum_{r=0}^{k-1}{\phi_{22}^{(k-1 : r+1)} \psi_2^{(r)}} \right] + c_1^{(kB)}
        		\end{aligned}
        		\\
        		& \begin{aligned}
        			= \mathbb{E}{\left[ F(\bar{\mathbf{w}}^{(kB)}) \right]} & - a_{11}^{(kB)} \mathbb{E}{\left[ {\left\| \nabla{F}(\bar{\mathbf{w}}^{(kB)}) \right\|}^2 \right]} + a_{12}^{(kB)} \phi_{22}^{(k-1 : 0)} {\left\| \mathbf{W}^{(0)} - \mathbf{1}_m \bar{\mathbf{w}}^{(0)} \right\|}^2 + a_{12}^{(kB)} \sum_{r=0}^{k-1}{\phi_{22}^{(k-1 : r+1)} \psi_2^{(r)}} + c_1^{(kB)}.
        		\end{aligned}
                \\
            	& \begin{aligned}
            		= F(\bar{\mathbf{w}}^{(0)}) & - \sum_{k=0}^{KB}{a_{11}^{(k)} \mathbb{E}{\left[ {\left\| \nabla{F}(\bar{\mathbf{w}}^{(k)}) \right\|}^2 \right]}} + \sum_{k=0}^{K-1}{\phi_{22}^{(k-1 : 0)} \sum_{b=0}^{B-1}{a_{12}^{(kB+b)} {\left\| \mathbf{W}^{(0)} - \mathbf{1}_m \bar{\mathbf{w}}^{(0)} \right\|}^2}}
            		\\
            		& + \phi_{22}^{(K-1 : 0)} a_{12}^{(KB)} {\left\| \mathbf{W}^{(0)} - \mathbf{1}_m \bar{\mathbf{w}}^{(0)} \right\|}^2 + \sum_{k=0}^{K-1}{\sum_{r=0}^{k-1}{\phi_{22}^{(k-1 : r+1)} \psi_2^{(r)}} \sum_{b=0}^{B-1}{a_{12}^{(kB+b)}}} + a_{12}^{(KB)} \sum_{r=0}^{K-1}{\phi_{22}^{(K-1 : r+1)} \psi_2^{(r)}}
            		\\
            		& + \sum_{k=0}^{KB}{c_1^{(k)}}
            	\end{aligned}
                \\
        		& \begin{aligned}
        			= F(\bar{\mathbf{w}}^{(0)}) & - \sum_{k=0}^{KB}{a_{11}^{(k)} \mathbb{E}{\left[ {\left\| \nabla{F}(\bar{\mathbf{w}}^{(k)}) \right\|}^2 \right]}} + {\left\| \mathbf{W}^{(0)} - \mathbf{1}_m \bar{\mathbf{w}}^{(0)} \right\|}^2 \left( \sum_{k=0}^{K-1}{\phi_{22}^{(k-1 : 0)} \sum_{b=0}^{B-1}{a_{12}^{(kB+b)}}} + \phi_{22}^{(K-1 : 0)} a_{12}^{(KB)} \right)
        			\\
        			& + \sum_{r=0}^{K-1}{\psi_2^{(r)} \sum_{k=r+1}^{K-1}{\phi_{22}^{(k-1 : r+1)}} \sum_{b=0}^{B-1}{a_{12}^{(kB+b)}}} + a_{12}^{(KB)} \sum_{r=0}^{K-1}{\phi_{22}^{(K-1 : r+1)} \psi_2^{(r)}} + \sum_{k=0}^{KB}{c_1^{(k)}}
        		\end{aligned}
        		\\
        		& \begin{aligned}
        			\le F(\bar{\mathbf{w}}^{(0)}) & - \sum_{k=0}^{KB}{a_{11}^{(k)} \mathbb{E}{\left[ {\left\| \nabla{F}(\bar{\mathbf{w}}^{(k)}) \right\|}^2 \right]}} + {\left\| \mathbf{W}^{(0)} - \mathbf{1}_m \bar{\mathbf{w}}^{(0)} \right\|}^2 \sum_{k=0}^K{\phi_{22}^{(k-1 : 0)} \sum_{b=0}^{B-1}{a_{12}^{(kB+b)}}}
        			\\
        			& + \sum_{r=0}^{K-1}{\psi_2^{(r)} \sum_{k=r+1}^K{\phi_{22}^{(k-1 : r+1)}} \sum_{b=0}^{B-1}{a_{12}^{(kB+b)}}} + \sum_{k=0}^{KB}{c_1^{(k)}}.
        		\end{aligned}
        	\end{aligned}
        \end{equation*}
        
        \ref{proposition:nonconvex:stepsize}
        For the loss to be decremented at every iteration, we must have
        \begin{equation*}
            a_{11}^{(k)} > 0 \qquad \Rightarrow 0 < \alpha^{(k)} < \frac1{2L}.
        \end{equation*}
        We can alternatively right this using a scalar $0 < \Gamma_1 < 1$ as
        \begin{equation*}
             \alpha^{(k)} < \frac{\Gamma_1}{2L} \qquad \Rightarrow a_{11}^{(k)} > \frac{1-\Gamma_1}2 \alpha^{(k)}.
        \end{equation*}
        Note that this also results in $a_{12}^{(k)} < \frac{L^2}{2m} \alpha^{(k)} ( 1 + \Gamma_1 )$ and $a_{22}^{(k)} = 2 [ 1 + 2 \Gamma_1^2 ]$.
        Next, in order to find explicit expressions for the upper bound, we enforce $\phi_{22}^{(k)} < 1$, which can equivalently established by ensuring $\phi_{22}^{(k)} < \frac{3 + \rho^2}4$. We first have
        \begin{equation*}
            \begin{aligned}
                \phi_{22}^{(k)} & = \frac{1+\rho^2}2 + 8 \frac{1+\rho^2}{1-\rho^2} B L^2 \sum_{r=kB+1}^{(k+1)B} {\left( \alpha^{(r-1)} \right)}^2 {\left( \rho^{((k+1)B-1:r)} \right)}^2 a_{22}^{(r-2:kB)}
                \\
                & < \frac{1+\rho^2}2 + 8 \frac{1+\rho^2}{1-\rho^2} B^2 L^2 {\left( \alpha^{(kB)} \right)}^2 a_{22}^{((k+1)B-1:kB)} < \frac{1+\rho^2}2 + 8 \frac{1+\rho^2}{1-\rho^2} B^2 L^2 {\left( \alpha^{(kB)} \right)}^2 {[ 2(1+2\Gamma_1^2) ]}^B.
            \end{aligned}
        \end{equation*}
        Therefore, we can write
        \begin{equation*}
            \begin{gathered}
                \phi_{22}^{(k)} < \frac{3 + \rho^2}4 \qquad \Rightarrow 8 \frac{1+\rho^2}{1-\rho^2} B^2 L^2 {\left( \alpha^{(kB)} \right)}^2 {[ 2(1+\Gamma_1) ]}^B < \frac{1 - \rho^2}4
                \\
                \Rightarrow \alpha^{(kB)} < \frac1{4\sqrt{2} B L} \frac1{{[ 2(1+\Gamma_1) ]}^{B/2}} \frac{1 - \rho^2}{\sqrt{1+\rho^2}}.
            \end{gathered}
        \end{equation*}
        To find the optimal value for $\Gamma_1$, we need to solve
        \begin{equation*}
            \frac1{4\sqrt{2} B L} \frac1{{[ 2(1+\Gamma_1) ]}^{B/2}} \frac{1 - \rho^2}{\sqrt{1+\rho^2}} = \frac{\Gamma_1}{2L} \qquad \Rightarrow \Gamma_1 (1 + \Gamma_1)^{B/2} = \frac{1 - \rho^2}{2\sqrt{2} B 2^{B/2} \sqrt{1 + \rho^2}},
        \end{equation*}
        which is an equation whose roots lie in the interval $\left( 0, \frac{1 - \rho^2}{2\sqrt{2} B 2^{B/2} \sqrt{1 + \rho^2}} \right)$ (see Appendix~\ref{appendix:proposition:nonIncreasing} for similar techniques). Choosing the larger value of this interval, and noting that $\frac{\Gamma_1}{2L}$ and $\frac1{4\sqrt{2} B L} \frac1{{[ 2(1+\Gamma_1) ]}^{B/2}} \frac{1 - \rho^2}{\sqrt{1+\rho^2}}$ are an increasing and a decreasing function in $\Gamma_1$, respectively, concludes the proof.

        \ref{proposition:nonconvex:dec} Starting from Proposition~\ref{proposition:nonconvex}-\ref{proposition:nonconvex:loss}, we further simplify the bound derived there using the fact that $\alpha^{(k)}$ is non-increasing. This implies that $a_{12}^{(k)}$, $c_1^{(k)}$, $\phi_{22}^{(k)}$ and $\psi_2^{(k)}$ are all non-increasing functions in $k$. We thus have
        \begin{equation*}
        	\begin{aligned}
        		& \begin{aligned}
        			\mathbb{E}{\left[ F(\bar{\mathbf{w}}^{(KB+1)}) \right]} \le F(\bar{\mathbf{w}}^{(0)}) & - \sum_{k=0}^{KB}{a_{11}^{(k)} \mathbb{E}{\left[ {\left\| \nabla{F}(\bar{\mathbf{w}}^{(k)}) \right\|}^2 \right]}} + {\left\| \mathbf{W}^{(0)} - \mathbf{1}_m \bar{\mathbf{w}}^{(0)} \right\|}^2 \sum_{k=0}^K{{\left( \phi_{22}^{(0)} \right)}^k B a_{12}^{(kB)}}
        			\\
        			& + \sum_{r=0}^{K-1}{\psi_2^{(r)} \sum_{k=r+1}^K{{ \left( \phi_{22}^{(r+1)} \right)}^{k-r-1}} B a_{12}^{(kB)}} + \sum_{k=0}^{KB}{c_1^{(k)}}
        		\end{aligned}
        		\\
        		& \begin{aligned}
        			\le F(\bar{\mathbf{w}}^{(0)}) & - \sum_{k=0}^{KB}{a_{11}^{(k)} \mathbb{E}{\left[ {\left\| \nabla{F}(\bar{\mathbf{w}}^{(k)}) \right\|}^2 \right]}} + {\left\| \mathbf{W}^{(0)} - \mathbf{1}_m \bar{\mathbf{w}}^{(0)} \right\|}^2 B a_{12}^{(0)} \sum_{k=0}^{\infty}{{\left( \phi_{22}^{(0)} \right)}^k}
        			\\
        			& + \sum_{r=0}^{K-1}{\psi_2^{(r)} B a_{12}^{((r+1)B)} \sum_{u=0}^{K-r-1}{{ \left( \phi_{22}^{(r+1)} \right)}^{u}}} + \sum_{k=0}^{KB}{c_1^{(k)}}
        		\end{aligned}
        		\\
        		& \begin{aligned}
        			\le F(\bar{\mathbf{w}}^{(0)}) & - \sum_{k=0}^{KB}{a_{11}^{(k)} \mathbb{E}{\left[ {\left\| \nabla{F}(\bar{\mathbf{w}}^{(k)}) \right\|}^2 \right]}} + {\left\| \mathbf{W}^{(0)} - \mathbf{1}_m \bar{\mathbf{w}}^{(0)} \right\|}^2 B a_{12}^{(0)} \sum_{k=0}^{\infty}{{\left( \phi_{22}^{(0)} \right)}^k}
        			\\
        			& + B \sum_{r=0}^{K-1}{\psi_2^{(r)} a_{12}^{((r+1)B)} \sum_{u=0}^\infty{{ \left( \phi_{22}^{(r+1)} \right)}^{u}}} + \sum_{k=0}^{KB}{c_1^{(k)}}
        		\end{aligned}
                \\
        		& \le F(\bar{\mathbf{w}}^{(0)}) - \sum_{k=0}^{KB}{a_{11}^{(k)} \mathbb{E}{\left[ {\left\| \nabla{F}(\bar{\mathbf{w}}^{(k)}) \right\|}^2 \right]}} + {\left\| \mathbf{W}^{(0)} - \mathbf{1}_m \bar{\mathbf{w}}^{(0)} \right\|}^2 \frac{B a_{12}^{(0)}}{1 - \phi_{22}^{(0)}} + B \sum_{r=0}^{K-1}{\psi_2^{(r)} \frac{a_{12}^{((r+1)B)}}{1 - \phi_{22}^{(r+1)}}} + \sum_{k=0}^{KB}{c_1^{(k)}}
        		\\
        		& \le F(\bar{\mathbf{w}}^{(0)}) - \sum_{k=0}^{KB}{a_{11}^{(k)} \mathbb{E}{\left[ {\left\| \nabla{F}(\bar{\mathbf{w}}^{(k)}) \right\|}^2 \right]}} + \frac{{\left\| \mathbf{W}^{(0)} - \mathbf{1}_m \bar{\mathbf{w}}^{(0)} \right\|}^2 B a_{12}^{(0)}}{1 - \phi_{22}^{(0)}} + B \sum_{k=0}^{K-1}{\frac{\psi_2^{(k)} a_{12}^{(kB)}}{1 - \phi_{22}^{(k)}}} + \sum_{k=0}^{KB}{c_1^{(k)}},
        	\end{aligned}
        \end{equation*}
        where we used the result from Proposition~\ref{proposition:nonconvex}-\ref{proposition:nonconvex:stepsize} in the last two inequalities that $\phi_{22}^{(k)} < 1$.
    \end{proof}
\end{proposition}

\begin{theorem} \label{theorem:nonconvex}
    Let Assumptions~\ref{assump:weights}, \ref{assump:smooth_convex_graddiv}-\ref{assump:smoothness}, \ref{assump:smooth_convex_graddiv}-\ref{assump:graddiversity}, \ref{assump:graderror} and \ref{assump:conn} hold, and a constant step size policy of Assumption~\ref{assump:stepsizes}-\ref{assump:stepsizes:constant} be used. If the step size $\alpha$ satisfies $\alpha < \frac1{4\sqrt{2} B L} \frac1{{[ 2(1+\Gamma_1) ]}^{B/2}} \frac{1 - \rho^2}{\sqrt{1+\rho^2}}$ with $\Gamma_1 = \frac{1 - \rho^2}{2\sqrt{2} B 2^{B/2} \sqrt{1 + \rho^2}}$, then the following bound holds
    \begin{equation*}
        \frac{\sum_{k=0}^{KB}{\mathbb{E}{\left[ {\left\| \nabla{F}(\bar{\mathbf{w}}^{(k)}) \right\|}^2 \right]}}}{KB+1} \le \frac1{w_1} \left[ \frac{F(\bar{\mathbf{w}}^{(0)}) - F^\star}{\alpha (KB+1)} + \frac{B w_2 {\left\| \mathbf{W}^{(0)} - \mathbf{1}_m \bar{\mathbf{w}}^{(0)} \right\|}^2}{KB+1} + \frac{w_2 w_3 B^3 \alpha}{KB+1} + w_4 \alpha^2 \right],
    \end{equation*}
    where $w_1 = \frac{1 - \Gamma_1}2$, $w_2 = \frac{2L^2 (1 + \Gamma_2)}{m (1 - \rho^2)}$, $w_3 = m \left( 8 \frac{1+\rho^2}{1-\rho^2} \left( \delta^2 + L^2 (16 \delta^2 + \sigma^2) \alpha^2 (B-1) {[2 (1 + \Gamma_1^2)]}^{(B-2)} \right) + \sigma^2 \right)$ and $w_4 = \frac{L \sigma^2}{2m}$.
    \begin{proof}
        First, we note that $c_1^{(k)} = \frac{L \sigma^2}{2m} \alpha^2$ and $c_2^{(k)} = m (16 \delta^2 + \sigma^2) \alpha^2$. Therefore, we will have
        \begin{equation*}
            \psi_2^{(k)} \le mB^2 \left( 8 \frac{1+\rho^2}{1-\rho^2} \left( \delta^2 + L^2 (16 \delta^2 + \sigma^2) \alpha^2 (B-1) {[2 (1 + \Gamma_1^2)]}^{(B-2)} \right) + \sigma^2 \right) \alpha^2
        \end{equation*}
        
        \begin{equation*}
            \begin{aligned}
                \mathbb{E}{\left[ F(\bar{\mathbf{w}}^{(KB+1)}) \right]} \le F & (\bar{\mathbf{w}}^{(0)}) - \frac{1 - \Gamma_2 }2 \alpha \sum_{k=0}^{KB}{\mathbb{E}{\left[ {\left\| \nabla{F}(\bar{\mathbf{w}}^{(k)}) \right\|}^2 \right]}} + \frac{2BL^2 (1 + \Gamma_2)}{m (1 - \rho^2)} \alpha {\left\| \mathbf{W}^{(0)} - \mathbf{1}_m \bar{\mathbf{w}}^{(0)} \right\|}^2
                \\
                & + \frac{2B^3L^2 (1 + \Gamma_2)}{(1 - \rho^2)} \left( 8 \frac{1+\rho^2}{1-\rho^2} \left( \delta^2 + L^2 (16 \delta^2 + \sigma^2) \alpha^2 (B-1) {[2 (1 + \Gamma_1^2)]}^{(B-2)} \right) + \sigma^2 \right) \alpha^2
                \\
                & + (KB+1) \frac{L \sigma^2}{2m} \alpha^3.
            \end{aligned}
        \end{equation*}
    \end{proof}
\end{theorem}
\textbf{Discussion on Theorem~\ref{theorem:nonconvex}.} This theorem implies that when our \texttt{EF-HC} is applied on non-convex loss functions, the running average ($\frac1{KB+1} \sum_{k=0}^{KB}{(\cdot)}$) of expected global gradient norms ($\mathbb{E}{[ \| \nabla{F}(\cdot) \|^2 ]}$) of the average model ($\bar{\mathbf{w}}^{(k)}$) is upper bounded. When using a constant step size of $\alpha^{(k)} = \alpha$, this upper bound has a convergence rate of $\mathcal{O}(\frac1K)$, where $K$ is the total number of iteration blocks of size $B$ that the model has been trained for. The stationarity gap of the final model is given as $\mathcal{O}(\frac{w_4}{w_1} \alpha^2) = \mathcal{O}(\frac{L \sigma^2}{m(1-\Gamma_1)} \alpha^2)$, which can be made as small as needed by decreasing the step size $\alpha$. Also, note that $F(\bar{\mathbf{w}^{(0)}}) - F^\star$ is the error of the initial loss value evaluated at the average model ($F(\bar{\mathbf{w}^{(0)}})$) from its minimal value ($F^\star$), and ${\left\| \mathbf{W}^{(0)} - \mathbf{1}_m \bar{\mathbf{w}}^{(0)} \right\|}^2$ is the initial consensus error of models from their average.

If we decide the total of iterations to run the algorithm prior to starting the training, we can choose the constant step size $\alpha$ based on that number. For example, letting $KB$ be the total number of iterations we run the algorithm for, we can use a constant step size $\alpha \propto \frac1{\sqrt{KB}}$, which will result in a convergence rate of $\mathcal{O}(\frac1{\sqrt{K}})$ and a stationarity gap of $\mathcal{O}(\frac{L \sigma^2}{m(1-\Gamma_1)} \frac1K)$. We see that in this case we end up with a slower sub-linear $\mathcal{O}(\frac1{\sqrt{K}})$ rate, but the stationarity gap approaches zero as $K \to \infty$.

\color{black}

\end{appendices}

\end{document}